\documentclass{article}

\usepackage[preprint]{neurips_2024}

\usepackage{amsmath,amsfonts,bm}

\def\eqref#1{equation~\ref{#1}}

\def\1{\bm{1}}

\DeclareMathAlphabet{\mathsfit}{\encodingdefault}{\sfdefault}{m}{sl}
\SetMathAlphabet{\mathsfit}{bold}{\encodingdefault}{\sfdefault}{bx}{n}

\newcommand{\R}{\mathbb{R}}

\usepackage{microtype}
\usepackage{graphicx}
\usepackage{booktabs} 
\usepackage{hyperref}

\usepackage{neurips_2024}

\usepackage{amsmath}
\usepackage{amssymb}
\usepackage{mathtools}
\usepackage{amsthm}

\usepackage[capitalize,noabbrev]{cleveref}
\usepackage{url}
\usepackage{hyperref}
\usepackage{mathtools}
\usepackage{float}
\usepackage{placeins}
\usepackage{titletoc}
\usepackage{algorithm}
\usepackage[noend]{algorithmic}

\usepackage{ dsfont }
\usepackage{amsfonts}  
\usepackage{nicefrac}      
\usepackage{bbm}
\usepackage[labelsep=period, labelfont=it, textfont=normal]{caption}
\usepackage{color}
\usepackage{enumerate}
\usepackage{enumitem}
\usepackage{stmaryrd}
\usepackage{stmaryrd}

\usepackage{subcaption}
\usepackage{xspace}

\usepackage{comment}
\usepackage{xcolor}
\usepackage{multirow}
\usepackage{array}
\usepackage{titletoc}
\usepackage{wrapfig}

\newtheorem{theorem}{Theorem}[section]

\newtheorem{lemma}[theorem]{Lemma}

\newtheorem{definition}[theorem]{Definition}

\newcommand{\code}[1]{{\texttt {#1}}}

\newcommand{\D}{\mathcal{D}}
\renewcommand{\P}{\mathcal{P}}

\newcommand{\M}{\mathcal{M}}

\newcommand{\G}{\mathcal{G}}

\newcommand{\N}{\mathbb{N}}
\newcommand{\p}{\phi}

\newcommand{\traj}{\mathcal{Z}}
\newcommand{\safe}{\text{safe}}

\newcommand{\choice}[2]{#1 \; \code{or} \; #2}
\newcommand{\reach}[1]{\code{reach} \; #1}
\newcommand{\avoid}[1]{\code{avoid} \; #1}

\newcommand{\semantics}[1]{{\llbracket #1 \rrbracket}}

\newcommand{\eventually}[1]{\code{achieve} \; #1}
\newcommand{\always}[1]{~ \code{ensuring} \; #1}
\newcommand{\true}{\code{true}}
\newcommand{\false}{\code{false}}

\newcommand{\task}{\mathsf{R}}
\newcommand{\updatepred}{\mathsf{update}\_\mathsf{pred}}
\newcommand{\updateinit}{\mathsf{update}\_\mathsf{init}}

\newcommand{\inducedistributions}{\mbox{\textnormal{\textsf{InduceDistrbution}}}}
\newcommand{\learnpolicy}{\mbox{\textnormal{\textsf{LearnBasePolicy}}}}
\newcommand{\traintasks}{\textsf{Train}}

\newcommand{\trainkappa}{\mbox{\textnormal{\textsf{LearnKappaCoefficients}}}}

\newcommand{\kplus}{\kappa_{plus}}
\newcommand{\kminus}{\kappa_{minus}}
\newcommand{\rplus}{r_{plus}}
\newcommand{\rminus}{r_{minus}}
\newcommand{\pplus}{\pi_{plus}}
\newcommand{\pminus}{\pi_{minus}}
\newcommand{\Rplus}{R_{plus}}
\newcommand{\Rminus}{R_{minus}}
\newcommand{\perturbkappa}{\mbox{\textnormal{\textsf{PerturbKappa}}}}
\newcommand{\Reward}{\textsf{Reward}}
\newcommand{\kpolicy}{\mbox{\textnormal{\textsf{KappaPolicy}}}}
\newcommand{\Score}{\textsf{Score}}
\newcommand{\ARSupdate}{\mbox{\textnormal{\textsf{DeltaUpdate}}}}
\newcommand{\sampledelta}{\mbox{\textnormal{\textsf{SampleDelta}}}}
\newcommand{\deltas}{\delta_{scale}}

\newcommand{\enqueue}{\textsf{enqueue}}
\newcommand{\dequeue}{\textsf{dequeue}}
\newcommand{\outgoing}{\textsf{OutVertices}}
\newcommand{\incoming}{\textsf{InVertices}}
\newcommand{\remove}{\textsf{remove}}
\newcommand{\append}{\textsf{append}}

\newcommand{\bestincomingedge}{\mbox{\textnormal{\textsf{bestIn}}}}
\newcommand{\decisionboundary}{\mbox{\textnormal{\textsf{LearnGuardConditions}}}}
\newcommand{\De}{\mathcal{D}_e}

\newcommand{\boundary}{\textsf{Guard}}
\newcommand{\traindecisiontree}{\textsf{TrainDecisionTree}}
\newcommand{\EnvInputValues}{\textsf{EnvInputValues}}

\newcommand{\spectrl}{\textsc{Spectrl}\xspace}

\newcommand{\dirl}{\textsc{Base}\xspace}

\newcommand{\genrl}{\mbox{\textnormal{\textsf{GenRL}}}}
\newcommand{\pz}[1]{[\pi_{#1}]}

\newcommand{\hp}[1]{%
  \ifnum#1>0
    \hphantom{\_}%
    \hp{\the\numexpr#1-1\relax}%
  \fi
}

\newcommand{\meta}{\mathbb{G}}

\title{Inductive Generalization in Reinforcement Learning from Specifications}

\author{%
  Vignesh Subramanian\\
  School of Computer Science\\
  Georgia Institute of Technology, USA\\
  \texttt{vignesh@gatech.edu} \\
  \And
  Rohit Kushwah\\
  Department of Computer Science\\
  Indian Institute of Technology - Kanpur, India\\
  \texttt{krohitk@cse.iitk.ac.in} \\
  \And
  Subhajit Roy\\
  Department of Computer Science\\
  Indian Institute of Technology - Kanpur, India\\
  \texttt{subhajit@iitk.ac.in} \\
  \And
  Suguman Bansal\\
  School of Computer Science\\
  Georgia Institute of Technology, USA \\
  \texttt{suguman@gatech.edu} \\
  }

\usepackage{adjustbox}
\usepackage{array}

\newcolumntype{R}[2]{%
    >{\adjustbox{angle=#1,lap=\width-(#2)}\bgroup}%
    l%
    <{\egroup}%
}

\newcommand\rot[1]{#1}

\begin{document}

\maketitle

\begin{abstract}
\label{sec:abstract}
We present a novel {\em inductive generalization framework} for RL from logical specifications. 
Many interesting tasks in RL environments have a natural inductive structure.
These \textit{inductive tasks} have similar overarching goals but they differ inductively in low-level predicates and distributions. We present a generalization procedure that leverages this inductive relationship to learn a higher-order function, a \textit{policy generator}, that generates appropriately \textit{adapted} policies for instances of an inductive task in a zero-shot manner. An evaluation of the proposed approach on a set of challenging control benchmarks demonstrates the promise of our framework in generalizing to unseen policies for long-horizon tasks. 
\end{abstract}
\section{Introduction}
\label{sec:intro}

Recent years have seen an emergence of {\em Reinforcement Learning from logical specifications}~\cite{aksaray2016q,alur2022framework,bansal2022specification,brafman2018ltlf,de2019foundations, hasanbeig2018logically,jothimurugan2022specification,littman2017environmentindependent, hasanbeig2019, yuan2019modular, moritz2019, ijcai2019-0557, jiang2020temporallogicbased,li2017reinforcement,icarte2018using,jothimurugan2021compositional}. Here, the task is expressed using high-level logical specifications rather than as low-level {\em rewards}. Logic specifications have received traction because of (a). the relative ease of expressing complex long-horizon tasks compared to rewards and (b). their impressive ability to efficiently scale learning to long-horizon tasks. 

This work investigates the problem of {\em generalization} in RL from logical specifications. Generalization refers to the ability to extrapolate to unseen tasks. In the context of RL from logical specifications, most works on generalization have focused on zero-shot generalization to unseen specifications that can be constructed from the seen (training) specifications~\cite{kuo2020encoding,vaezipoor2021ltl2action,liu2023skill,xu2022generalizing,leonsystematic}. 

We propose an alternate form of generalization from logical specifications. Our framework is based on the hypothesis that if there exists an \textit{inductive} structure amongst task instances, it is likely that there is an inductive structure amongst their policies. To this end, we design a generalization approach that leverages the inductive relation between the policies of the training tasks to obtain policies for unseen tasks in a zero-shot manner. Let us detail the motivations for our work via an example.

\begin{figure*}[t]
    \centering
    \begin{subfigure}{0.22\textwidth}
        \centering
        \includegraphics[width=\linewidth]{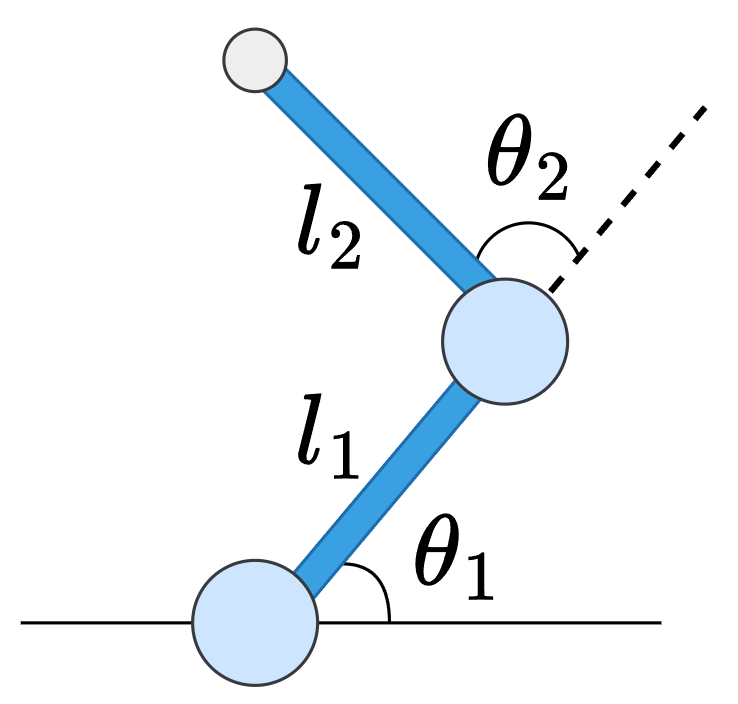}
        \caption{Agent dynamics}
        \label{fig:reacherdyn}
    \end{subfigure}
    \hfill
    \begin{subfigure}{0.35\textwidth}
        \centering
        \includegraphics[width=\linewidth]{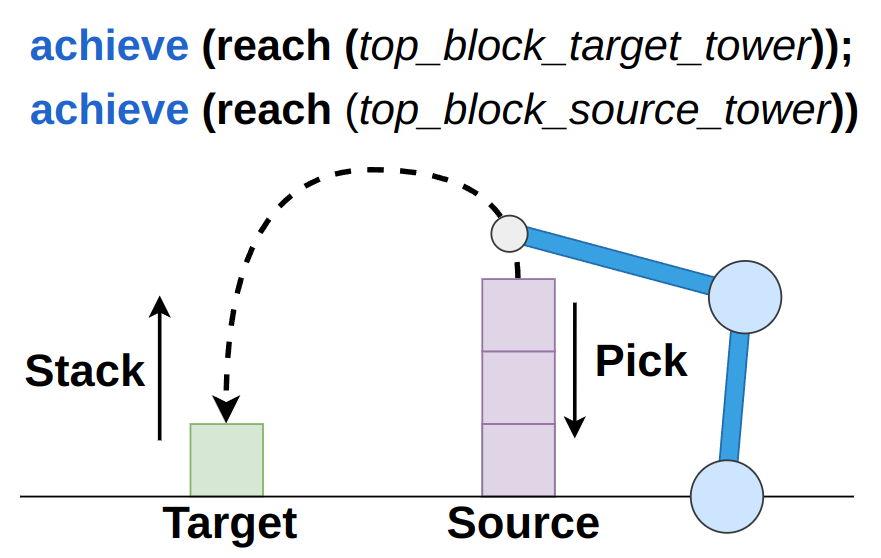}
        \caption{Inductive Task}
        \label{fig:r4illus}
    \end{subfigure}
    \hfill
    \begin{subfigure}{0.30\textwidth}
        \centering
        \includegraphics[width=\linewidth]{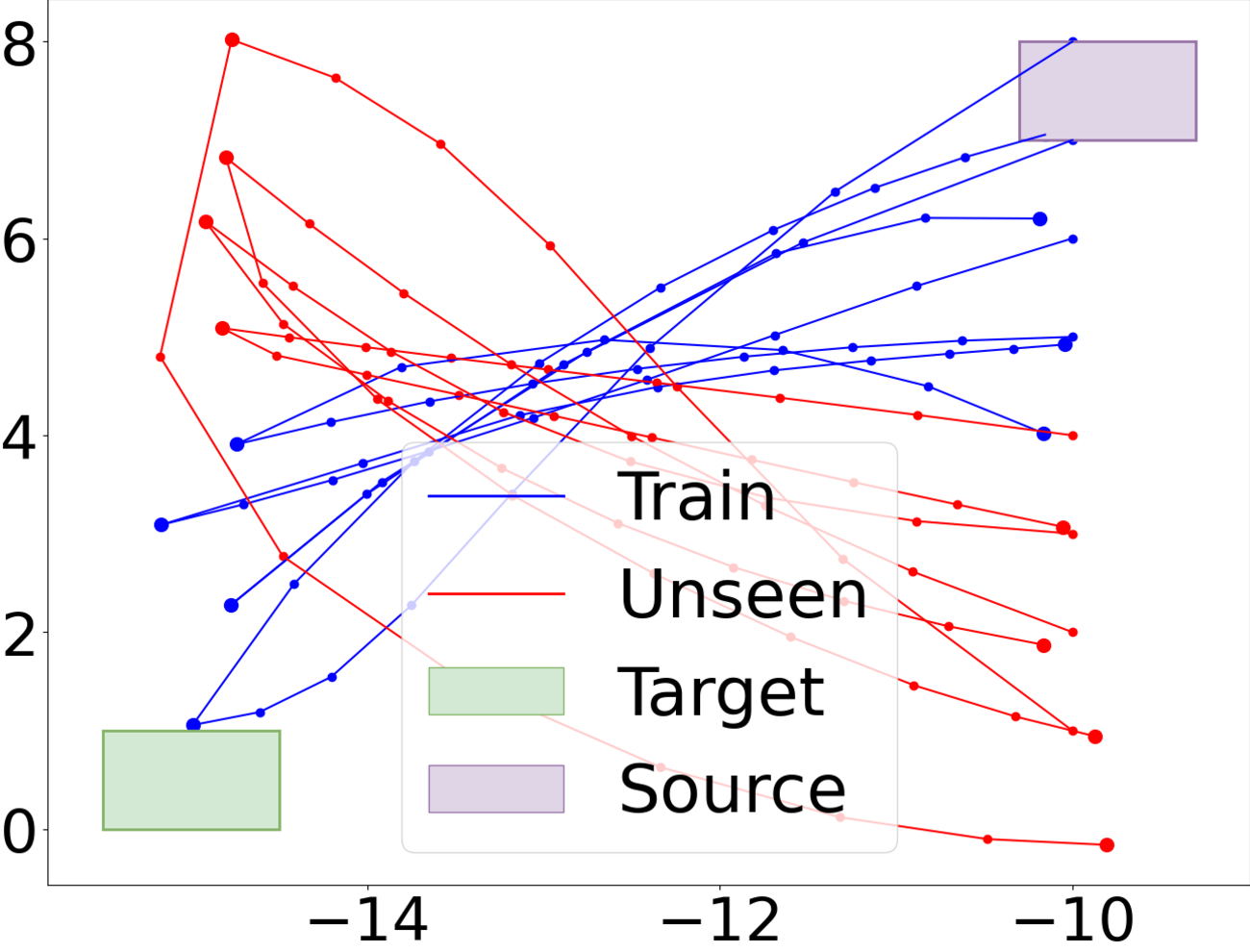}
        \caption{Trajectories}
        \label{fig:r4traj}
    \end{subfigure}
    \caption{Tower Destacking: The task is to pick boxes from \textit{Source} and stack it on \textit{Target}.}

    \label{fig:motivating_example_reacher}
\end{figure*}

{\em Motivating Example \#1.} Figure~\ref{fig:reacherdyn} shows a two-arm robot (inspired by the Reacher-v2 environment in OpenAI gym) that is required to perform certain pick-n-place tasks. The robot controller manipulates the two angles, $\theta_1$ and $\theta_2$, to control the position of the magnetic pickup head. In Figure~\ref{fig:motivating_example_reacher}, the robot is required to pick up the boxes from the \textit{Source} pile and stack them up at \textit{Target}. 
This overall task is essentially a composition of a sequence of smaller \textit{task instances}, where the $i$-{th} instance is: pick the topmost box from a height of $i$ in \textit{Source} and place it at the top of the \textit{Target} pile at height $(h-i)$ (where $h$ is the total number of blocks). Though each of these task instances are ``similar", the respective policies that control $\theta_1$ and $\theta_2$ to position the pickup head, require non-trivial \textit{adaptations} across task instances. The question we explore in this work is: if trained on a few  task instances, can the robot \textit{learn to adapt}, in a \textit{zero-shot} manner,  so as to  accomplish the complete task?

Such generalization is difficult, in general. However, in the above case, the task instances has a well-defined \textit{inductive} structure.
We hypothesize that {\em inductively-related task instances may have inductively-related policies}. Based on this hypothesis, we focus on learning the inductive relationship between the policies of the task instances to extract a \textit{policy generator}: a \textit{higher-order function}, that returns an \textit{adapted} policy for a given inductive task instance.

Figure~\ref{fig:motivating_example_reacher}(c) shows the trajectories of the pickup head with $h=8$ blocks: we trained a policy generator for the robot on picking and placing the first four blocks (shown in blue); The robot could complete the whole task, with adapted policies from the learnt policy generator for the unseen task instances, i.e. pick-n-place of the bottom four blocks are shown in red. 
We see that the policy generator lends \textit{significant adaptability} to the robot to control its $\theta_1$ and $\theta_2$, as the trajectories of the task instances are quite different.

However, our hypothesis may not always hold. It is possible that despite the task being inductive, the policies are not inductive. The motivating example below illustrates this complication.

\begin{figure*}[t]
\centering
\begin{subfigure}[b]{0.32\textwidth}
\centering
\includegraphics[width=\textwidth]{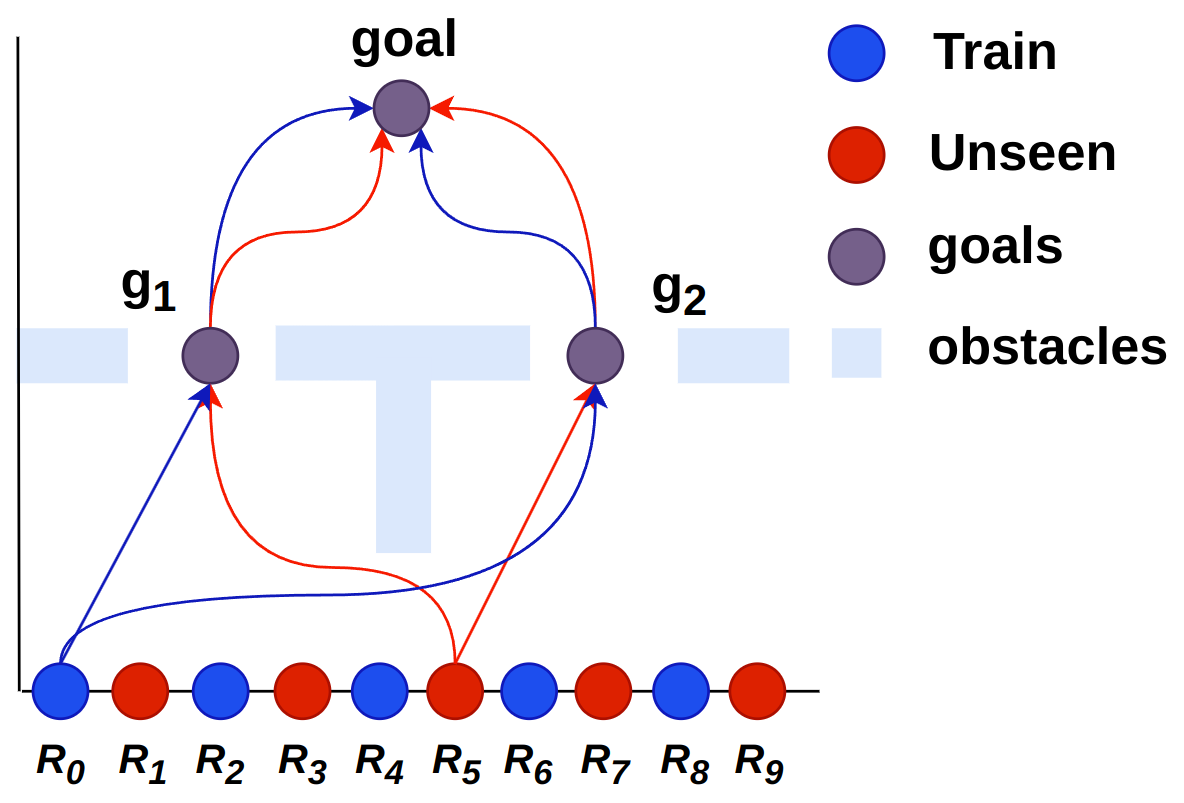}
\caption{Inductive Task}
\label{fig:choice1illus}
\end{subfigure}
\hfill 
\begin{subfigure}[b]{0.32\textwidth}
\centering
\includegraphics[width=0.9\textwidth]{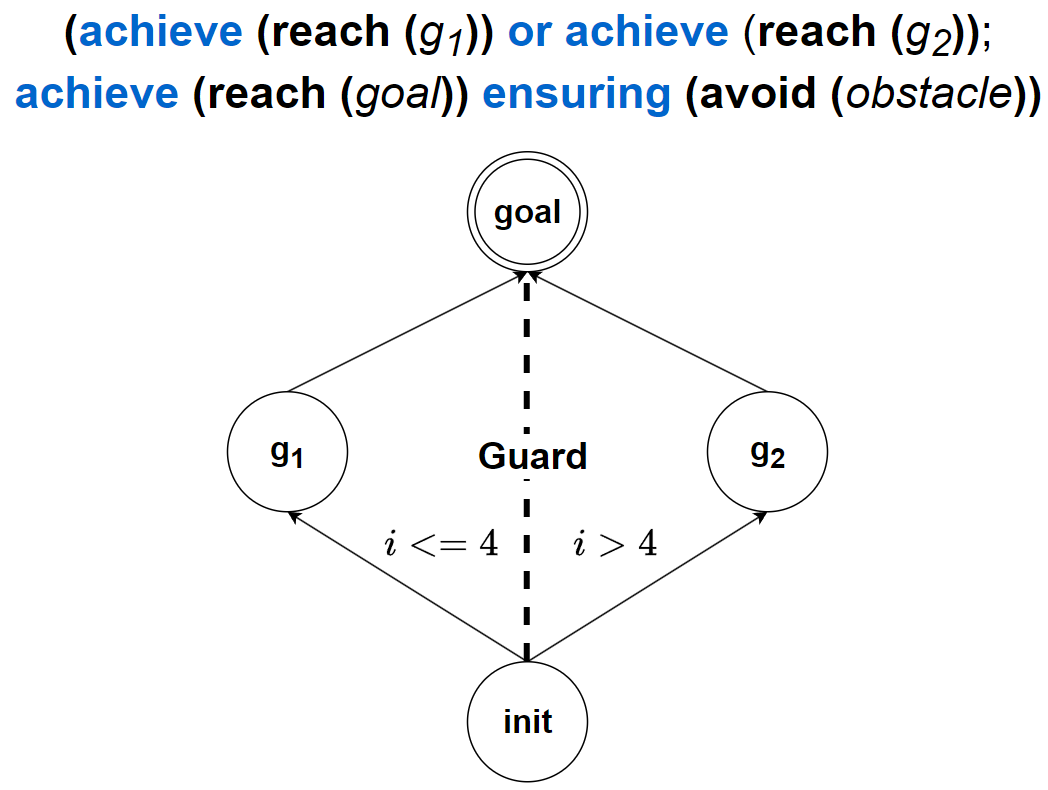}
\caption{Abstract Graph}
\label{fig:graph}
\end{subfigure}    
\hfill
\begin{subfigure}[b]{0.32\textwidth}
\centering
\includegraphics[width=0.9\textwidth]{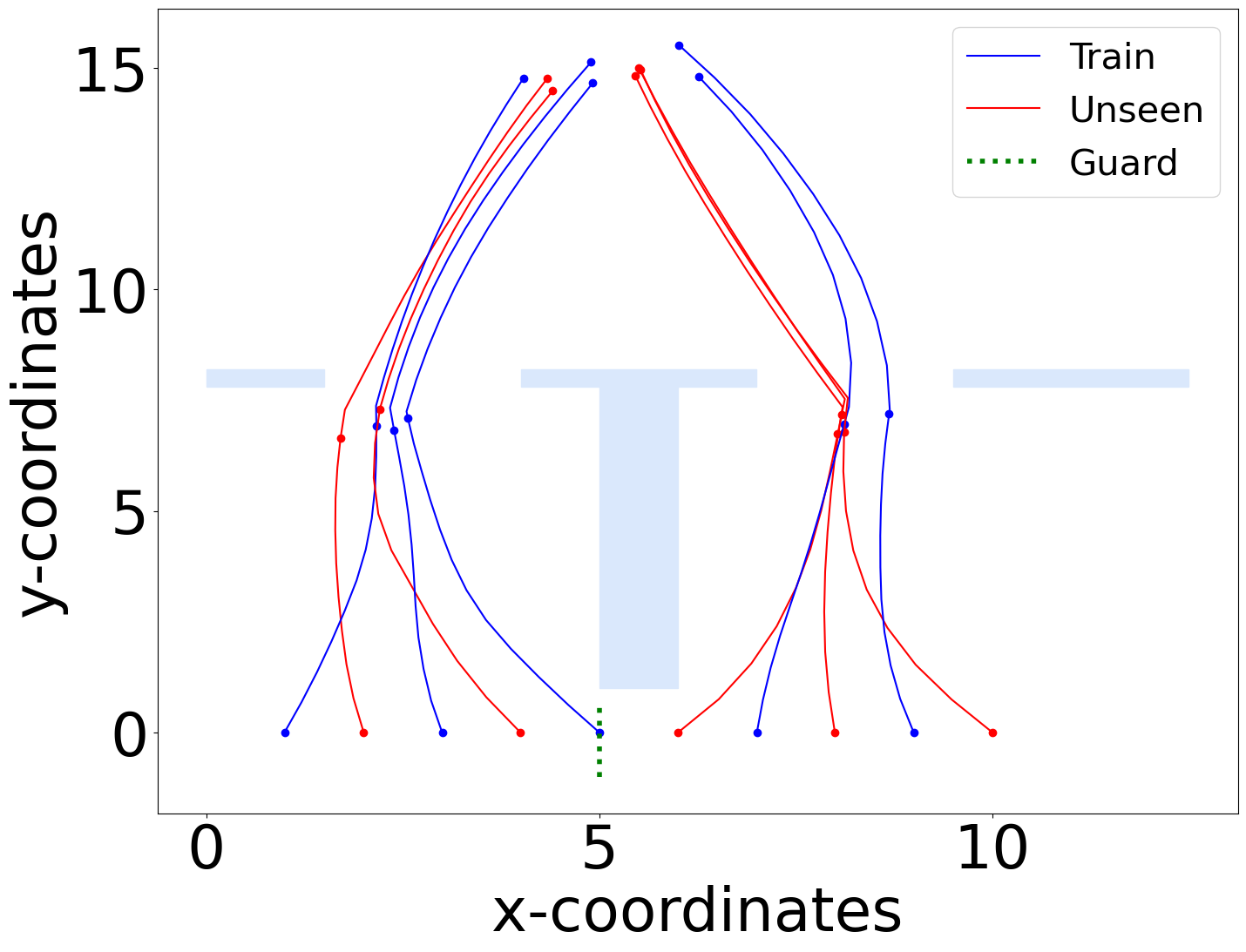}  
\caption{Trajectories}
\label{fig:choice1traj}
\end{subfigure}    
\caption{Choice: visit either $g_1$ or $g_2$, then visit \textsf{goal}; task instances differ in initial state distribution.}

\label{fig:motivating_example}
\end{figure*}

{\em Motivating Example \#2.}
Figure~\ref{fig:motivating_example} illustrates an \textit{inductive task} in a 2D Cartesian plane: in a task instance, the agent is initially located in one of the blue- or red regions marked $R_k$. The goal is to \textit{visit the region} marked $\mathsf{goal}$, \textit{after visiting} one of the intermediate regions $\mathsf{g_1}$ or $\mathsf{g_2}$, while \textit{always avoiding} the obstacles shown in light blue. The task is inductive on the initial position: the $(k+1)$-{th} task can be defined in terms of the $k$-{th} task, by shifting the initial location to the right by $c $ units. 

However, the policies are not inductive: there is a task $\task_k$ such that its policy needs to route through $\mathsf{g_1}$ but the policy of task $\task_{k+1}$ must route through $\mathsf{g_2}$ (eg. $\task_4$ and $\task_5$). But, we may be able to \textit{classify} the task instances into multiple groups, such that all tasks in each group is \textit{inductive} (eg. \{$\task_0$, \dots, $\task_4$\} and \{$\task_5$, \dots, $\task_{9}$\}). 
Our policy generator learns such \textit{branches} such that the task instances on the same decision of the branch have inductive policies.

We summarize our contributions: (a). We introduce a framework to learn inductively generalizable policies for long-horizon tasks. This comprises formalizing the notion of inductively-related tasks based on their logical specification and describing the generalization problem  as learning a higher-order policy generator (Section~\ref{sec:problemformulation}). (b). We describe a procedure to learn a neural policy generator by leveraging the inductive relationship between task instances (Section~\ref{sec:compositional}-Section~\ref{sec:algo}). (c). 
We perform an empirical evaluation of our inductive framework for generalization in learning unseen tasks in complex, long-horizon specifications in continuous environments, popular control environments and robotic pick-n-place tasks, demonstrating the promise of our inductive approach (Section~\ref{sec:empiricaleval}).

\paragraph{Related Work.}
\label{sec:related}

Closest to our work is PSMP~\cite{inala2020synthesizingpp}. PSMP learns inductive policies in the planning setting (with known MDP), not in the RL setting. Despite the planning setting, it is unable to \textit{adapt} its policy to accommodate subgoals that have similar high-level structures but slightly different low-level goals as they learn a single policy for all task instances. Whereas, our approach learns a higher-order policy generator to produce specially adapted policies for each task instance.

Zero-shot generalization is closely related to {\em multi-task learning} and {\em skill transfer} where the common theme is to distil transferable skills from seen tasks to generalize to unseen tasks~\cite{oh2017zero,sohn2018hierarchical,kirk2023survey,sodhani2021multi}. When using logical specifications, existing approaches learn policies for sub-specifications, such as the predicates, and then generalize to boolean and sequential combinations of the learned tasks~\cite{kuo2020encoding,vaezipoor2021ltl2action,liu2023skill,xu2022generalizing,leonsystematic}. Our generalization problem is orthogonal to these: These problems share a common predicate set between the seen (training) and unseen specifications, and their environment distribution remains fixed. In contrast, both the predicate set and the environment distribution differ between every training and unseen RL task in our problem. \cite{jothimurugan2023robust} considers changing distributions, but the predicates are intact. Reward-based generalizable RL have been explored in~{\cite{zisselman2024explore},\cite{taiga2023investigating}}.

Programmatic/logic-based representations of RL policies demonstrate better generalizability than plain function representations of policies (such as NN-policies)~\cite{bastani2018verifiable,verma2018programmatically,cao2022galois,zhu2019inductive}. Non-programmatic policy sketches have been explored~\cite{andreas2017modular}.
Our work differs as we exploit the \textit{natural} inductiveness in task specifications to extract inductive relations for policies and learn a higher-order  {\em policy generator}. 

{\em Meta-learning}~\cite{finn2017model} also learns policies for similar tasks with varying distributions; to the best of our knowledge, none of the meta-learning approaches perform inductive generalization.

\section{Preliminaries}
\label{sec:prelims}

\paragraph{Markov Decision Process (MDP).} The environment in RL is given by a  MDP $\M = (S, A, P, \eta)$ with continuous states $S \subseteq \R^n$, continuous actions $A \subseteq \R^m$, transitions $P(s,a,s') = p(s'\mid s,a)\in\R_{\geq 0}$ (i.e.,  probability density of transitioning from state $s$ to state $s'$ upon taking action $a$), and initial states $\eta: S \rightarrow \R_{\geq 0}$ (i.e., $\eta(s)$ is the probability of the initial state being $s$). 
A \emph{trajectory} $\zeta\in\traj$ is either an infinite sequence $\zeta = s_0\xrightarrow{a_0}s_1\xrightarrow{a_1}\cdots$ or a finite sequence $\zeta=s_0\xrightarrow{a_0}\cdots\xrightarrow{a_{t-1}} s_t$ where $s_i \in S$ and $a_i \in A$. A subtrajectory of $\zeta$ is a subsequence $\zeta_{\ell:k} = s_\ell\xrightarrow{a_\ell}\cdots\xrightarrow{a_{k-1}} s_k$.
We let $\traj_f$ denote the set of finite trajectories.
A (deterministic) \emph{policy}  $\pi:\traj_f \to A$ maps a finite trajectory to a fixed action. Given $\pi$, we can sample a trajectory by sampling an initial state $s_0\sim\eta(\cdot)$, and then iteratively taking the action $a_i=\pi(\zeta_{0:i})$ and sampling a next state $s_{i+1}\sim p(\cdot\mid s_i,a_i)$.

\paragraph{\spectrl Specification Language.}

A \spectrl specification~\cite{jothimurugan2019composable} is defined over a set of \emph{atomic predicates} ${\P}_0$ that ground  environment states, where every $p \in {\P}_0$ is associated with a function $\semantics{p}:S\to\mathbb{B}=\{\true, \false\}$; we say a state $s$ \emph{satisfies} $p$ (denoted $s\models p$) if and only if $\semantics{p}(s)=\true$. 
 For $b\in\mathcal{P}$, the syntax of \spectrl is:
$\p ~::=~ \eventually{b} \mid \p_1 \always{b} \mid \p_1; \p_2 \mid \choice{\p_1}{\p_2}.$
 Each specification $\phi$ corresponds to a function $\semantics{\phi}:\traj\to\mathbb{B}$, and we say $\zeta\in\traj$ satisfies $\phi$ (denoted $\zeta\models\phi$) if and only if $\semantics{\phi}(\zeta)=\true$. Intuitively, `$\eventually$' and `$\always$' are reachability and safety goals, respectively. `;' and `\texttt{or}' refer to sequencing and disjunction, respectively. 
 The formal semantics is present in Appendix~\ref{ap:spectrlsemantics}.

\paragraph{Abstract Graph.}

An  {\em abstract graph} of a \spectrl specification is a DAG-like structure in which vertices represent sets of states (called subgoal regions) and edges represent sets of MDP trajectories that can be used to transition from the source to the target vertex without violating safety constraints. 

\begin{definition}
\rm
An {\em abstract graph} $\G = (U,E,u_0,F,\beta,\traj_{\safe})$ is a directed acyclic graph (DAG) with vertices $U$,
(directed) edges $E\subseteq U\times U$, initial vertex $u_0\in U$, final vertices $F\subseteq U$, subgoal region map $\beta:U\to2^S$ such that for each $u\in U$, $\beta(u)$ is a subgoal region, and \emph{safe trajectories}
$
\traj_\safe = \bigcup_{e \in E}\traj_\safe^e\cup\bigcup_{f \in F}\traj_\safe^f,
$
where $\traj_\safe^e\subseteq\traj$ denotes the safe trajectories for edge $e \in E$ and $\traj_\safe^f\subseteq\traj$ denotes the safe trajectories for final vertex $f\in F$.
\end{definition}
Intuitively, $(U,E)$ is a DAG, and $u_0$ and $F$ define a graph reachability problem for $(U,E)$. Furthermore, $\beta$ and $\traj_{\safe}$ connect $(U,E)$ back to the original MDP $\M$; in particular, for an edge $e=u\to u'$, $\traj_{\safe}^e$ is the set of safe trajectories in $\M$ that can be used to transition from $\beta(u)$ to $\beta(u')$.

The {\em edge policy} $\pi_e$ for an edge $e=u\to u'$
is one that safely transitions from a state in $\beta(u)$ to a state in $\beta(u')$. 
Given edge policies $\Pi$ along with a path
$
\rho=u_0\to u_1\to \cdots\to u_k = u
$
in $\G$, the \emph{path policy} ${\pi}_{\rho}$ navigates from $\beta(u_0)$ to $\beta(u)$. In particular, ${\pi}_{\rho}$ executes $\pi_{u_j\to u_{j+1}}$ (starting from $j=0$) until reaching $\beta(u_{j+1})$, after which it increments $j\gets j+1$ (unless $j=k$). 
Learning an optimal policy for \spectrl is reduced to learning an optimal path policy from the initial to final vertex. This gives rise to a natural compositional learning approach that first learns edge policies and then returns the path policy with the maximum probability of reaching a final vertex~\cite{jothimurugan2021compositional}.

\section{Generalizable RL for Inductive Tasks}
\label{sec:problemformulation}
\sloppy

We introduce the problem of learning generalizable policies for tasks that are inductively related, called {\em inductive tasks}. These appear naturally in several scenarios, as shown in  Figures~\ref{fig:motivating_example_reacher}-\ref{fig:motivating_example}.
Differing from prior work, we learn a {\em policy generator} that supplies separate policies for all task instances.

Notation: 
An {\em RL task} be given by the tuple $(\phi, \eta)$ where $\phi$ is a \spectrl specification and $\eta$ is the initial state distribution in the MDP. We say a trajectory $\zeta = s_0\dots s_t$ satisfies an RL task $(\phi, \eta)$, denoted $\zeta \models (\phi, \eta)$, if $s_0 \sim \eta$ and $\zeta \models \phi$, I.e., $\zeta$ begins in a state sampled from $\eta$ and $\zeta$ satisfies  $\phi$.

\paragraph{Inductive Tasks.}
\label{sec:repeatedtask}

We introduce the notion of an {\em inductive task} as a family of {\em RL tasks} that demonstrate the same overarching structure but differ inductively in the low-level details. 
I.e., an {\em inductive task} is given by a set of enumerable RL tasks such that the $(i+1)$-th
task builds on the $i$-th task by updating the predicates in the specification and/or 
the MDP initial distribution. Formally,

\begin{definition}
    \label{def:repeatedTask}
Let $\P$ and $D(S)$ denote the sets of predicates and state distributions in an MDP, respectively. Let $\p(P)$ denote a \spectrl specification defined over predicates $P \subseteq \P$.

Then, an {\em inductive task} is given by $ \task =(\task_0$, $\updatepred$, $\updateinit)$ where RL task $\task_0 = (\phi(\P_0), \eta_0) $ is the {\em base task},
$\updatepred: \P \mapsto \P$ is the {\em predicate update function}, and
$\updateinit: \D(S) \mapsto \D(S)$ is the  {\em initial distribution update function}.
The enumerable {\em task instances} in $\task$ are given by $\task_0 = (\phi(\P_0), \eta_0)$ and 
$\task_{i+1} = (\phi(\P_{i+1}), \eta_{i+1})$ for  $i> 0$
where 
$\P_{i+1} = \{\updatepred(p) ~|~ p \in \P_i\}$ and 
 $\eta_{i+1}(s) =  \eta_{i}(\updateinit(s))$.
\end{definition}

We denote the $i$-th task instance $\task_i$ by $(\phi_i, \eta_i)$ and refer to task instances $\task_i$ and $\task_{i+1}$ as \textit{adjacent}.

{\em Motivating Example \#1.} For Figure~\ref{fig:motivating_example_reacher}, the inductive task is formalized as: For $j \in \{0, \dots, h\}$, let the predicates $\mathsf{source\_j}$ and $\mathsf{target\_j}$ denote the location of the block at height $j$ in the source and target tower, respectively; let $\eta\_\mathsf{source\_j}$ be a distribution around the block at height $j$ in the source tower.

The base task $\task_0$ is given by  $ ((\eventually(\mathsf{target\_0}));(\eventually(\mathsf{source\_{(h-1)}}));, \eta\mathsf{\_source\_h})$.

The predicate update function updates predicates $\mathsf{source\_j}$ and $\mathsf{target\_j}$ to  $\mathsf{source\_(j-1)} $ and $\mathsf{target\_(j+1)}$, resp. The initial distribution update function updates  $\eta\mathsf{\_source\_j}$ to $\eta\mathsf{\_source\_{(j-1)}}$.

Then, the $j$-th task instance $\task_j  = ((\eventually(\mathsf{target\_{j}}));(\eventually(\mathsf{source\_{(h-j-1)}})), \eta\mathsf{\_source\_{h-j}})$.

{\em Motivating Example \#2.} The Choice example from Figure~\ref{fig:motivating_example} is an inductive task that updates the initial state distribution, the distribution shifted to the right by constant units for adjacent task instances.

\begin{lemma}
\label{lem:commongraph}
For an inductive task $\task$, let $\G_i$ be the abstract graph of the specification of the $i$-th task instance $\task_i$. Then, all the $\G_i$s share a common DAG structure with the same initial and final vertices. 
\vspace{-10pt}
\begin{proof}
The proof follows from the construction of abstract graphs in~\cite{jothimurugan2021compositional}. E.g. for Figure~\ref{fig:motivating_example}, Figure~\ref{fig:graph} shows the \spectrl specification and abstract graph. The abstract graph consists of diamond-shaped DAG with vertices $\mathsf{init}$, $\mathsf{g_1}$, $\mathsf{g_2}$, $\mathsf{goal}$ and edges as shown in Figure~\ref{fig:graph}. 
\end{proof}
\end{lemma}

\paragraph{Generalizable RL for Inductive Tasks.}
\label{sec:policygenerator}

We solve the problem of learning generalizable policies for an inductive task by learning a \textit{policy generator}. The {\em policy generator} for an inductive task $\task$ is a function $\meta: \task \rightarrow \Pi $, where $\Pi$ is the set of all policies in the MDP. E.g., the policy generator for tower-destacking from Figure~\ref{fig:motivating_example_reacher} maps the $j$-th task instance to the policy that displaces the source's $ \mathsf{(h-j)}$-th block to the target's $\mathsf{j}$-th block, then returns to the source's $\mathsf{(h-j-1)}$-th block by manipulating the motor controls $\theta_1$, $\theta_2$. Note these policies are different for each task instance $\task_j$.

\begin{definition}[Learning a Policy Generator]
\label{def:policygenerator}
Given an MDP with unknown transitions, an {\em inductive task} $\task$ and a set of a {\em training task instances} $\traintasks$ s.t. the base task $\task_0 \in\traintasks$, the problem of {\em generalizable RL} is to learn a {\em policy generator} $\meta^*: \task \to \Pi$ such that
    $${\meta}^* \in \operatorname*{\arg\max}_{\meta} \frac{1}{|\traintasks|}\cdot \operatorname*{\sum}_{\task_j\in \traintasks}\Pr_{s_0\sim\eta_j \zeta\sim\D_{\pi_j, s_0}}[\zeta \models \p_j, \eta_j] {\text{ where the policy } \pi_j = \meta^*(\task_j) } $$   
    Then, $\pi_j = \meta^*(\task_j)$  for all $j \in \N$.
\end{definition}
I.e., the policy generator optimizes the policies for all training task instances simultaneously, in an attempt to \textit{generalize}, so as to also derive policies for all task instances not present in $\traintasks$.

\section{Inductive Learning of Policy Generator}
\label{sec:compositional}

Learning a higher-order function such as the policy generator is difficult. To make learning a policy generator feasible, we 
(a) assume {\em inductive relations between policies} of task instances that are inductively related,
(b) leverage similarity between the structure of inductive tasks (Lemma~\ref{lem:commongraph}), and 
(c) leverage compositionality of \spectrl specifications~\cite{jothimurugan2021compositional}.

We leverage the inductive nature of the inductive tasks to learn the policy generator. We base our work on the following hypothesis: 
{\em As two adjacent task instances are related by an inductive relation, there may also exist an inductive relation over the corresponding policies of these tasks}. However, this may not hold for certain tasks (eg. Figure~\ref{fig:motivating_example}). We attempt to overcome this with \textit{compositonality}: instead of learning an inductive policy for the whole task, we divide the task into \textit{subtasks} via the abstract graph, where each edge in the abstract graph corresponds to a subtask. 

\cite{jothimurugan2021compositional} ensures that a policy for a task instance $\task_i$ is given by a path policy in its abstract graph $\G_i$. Lemma~\ref{lem:commongraph} informs that the DAG structure of all graphs $\G_i$ are identical, say $\G$. Hence, the policy generator can be viewed as a map from task instances to path policies from initial to final vertex in the same graph $\G$ (with appropriate instantiation for edge policies in each task instance). Hence, we learn an inductive relation between the corresponding edge policies of the abstract graphs. We formulate the problem to learn such inductive relations in Section~\ref{sec:edgepolicyinduction}.

Last but not least, the edge policies obtained from the inductive relation will result in multiple path policies for each task instance. We are interested in the policy generator to choose the optimal path policy for each task instance. We ensure this by incorporating {\em guards} along vertices in the common DAG $\G$ that route each task instance along the optimal path in the DAG (Section~\ref{sec:finalpolicygenerator}).

\subsection{Learning an Inductive Relation on Edges}
\label{sec:edgepolicyinduction}

This section defines an inductive relation between corresponding edges of the abstract graphs of an inductive task and formulates our approach to learn neural inductive relations.   

Let $e = u\to v$ be an edge in the common DAG $\G$ of an inductive task. Let $\pi_i$ denote the edge policy for the $i$-th task on the edge $e$ in  $\G_i$.

Then, an {\em inductive relation} between these policies is a function $\Omega: \Pi\to \Pi$ s.t. $\pi_{i+1} = \Omega(\pi_i) $. Thus, given the edge policy $\pi_0$ in the base task, the inductive relation $\Omega$ can be inductively ``unrolled" to construct the edge policy for any instance $\task_i$ of an inductive task $\task$. That is,
 \[ \pi_i = \Omega(\pi_{i-1}) = (\Omega (\Omega(\pi_{i-2}))) = \dots = \Omega^{i} (\pi_0) \text{ where, $\Omega^i$ composes $\Omega$ with itself $i$ times. 
}\]

As learning the inductive relation $\Omega$ is difficult, we resort to {\em polynomial approximation}: we approximate the inductive relation $\Omega$ over the policies as an $m$-degree polynomial. This reduces learning $\Omega$ on edges to inferring the $\kappa$-coefficients $(\kappa_m, \cdots, \kappa_0)$ of an $m$-degree $\kappa$-polynomial. Details below:

\paragraph{Neural Policies.} If the policy for the $i$-th task instance  $\pi_i\in \Pi$ is implemented by a neural network with parameter vector $\pz{i}$, then the $m$-degree polynomial inductive relation is given by 
\begin{align}
\label{eq:kappa-polynomail}
  \pz{i+1} = \kappa_{m}\odot \pz{i}^{m} + \kappa_{m-1}\odot \pz{i}^{m-1} + \cdots + \kappa_0  
\end{align}
where, the polynomial coefficients, $\kappa_i$, are vectors with the same dimension as $\pz{i}$; the $\odot$ operator is the  Hadamard product (element-wise multiplication) of the coefficient vectors $\kappa_i$ with the parameter vector (weights and biases) of the policy network $\pi_i$, and `+' is element-wise vector addition. Then as described earlier, with $\pi_0$ as the base policy with parameters $\pz{0}$, the inductive relation $\Omega$ can be inductively ``unrolled" to construct the policy network for $\pi_i$. Hence, in this case, we attempt to learn an inductive relation between the parameter vectors of the policy (neural) network of task instances.

\subsection{Learning the Policy Generator}
\label{sec:finalpolicygenerator}

Next, we describe a policy generator on the common DAG $\G$ between all task instances in $\task$.
Given the inductive relation and base policy for every edge in $\G$, our goal is to describe a mapping for task instances in $\task$ to a path policy in $\G$, as per the edge policies inferred by Equation~\ref{eq:kappa-polynomail}.

Every path from the initial to the final vertex corresponds to a path policy for the $i$-th task. Elaborating further, for degree $m$, let $\kappa^e = (\kappa^e_{m}, \cdots, \kappa^e_0)$ denote the $\kappa$-coefficients on the edge $e\in \G$. Let $\rho = e_1\cdots e_k$ be a path from the initial to a final vertex. Then, a policy for a task $\task_i$ is given by the path policy $\pi_i^{e_1}\cdots \pi_i^{e_k}$ where $[\pi_i^{e_j}]= \Omega^i[\pi_0^{e_j}]$. This requires \textit{selection} of a path policy for each $\task_i$.

We assign {\em guards} at vertices with multiple outgoing edges in $\G$ such that each vertex routes task instances to a unique outgoing edge, ensuring a unique path for every task instance from initial to a final vertex. Formally, a guard in a vertex maps task instances to the outgoing edges from the vertex.

Then, the policy generator for an abstract graph executes as follows: Given a task instance $\task_i$,  it uses the guard conditions to determine its unique path from the initial to a final vertex. It returns the path policy along this path as described above. For example, the abstract graph in Figure~\ref{fig:graph}  has two possible paths: via $g_1$ or $g_2$. We learn a guard, $(i \leq 4)$, that resolves this branching decision at the \textsf{init} node: a task like $\task_2$ would select pass via $g_1$ while $\task_6$ will via $g_2$.

Hence, learning a policy generator for a DAG entails learning the $(m+1)$ $\kappa$-coefficients of the $m$-degree $\kappa$-polynomial  and a base policy for every edge; along with, guard conditions for all vertices with multiple outgoing edges.

\section{Algorithm}
\label{sec:algo}

Algorithm~\ref{algo:genrl-short} (\genrl)~takes as input an inductive task $\task$, the degree $m$ of the $\kappa$-polynomial, and a finite set of training task instances $\traintasks$ (we assume $0 \in \traintasks$) and outputs a {\em policy generator} for $\task$. 
As described above, this entails learning a base policy and the $(m+1)$ $\kappa$-coefficients  for edges and guard conditions for vertices in the common DAG $\G$ (with initial vertex $u_0$) of the inductive task. 

\genrl~operates in two phases: 
 (1) learn $\kappa$-coefficients and base policy for all edges, and (2) learn guard conditions at vertices with multiple outgoing edges.

In the first phase, \genrl~ traverses the vertices in $\G$ in topological order. While processing a vertex $u$,  we record the {\em success probability} of the best probability path from the initial vertex $u_0$ to $u$ for the $i$-th task instance in $P(u,i)$.  We also record $\bestincomingedge(u,i)$ to be the set of incoming vertices to $u$ that are along a best probability path from $u_0$ to $u$ for the task instance $i \in \traintasks$.   Then, 
    \begin{align*}
    P(u_0,i) = 1 &~\text{ and }~ P(u,i) = \operatorname*{\max} \{ P(w,i)\cdot p_i^{w\to u} \mid w\to u \in \mathsf{InEdges}(u) \} \text{ for all }u \neq u_0 \\
    \bestincomingedge(u_0,i) = \emptyset &~\text{ and }~ \bestincomingedge(u,i) = \operatorname*{\arg\max}_w \{ P(w,i)\cdot p_i^{w\to u} \mid w\to u \in \mathsf{InEdges}(u) \} \text{ for all }u \neq u_0 
    \end{align*}
    where $p_i^{w\to u}$ is the estimated success probability of edge policy of $i$-th task instance on edge $w\to u$.

 Next, we induce a state distribution $\eta_i^u$ on vertex $u$ for all task instances $i\in \traintasks$. $\eta_i^{u_0}$ is given by the initial distribution of the task instance $\task_i$.
    For all other vertices $u \neq u_0$, the state distributions are induced along the {\em best probability path} from $u_0$ to $u$. To this end, $\eta_i^u$ is induced from states in $\bestincomingedge(u,i)$ using the leaned edge policies along these incoming edges.
    
     Finally, for all outgoing edges $e= u\to v$, we learn the base policy and $(m+1)$ $\kappa$-coefficients. 
    The base policy $\pi^e_0$  is learned as a neural-network policy using standard RL such that $\pi^e_0$ maximizes the satisfaction of the edge $e$ for the $0$-th task instance.
    I.e., the rewards are designed to encourage $\pi^e_0$ to safely transition from an MDP state in $u$ to an MDP state in $v$ for the 0-th task instance. 
    
    The $\kappa$-coefficients are learned using an adaptation of the ARS (Augmented Random Search) (see Appendix~\ref{ap-sec:algo:kappalearning}). 
    The $\kappa$-coefficients capture an inductive relation between the parameters of the policy networks of adjacent task instances; $[\pi^e_i]$ is the parameter vector for the policy network corresponding to the ${i}$-th task instance. 
    Let $\pi^e_i$ be obtained by unrolling the $\kappa$-polynomial on the base policy parameters for all $i \in \traintasks$. Taking rewards of $\pi^e_i$ to be based on satisfaction of the edge for the $i$-th task instance (as done for the base policy), $\kappa$-polynomials are trained to optimize the $\mathsf{softmin}$ of the rewards over all training task instances. 

The second phase learns the guard conditions  (see Appendix~\ref{sec:algo:decisionboundary}): In addition to ensuring the uniqueness of the path, we require that the guards choose an edge such that the resulting path policy has a high probability of satisfaction. To this end, for each edge $e$, we create a set $S^e$ of task instances such that the $e$ appears on a best probability path to a final vertex for those task instances, using backward DAG traversal and $\bestincomingedge$. 
Finally, the guard on a vertex $u$ is learned as a multi-task classifier with (a) outgoing edges as the class labels, (b). task instance indices as features, and (c). the dataset consists of data points $(i, e)$ s.t. $i \in S^e$ for all outgoing edges $e$ from $u$.

\begin{algorithm}[t]
\small
\caption{\genrl($\task, m, \traintasks$)}
\begin{algorithmic}[1]
\STATE $\G \gets \mathsf{CommonDAG(\task)}$
\WHILE{vertex $u \in \G$ is chosen in topological order}
    \STATE Compute $P(u,i), \bestincomingedge(u,i) $ \textbf{for} all $i \in \traintasks$ 
    \STATE $\eta_i^u \gets \inducedistributions(u, \bestincomingedge(u,i))$ \textbf{for} all $i \in \traintasks$
    \FOR{edge $e = (u,v)\in \mathsf{OutEdges}(u)$} 
    \STATE $\pi^{e}_0  \gets \learnpolicy(e, \eta^{u}_0)$
    \STATE $\kappa^e \gets \trainkappa(e, m+1, \pi^{e}_0, \eta^u, \traintasks)$
    \ENDFOR
\ENDWHILE
\STATE $\boundary \gets \decisionboundary(\G, \bestincomingedge)$ \label{line:decisionboundary}
\STATE \textbf{return} $\kappa^e$, $\pi_0^e$ for all edge 
 and $\boundary$ for all vertices
\end{algorithmic}
\label{algo:genrl-short}
\end{algorithm}

\section{Empirical Evaluation}
\label{sec:empiricaleval}

\paragraph{Baselines.}

Due to the lack of baselines of generalizable RL for long-horizon tasks (see Section~\ref{sec:related})), we choose the closest baseline PSMP~\cite{inala2020synthesizingpp} from inductive generalization in the planning setting (MDP is known)  and create three baselines in the RL setting (MDP is unknown) as ablations of our tool \genrl.  The key distinction between $\genrl$ and the baselines is that \genrl~ learns a policy generator that discharges an \textit{adapted} policy for each task instance. Whereas, the baselines learn a {\em single policy} for all task instances, as is usual in prior generalizable RL approaches. 

Given a set of training task instances \traintasks, our ablations $\dirl$1, $\dirl$2, and $\dirl$3 use different training approaches to learn a single policy that optimizes the satisfaction of all task instances in \traintasks. In each policy update during training, $\dirl$1 updates the policy sequentially for every task instance in \traintasks. Consequently, $\dirl$1 may demonstrate weak convergence and weak generalization. $\dirl$2 and $\dirl$3 overcome this issue by updating the policy w.r.t. an {\em aggregate} (\textsf{softmin}) of updates obtained from sequential updates to the policies of all tasks; $\dirl$3 is similar to $\dirl$2 but allows the policy to also learn over the index of the task instance in addition to the environment state. The intuition is that knowledge of the inductive \textit{ordering} of the task instances can aid generalization.

\paragraph{Setup.} 
$\genrl$ with experimental setup is available\footnote{\href{https://anonymous.4open.science/r/GenRL_Zenith-7EEB/}{https://anonymous.4open.science/r/GenRL\_Zenith-7EEB/}}; details in Appendix~\ref{sec:supplemental}.
We train a $1$-degree $\kappa$-polynomial.  
In each experiment, we record the number of successful task instances on training set \traintasks~and on a testing set of task instances, called {\em Unseen}. 
The probability of success is estimated on 1000 rollouts per task instance. 
We say that a policy $\pi_i$ \textit{succeeds} on task instance $\task_i: (\phi_i, \eta_i)$ if the rollouts $\zeta$ satsify the specification with a probability above $\delta$, that is $\Pr_{\zeta\sim\D_{\pi_i}}[\zeta \models \p_i, \eta_i] > \delta$;.  We set $\delta$ as 0.9.  We report the median of five runs.  
We show selected trajectories produced by trained policies for task instances in \traintasks~(in \textcolor{blue}{blue}) and Unseen (in \textcolor{red}{red}).
For a fair comparison, tools are compared by training and testing on identical \traintasks~ and Unseen, resp. 
We use a cluster running Intel Xeon Gold 6226 CPUs, operating at 2.7 GHz with 24 cores per node. Each node has 192GB of RAM.

\subsection{Long-horizon tasks in environments with simple dynamics}
\label{sec:longhorsimdyn}

\begin{table}[t]
    \centering
    \begin{minipage}{\textwidth} 
        \centering
    \footnotesize
        \setlength{\tabcolsep}{1pt}
        
        \begin{tabular}{|l||c|c|c|c|c||c|c|c|c|c|}
            \hline
            \textbf{Benchmark} & \multicolumn{5}{c||}{\textbf{Successful \traintasks}} & \multicolumn{5}{c|}{\textbf{Successful \textsf{Unseen}}} \\
            \cline{2-11}
             & {\rot{PSMP}} & {\rot{$\dirl$1}} & {\rot{$\dirl$2}} & {\rot{$\dirl$3}} & {\rot{\genrl}} & {\rot{PSMP}} & {\rot{$\dirl$1}} & {\rot{$\dirl$2}} & {\rot{$\dirl$3}} & {\rot{\genrl}} \\
            \hline
            Figure~\ref{fig:1reachmovpsmp} $|\traintasks| = 5$ & \textbf{All} & 0 & \textbf{All} & \textbf{All} & \textbf{All} & \textbf{8} & 0 & 2 & 1 & 4 \\
            Figure~\ref{fig:1reachpsmp}  $|\traintasks| = 10$  & 1 & 0 & 8 & 8 & \textbf{All} & 3 & 0 & 3 & 1 & \textbf{10} \\
            Figure~\ref{fig:1reachobspsmp} $|\traintasks| = 10$ & 1 & 0 & 6 & 9 & \textbf{All} & 1 & 0 & 1 & 0 & \textbf{10} \\
            \hline
        \end{tabular}
        \captionof{table}{Number of successful task instances on $\traintasks$ and Unseen on simple long-horizon tasks}
        \label{tab:1reachability_alltools}
    \end{minipage}
\end{table}

\begin{figure}
\begin{minipage}{0.3\textwidth} 
    \begin{subfigure}{\textwidth}
        \includegraphics[width=\linewidth, height=0.7\linewidth]{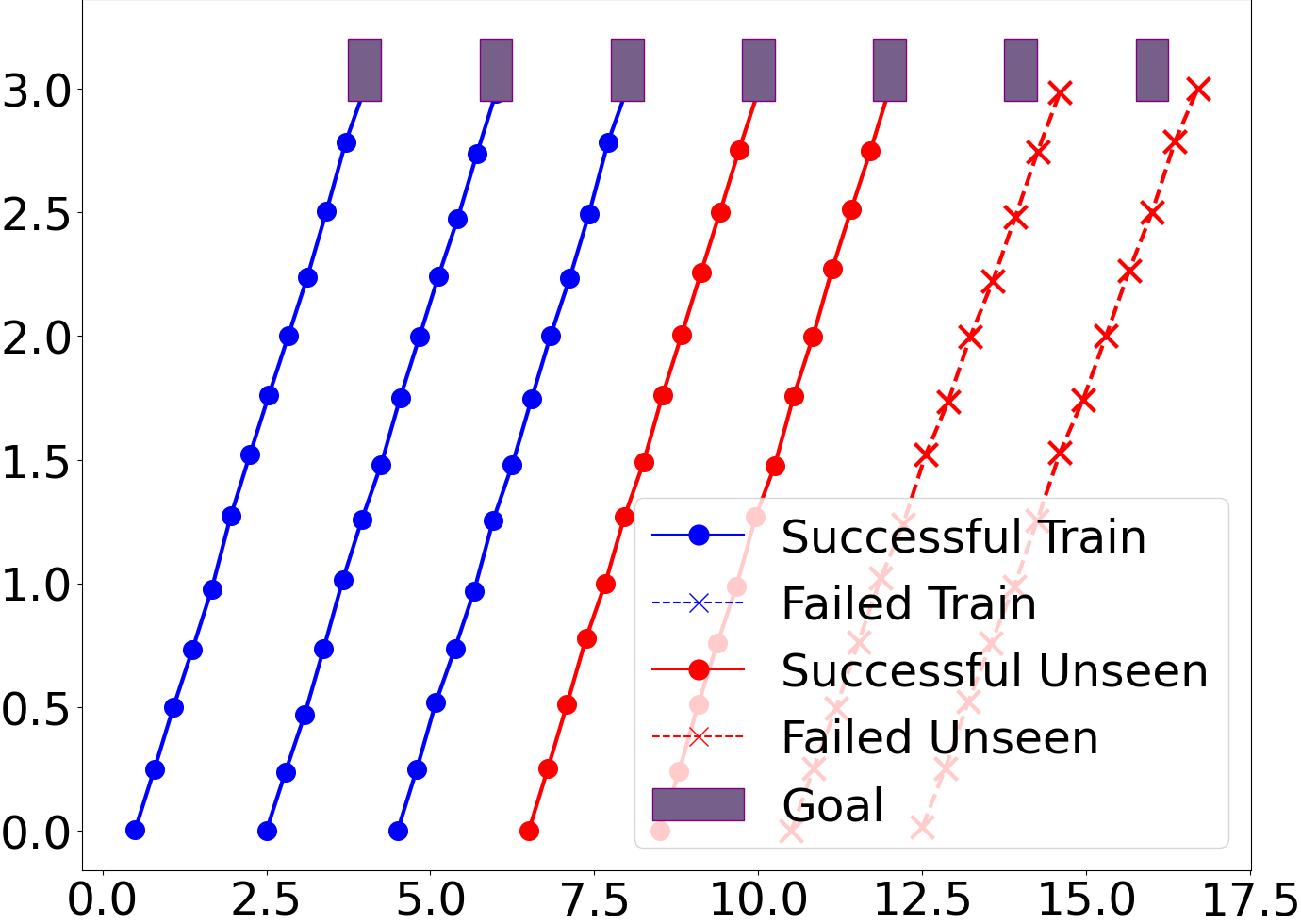}
        
        \caption{GenRL}
        \label{fig:1a}
    \end{subfigure}
    
    \medskip 
    \begin{subfigure}{\textwidth}
        
        \includegraphics[width=\linewidth, height=0.7\linewidth]{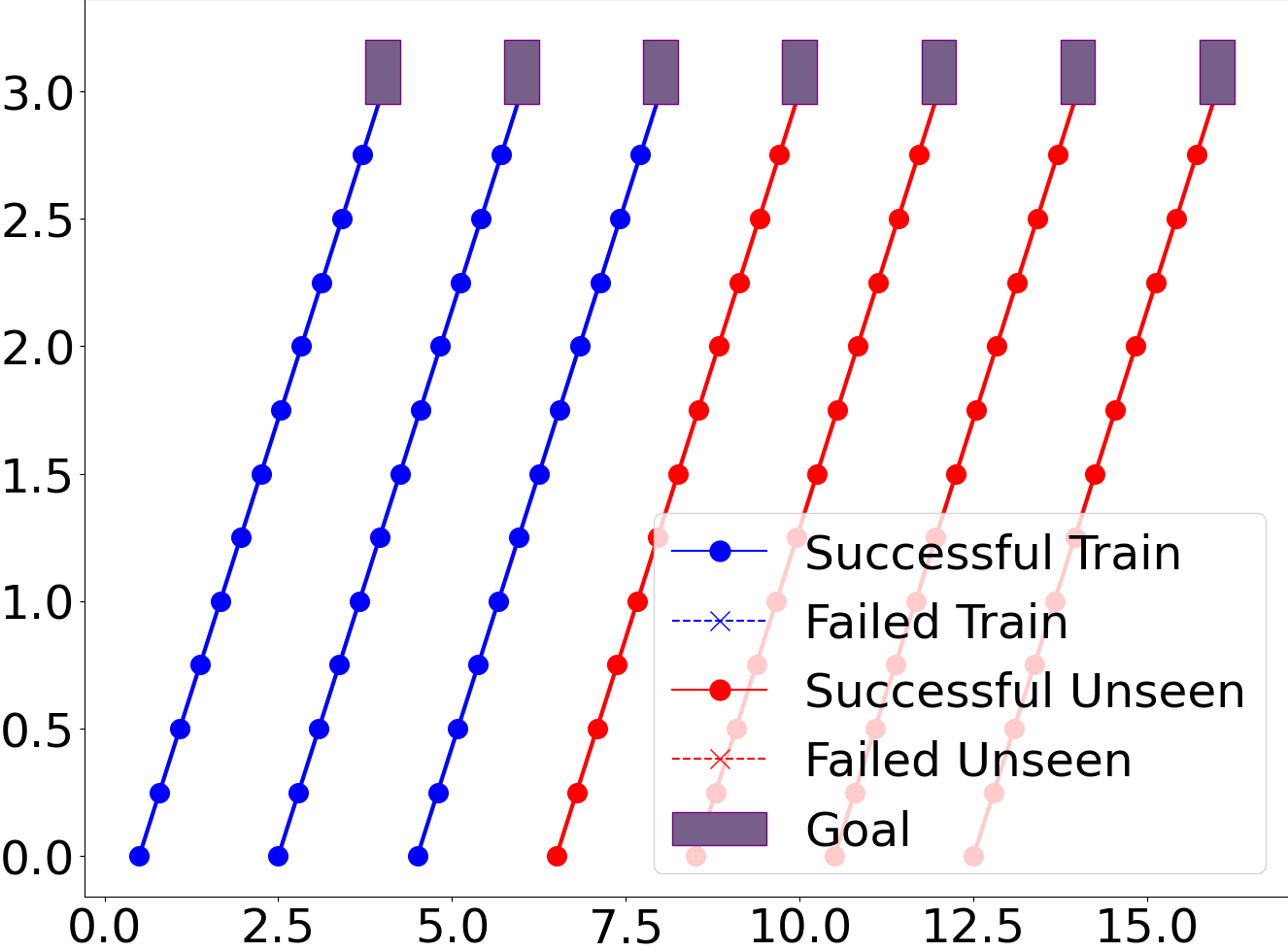}
        \caption{PSMP}
        \label{fig:1b}
    \end{subfigure}
    \caption{Moving initial and goal distributions}
    \label{fig:1reachmovpsmp}
\end{minipage}
\hfill
\begin{minipage}{0.3\textwidth} 
    \begin{subfigure}{\textwidth}
        
        \includegraphics[width=0.95\linewidth, height=0.7\linewidth]{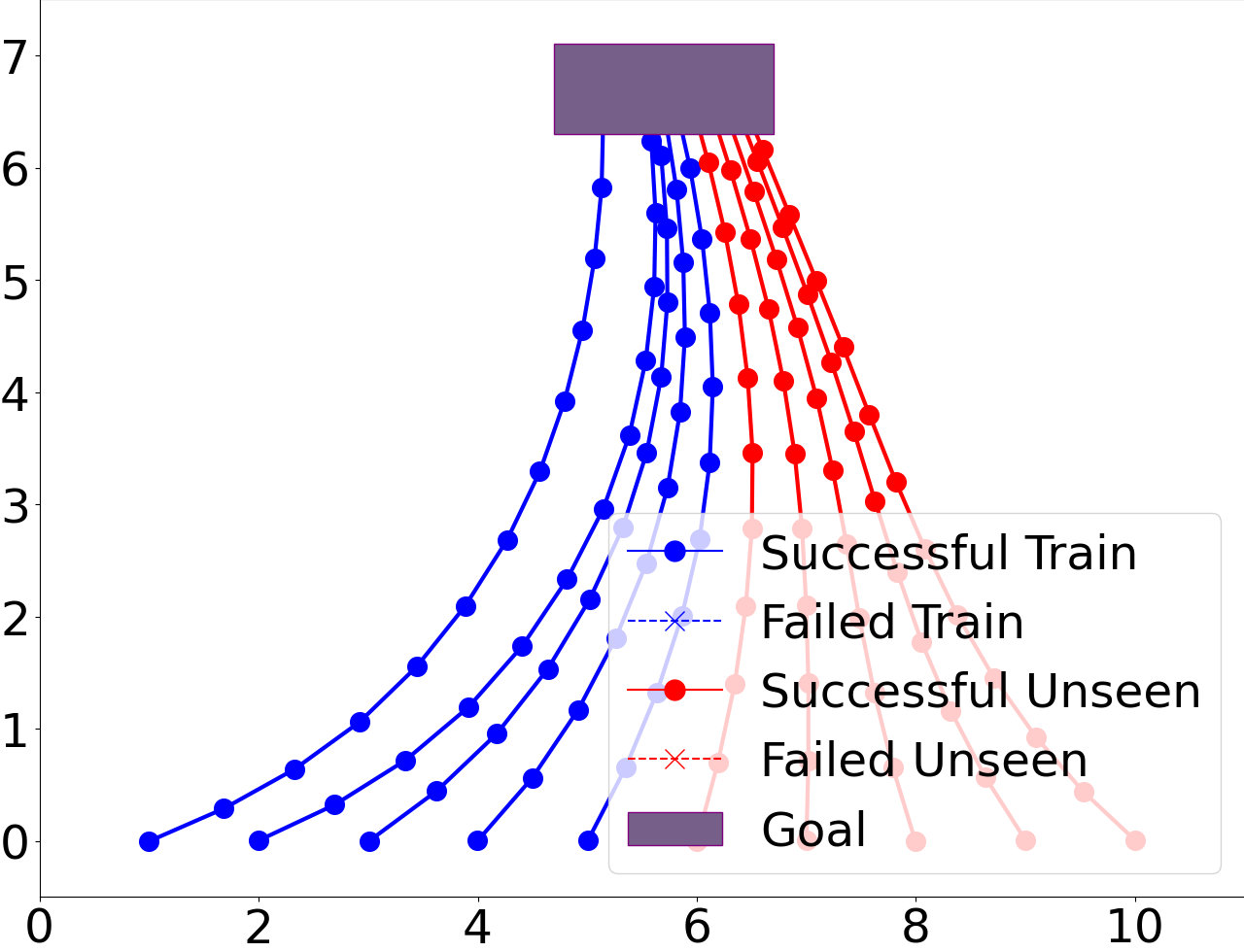}
        \caption{GenRL}
        \label{fig:2a}
    \end{subfigure}

    \medskip
    \begin{subfigure}{\textwidth}
        
        \includegraphics[width=0.95\linewidth, height=0.7\linewidth]{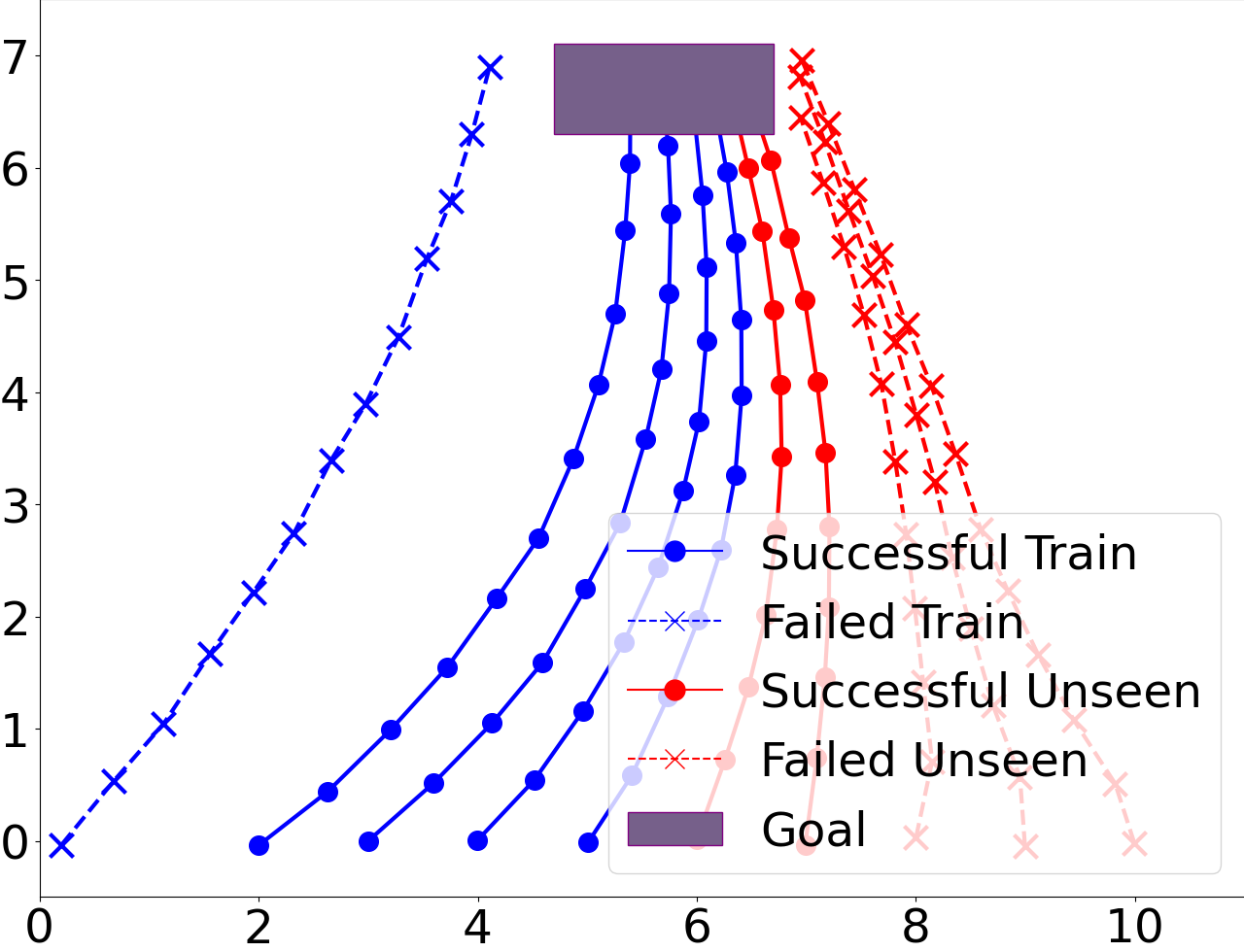}
        \caption{\textsc{Base2}}
        \label{fig:2b}
    \end{subfigure}
    \caption{Moving initial distribution, goal stationary}
    \label{fig:1reachpsmp}
\end{minipage}
\hfill
\begin{minipage}{0.3\textwidth} 
    \begin{subfigure}{\textwidth}
        
        \includegraphics[width=0.95\linewidth, height=0.7\linewidth]{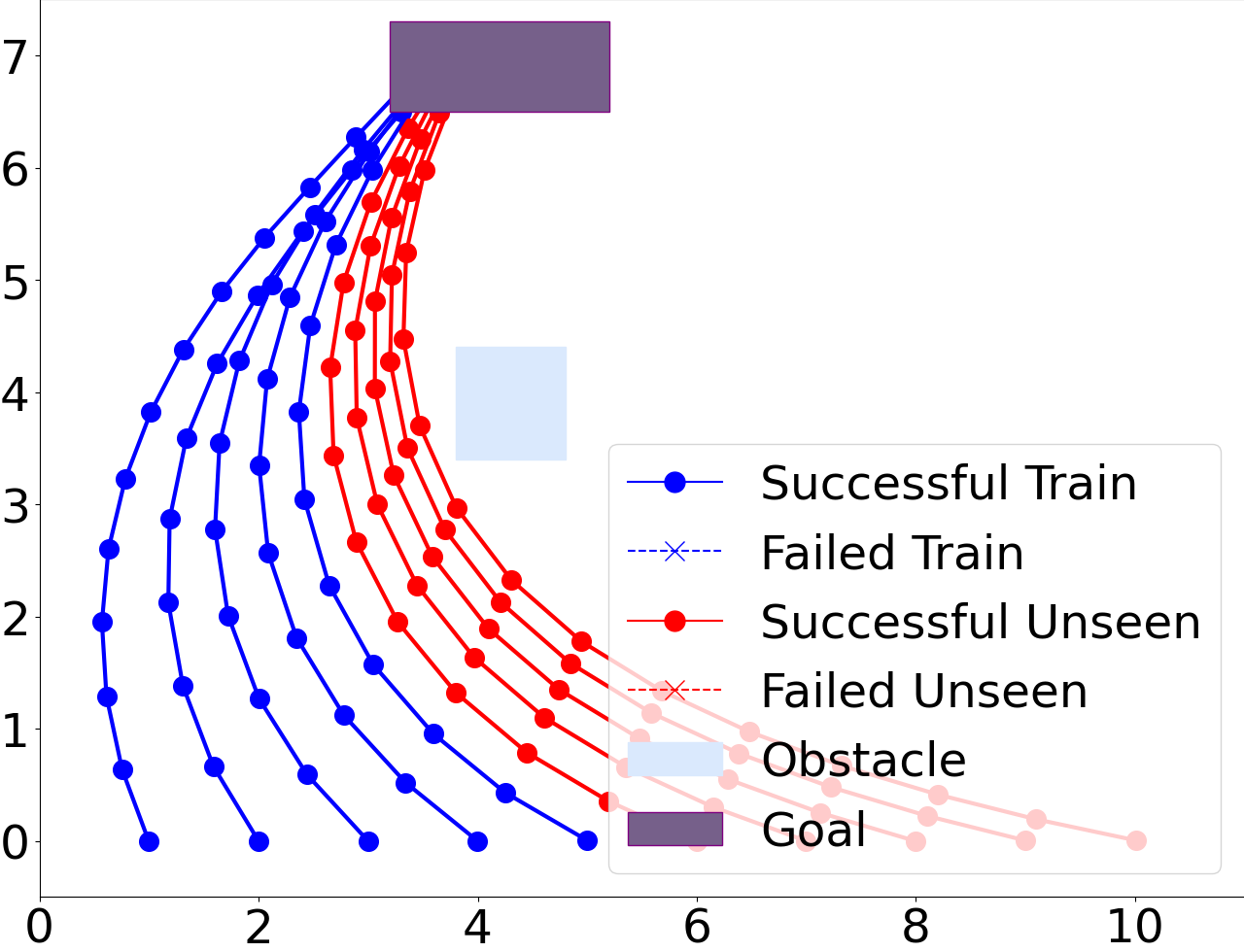}
        \caption{GenRL}
        \label{fig:3a}
    \end{subfigure}

    \medskip
    \begin{subfigure}{\textwidth}
        
        \includegraphics[width=0.95\linewidth, height=0.7\linewidth]{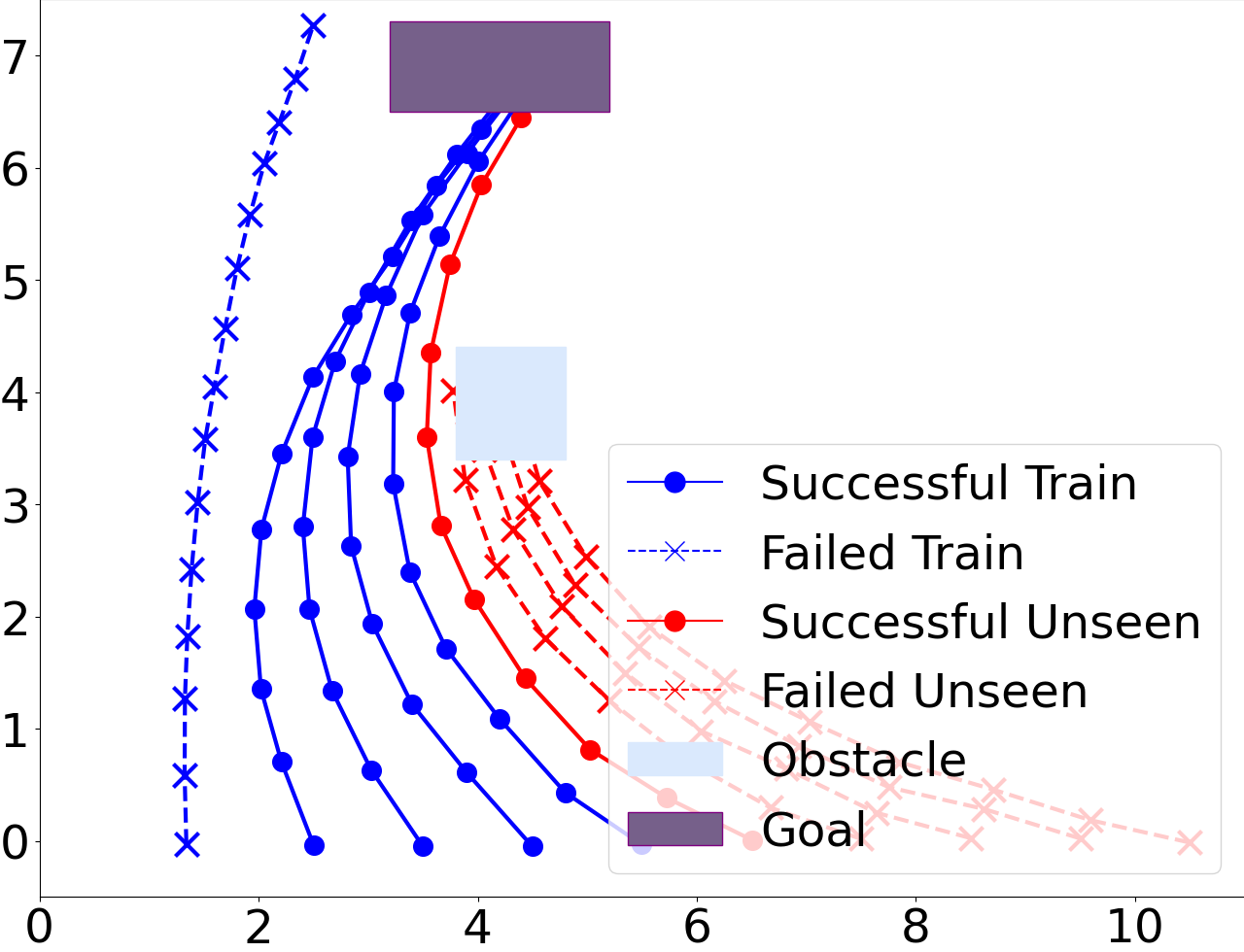}
        \caption{\textsc{Base2}}
        \label{fig:3b}
    \end{subfigure}
    \caption{Moving initial distribution with obstacle}
    \label{fig:1reachobspsmp}
    \label{fig:2}
\end{minipage}
\end{figure}

\paragraph{Simple reachability tasks.}

The inductive tasks in Figure~\ref{fig:1reachmovpsmp}, \ref{fig:1reachpsmp}, \ref{fig:1reachobspsmp} are to navigate a point object in the Cartesian plane from the initial distribution (blue and red dots on the $x$-axis) to their corresponding target position (grey box(es) connected to the initial dots by dashed lines). Base task instance is the leftmost task; other task instances are obtained by shifting either only the initial (Figures~\ref{fig:1reachpsmp}, \ref{fig:1reachobspsmp}) or both of the initial and goal positions (Figure~\ref{fig:1reachmovpsmp}) to the right by the same amount. Additionally, the point object must avoid the square obstacle in Figure~\ref{fig:1reachobspsmp}. In each of these figures, we compare GenRL with the best-performing baseline (PSMP in Figure~\ref{fig:1reachmovpsmp}, \textsc{Base2} in Figures~\ref{fig:1reachpsmp}, \ref{fig:1reachobspsmp}). We show trajectories of successful task instances by solid lines containing $\bullet$, unsuccessful ones as dotted lines with $\times$.   

Table~\ref{tab:1reachability_alltools} shows the number of task instances learned successfully from the training and testing sets. 
{The baselines do not even succeed on all task instances in $\traintasks$}. On the Unseen task instances, \genrl~learns the most in all but Figure~\ref{fig:1reachmovpsmp} where PSMP generalizes the most. 

We can conclude that \genrl~exhibits strong generalization as the policy generator discharges \textit{adapted} policies for each task instance.  The baselines generalize well on almost identical task instances (like Figure~\ref{fig:1reachmovpsmp}), but fail when a certain degree of adaptation is needed. The policies learned by PSMP and Base2 produce almost similar trajectories for all task instances; on the other hand, \genrl~is able to adapt its policies to trajectories that are related-but-different.

\paragraph{Complex reachability tasks with choice.} 

\begin{figure*}[t]
\centering
\begin{minipage}[t]{0.27\textwidth}
\centering
\includegraphics[width=\textwidth]{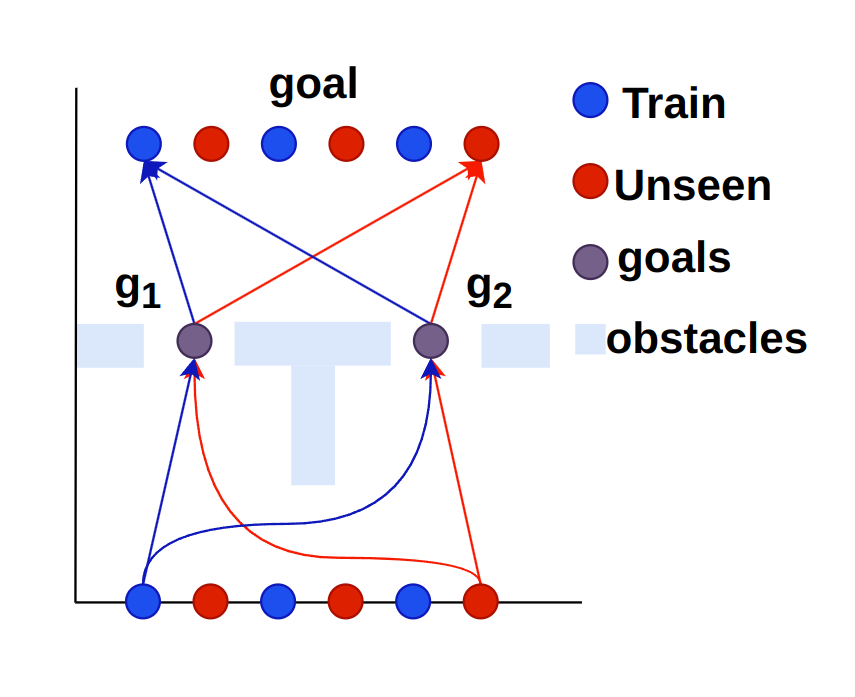}
\caption{Choice with moving goal}
\label{fig:choice2illus}
\end{minipage}
\hspace{0.1cm}
\begin{minipage}[t]{0.27\textwidth}
\centering
\includegraphics[width=\textwidth]{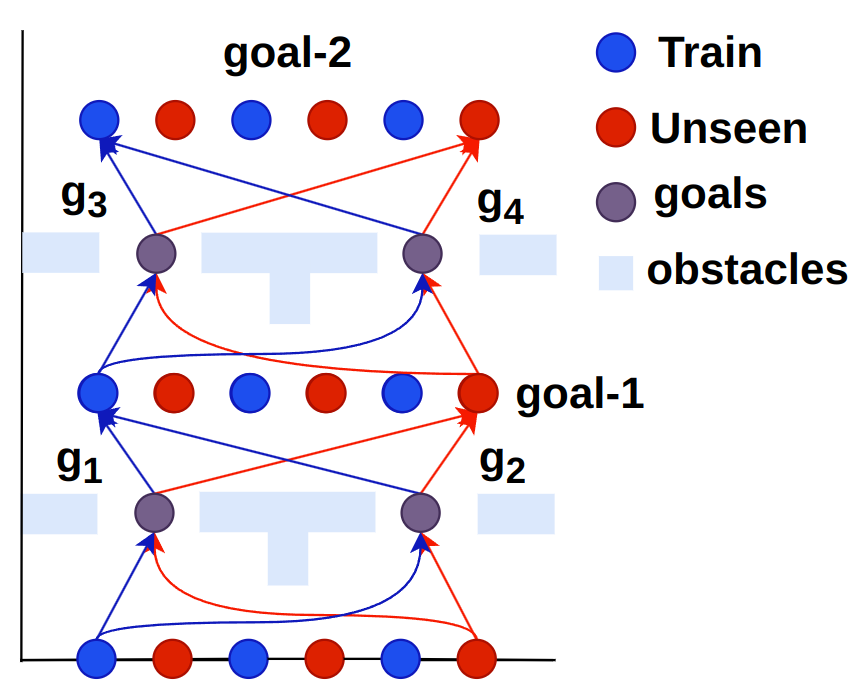}
\caption{Two levels of choice with moving goal}
\label{fig:choice3illus}
\end{minipage}
\hspace{0.1cm}
\begin{minipage}[h]{0.4\textwidth}
\centering
\setlength{\tabcolsep}{1pt}
\small
\vspace{-40pt}
\begin{tabular}{|cc|c|c|c|}
\hline
\multicolumn{2}{|c|}{\textbf{Benchmark}} & \multirow{2}{*}{\begin{tabular}[c]{@{}c@{}}\textbf{Successful}\\ \textbf{Train} \end{tabular}} & \multirow{2}{*}{\begin{tabular}[c]{@{}c@{}}\textbf{Successful}\\ \textbf{Unseen} \end{tabular}} & \multirow{2}{*}{\begin{tabular}[c]{@{}c@{}}\textbf{Guard}\\ \textbf{predicate}\end{tabular}} \\
\multicolumn{2}{|c|}{$|\traintasks| = 6$} &  &  &  \\ 
\hline
\multicolumn{2}{|c|}{Figure~\ref{fig:motivating_example}} &  \textbf{All} & 7 & $(i \leq 4)$ \\ 
\multicolumn{2}{|c|}{Figure~\ref{fig:choice2illus}} & \textbf{All} & 5 & $(i\leq 4)$ \\ 
\multicolumn{2}{|c|}{Figure~\ref{fig:choice3illus}} & \textbf{All} & 5 & $(i\leq 4)$, \\ 
\multicolumn{2}{|c|}{} &  &  & $(i\leq 4)$ \\ 
\hline
\end{tabular}
\captionof{table}{No. of successful task instances on $\traintasks$ and Unseen on Choice} 
\label{table:car2dchoice:merged}
\end{minipage}
\end{figure*}

In more complex long-horizon tasks, we examine Choice from Figure~\ref{fig:motivating_example}, and its variations Figure~\ref{fig:choice2illus} where the final goal position is vertically above the initial position (hence moving to the right) and Figure~\ref{fig:choice3illus} which stacks Figure~\ref{fig:choice2illus} on top of itself, hence two moving goals. 
See Appendix~\ref{sec:ap:environmentcar} for full descriptions and more experiments on multi-goal reachability.

These tasks are significantly harder than the previous ones as the policy generator also learns guard conditions to determine intermediate goal positions ($g_1$, $g_2$, $g_3$, or $g_4$) for every task instance. Since the baselines failed on the simpler tasks, Table~\ref{table:car2dchoice:merged} presents \genrl~only. \genrl~successfully learns all in \traintasks~ and shows high generalization on unseen tasks as well. We observe that the policy generator always chooses the closer intermediate goal position on every task instance.

\subsection{Long-horizon tasks in environments with complex dynamics}
\label{sec:longhorizoncompdyn}

\paragraph{Tower-destacking with robotic arm.}

\begin{figure*}[t]
    \centering
    \begin{minipage}[t]{0.23\textwidth}
        \begin{subfigure}[t]{\textwidth}
            \includegraphics[width=\textwidth]{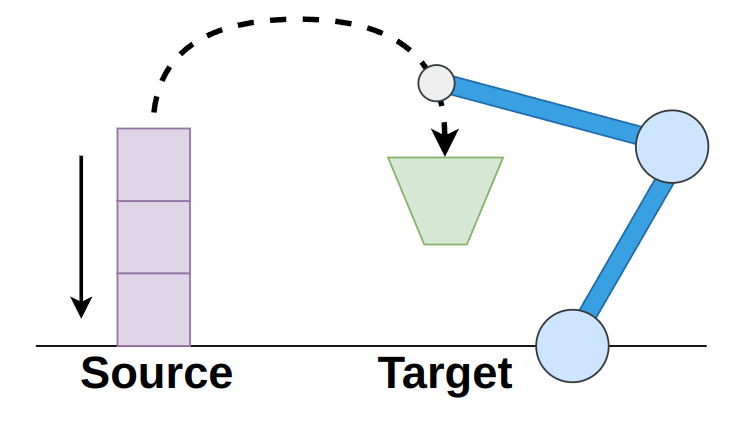}

        \end{subfigure}
        \vfill
        \begin{subfigure}[t]{\textwidth}
            \includegraphics[width=\textwidth]{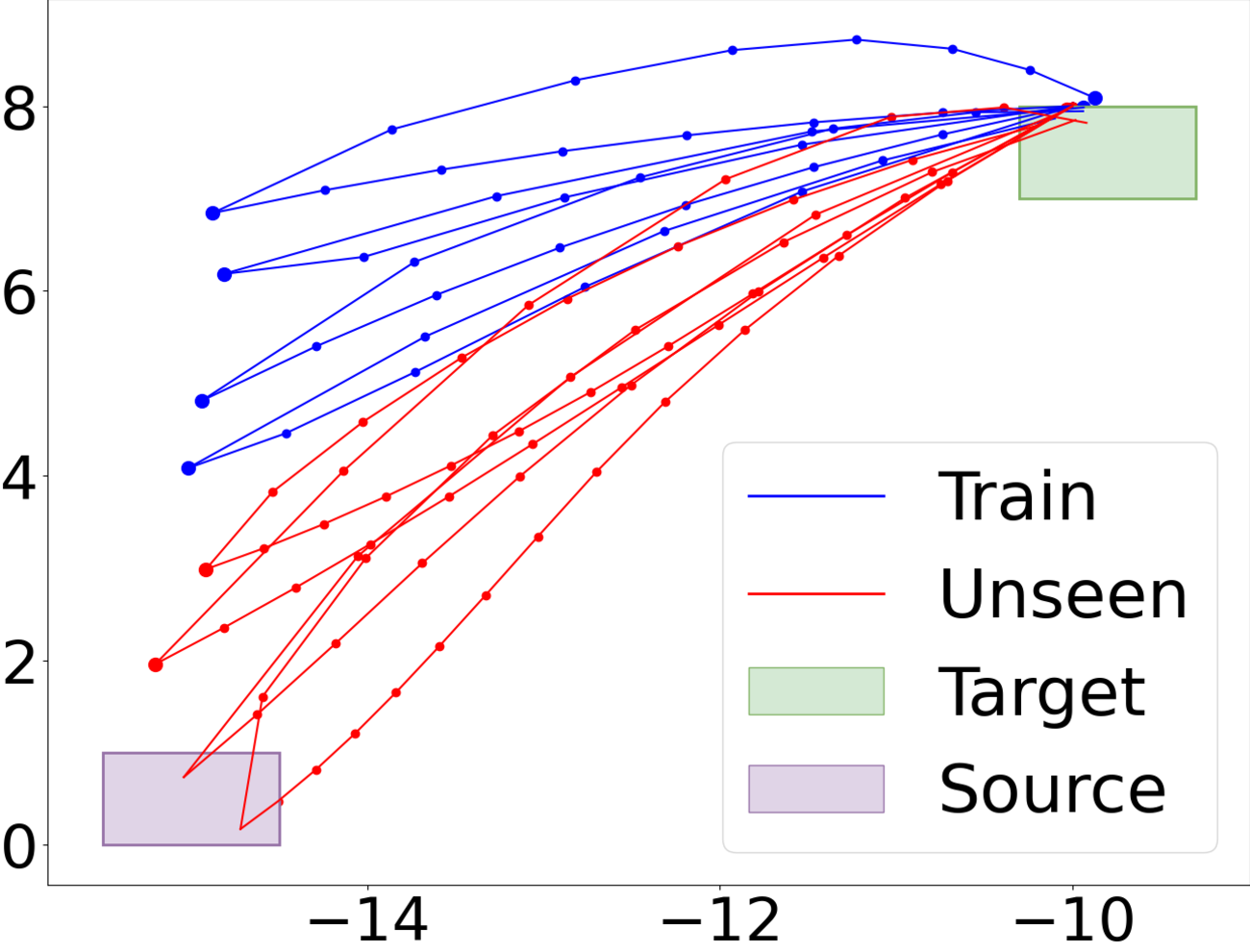}

        \end{subfigure}
        \caption{Pick and Drop: Same side}
        \label{fig:r3}
    \end{minipage}
    \hfill
    \begin{minipage}[t]{0.23\textwidth}
        \begin{subfigure}[t]{\textwidth}
            \includegraphics[width=\textwidth]{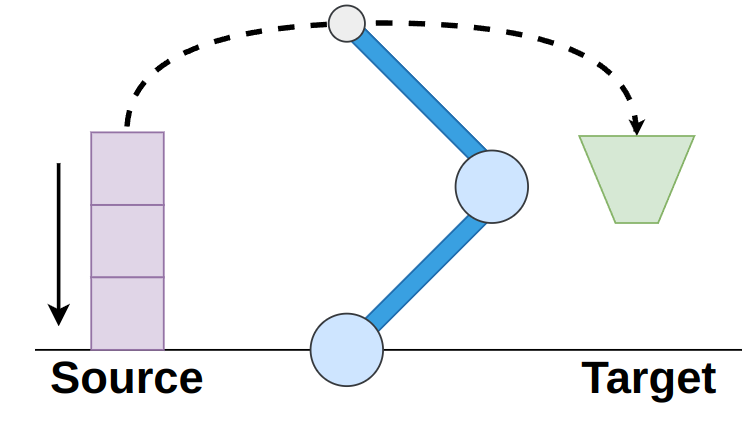} 

        \end{subfigure}
        \vfill
        \begin{subfigure}[t]{\textwidth}
            \includegraphics[width=\textwidth]{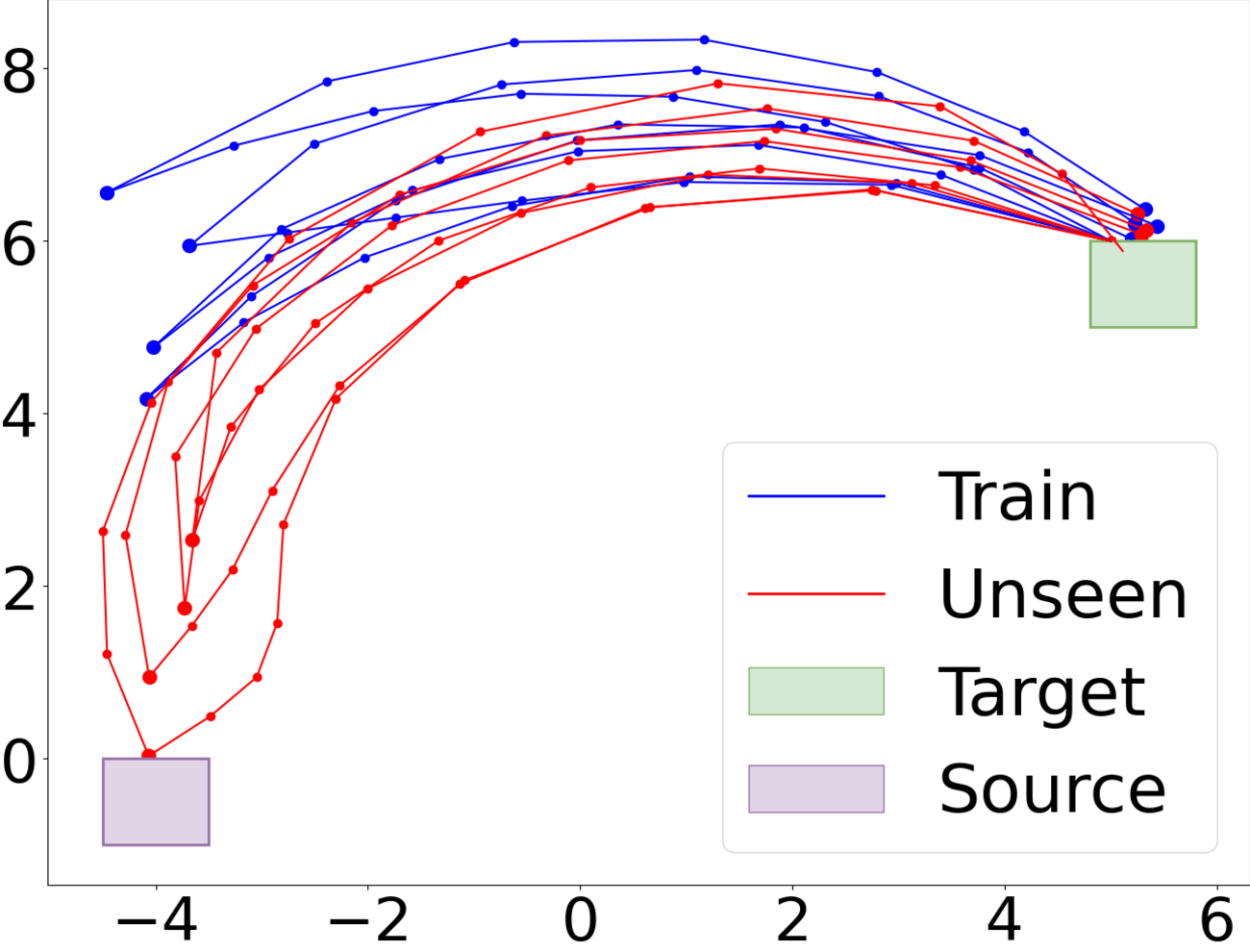}

        \end{subfigure}
        \caption{Pick and Drop: Opposite sides}
        \label{fig:r1}
    \end{minipage}
    \hfill
    \begin{minipage}[t]{0.23\textwidth}
        \begin{subfigure}[t]{\textwidth}
            \includegraphics[width=\textwidth]{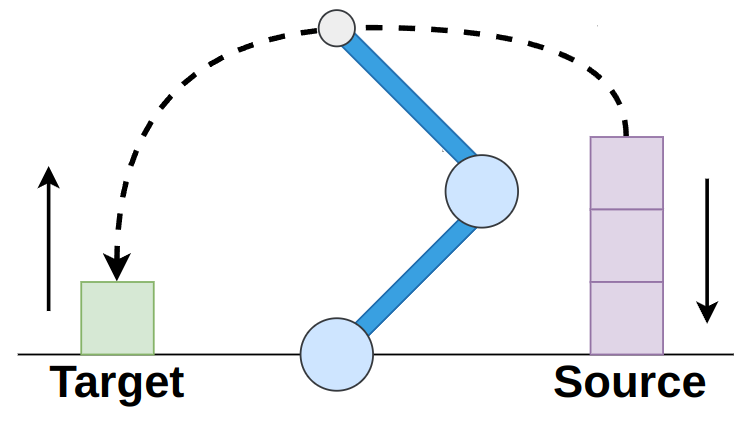}

        \end{subfigure}
        \vfill
        \begin{subfigure}[t]{\textwidth}
            \includegraphics[width=\textwidth]{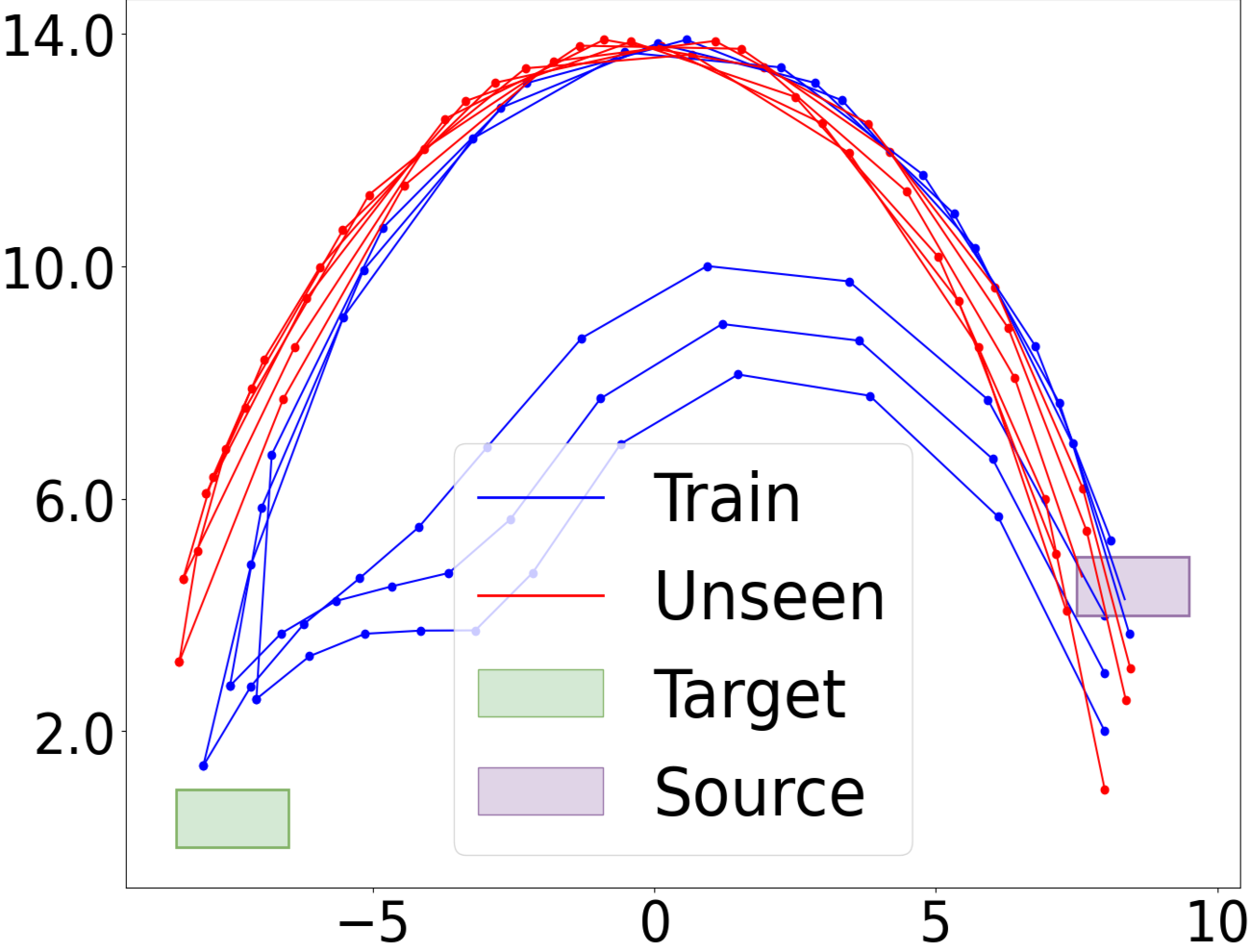}

        \end{subfigure}
        \caption{Pick and Vert. Stack: Same Side}
        \label{fig:r2}
    \end{minipage}
    \hfill
    \begin{minipage}[t]{0.23\textwidth}
        \begin{subfigure}[t]{\textwidth}
            \includegraphics[width=\textwidth]{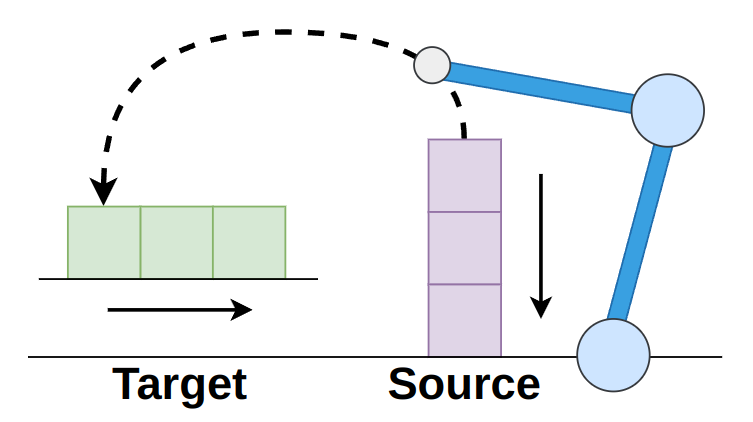}

        \end{subfigure}
        \vfill
        \begin{subfigure}[t]{\textwidth}
            \includegraphics[width=\textwidth]{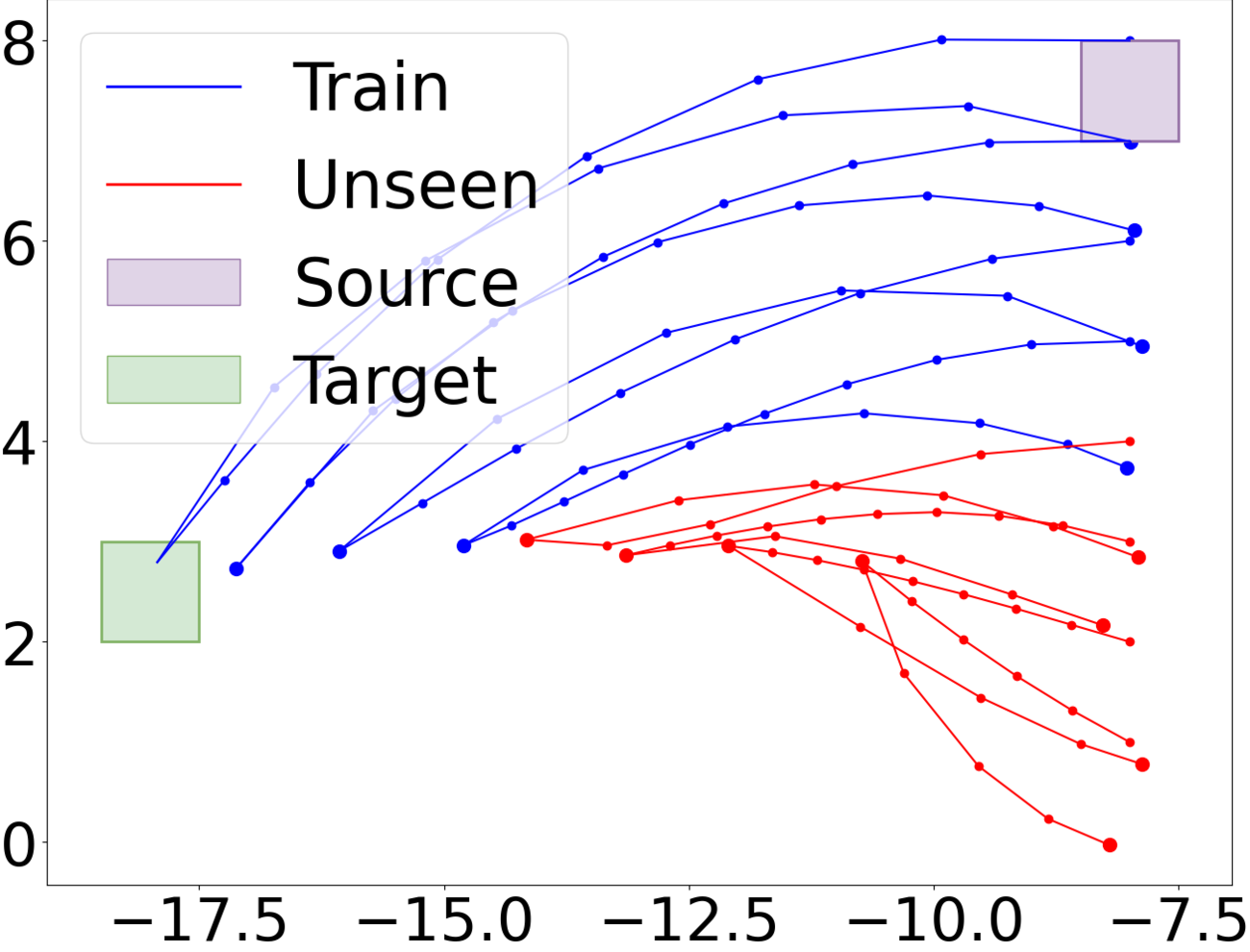}

        \end{subfigure}
        \caption{Pick and Hor. Stack: Same side}
        \label{fig:r5}
    \end{minipage}
\end{figure*}

We examine variations of Figure~\ref{fig:motivating_example_reacher} (stacking on the same side), like destacking the tower into a dropbox on same side (Figure~\ref{fig:r3}) and opposite side (Figure~\ref{fig:r1}); stacking on opposite sides (Figure~\ref{fig:motivating_example_reacher}); horizontal stacking (Figure~\ref{fig:r5}). In all cases, each task instance displaces the topmost block from the source tower. Formal descriptions can be found in Appendix~\ref{sec:reacherenv}.

\begin{figure*}[t]
    \begin{minipage}[t]{0.48\linewidth}
    \centering
    \includegraphics[width=\linewidth]{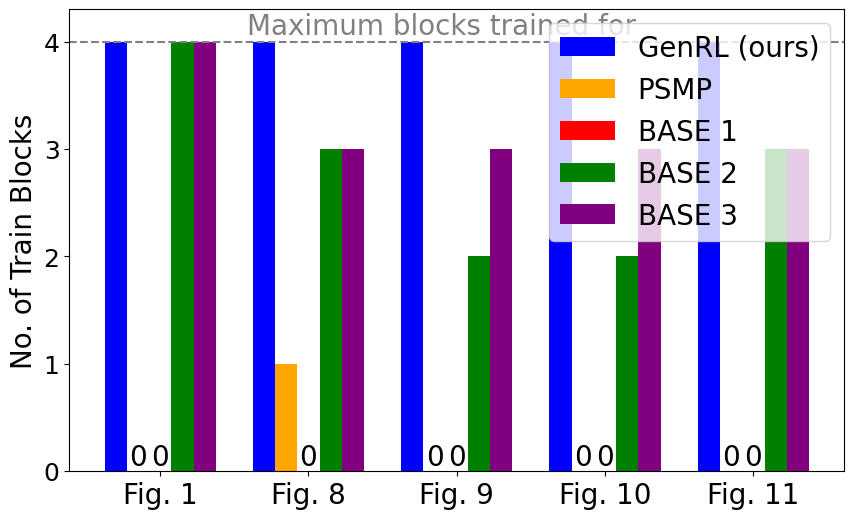}
    \caption{Number of successful task instances on Train on Reacher Benchmarks}
    \label{fig:reachertrain}
    \end{minipage}
    \hfill
    \begin{minipage}[t]{0.48\linewidth}
        \centering
        \includegraphics[width=\linewidth]{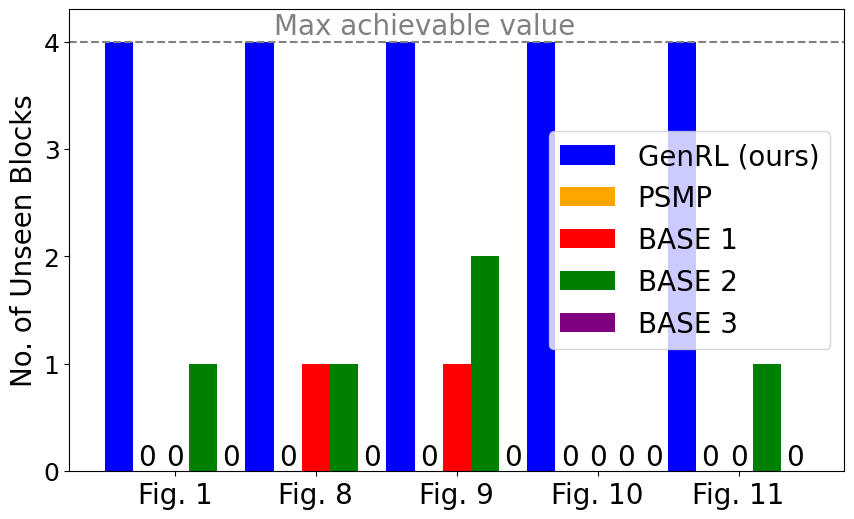}
        \caption{Number of successful task instances on Unseen on Reacher Benchmarks}
        \label{fig:armexphist}
    \end{minipage}   
\end{figure*}

In our experiments, the source tower is initially eight blocks high. We train on the top four blocks of the source tower and test on its remaining four blocks.  In all experiments, \genrl~successfully displaces all training blocks.  
Furthermore, Figure~\ref{fig:reachertrain} and Figure~\ref{fig:armexphist} show that on \genrl~could learn to displace all the training and unseen blocks in the experiments, respsectively. It is, infact, the only tool to do so. Figures~\ref{fig:r3}-\ref{fig:r5} show the trajectories produced by \genrl.

\paragraph{Classical control benchmarks.}

\begin{figure}
\centering
\includegraphics[width=0.5\linewidth]{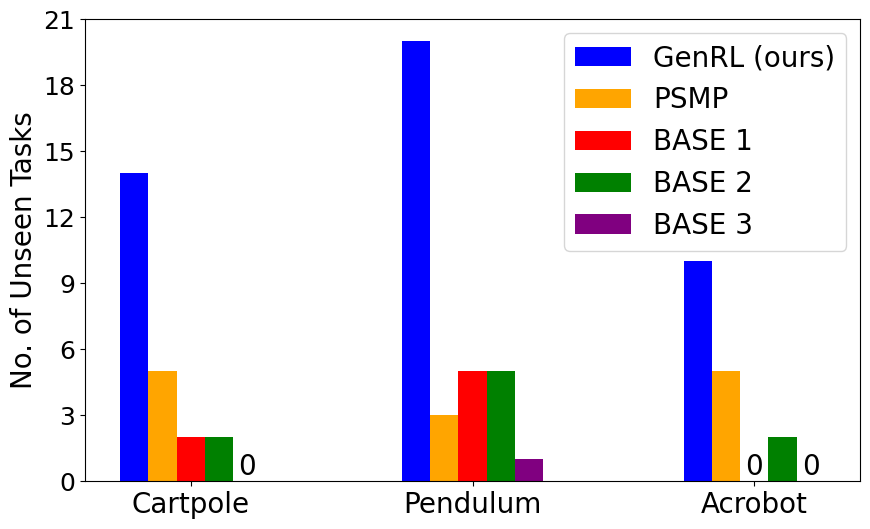}
\caption{Classical control benchmarks on Unseen}
\label{fig:gymevnhist} 
\end{figure}

We examine $\genrl$ on classical RL control benchmarks~\cite{openaigym}. 
The inductive tasks are designed such that the induction occurs on the mass of the object in the Pendulum and Acrobot environments and on the length of the pole in the Cartpole environment. For details, refer to Appendix~\ref{sec:gymexp}. As earlier, $\genrl$ learns the most number of successful task instances (see Figure~\ref{fig:gymevnhist}). This demonstrates our framework's ability to generalize even when the induction is on an environment-parameter as opposed to a specification-parameter.

\section{Future Works}
\label{sec:limitations}
Currently, $\genrl$ performs effectively in environments with lower-dimensional action and state spaces. However, its scalability to more complex environments with higher-dimensional spaces remains a challenge. Future work will focus on enhancing the algorithm's capability to handle these more complex scenarios. Additionally, while defining tasks via logical specifications is generally easier than specifying rewards, it still requires considerable effort to design these specifications accurately. To address these, future research will aim at developing more streamlined and user-friendly methods for task specification to make the specification process as lightweight as possible. This will help broaden the applicability of $\genrl$ or any specification-guided learning to a wider range of tasks and environments. Even though the logic and formulation behind our research are principally motivated by a foundational hypothesis and empirically validated for its performance and results, we still not have investigated the possibilities of providing theoretical guarantees and this is also something our future work will focus on.

\section{Societal Impacts}
\label{sec:impacts}
Our work focuses on improving the generalization capabilities in reinforcement learning. This could have a substantial impact in the automation and robotics domain as the ability to learn generalized policies for new tasks can allow agents to adapt to new environments and tasks with minimal retraining, therefore reducing the compute requirements and costs involved.

A potential negative impact of our work is the military applications which is also a common concern with robotics-related research.

\medskip

\bibliography{reference}
\bibliographystyle{neurips2024}

\newpage
\appendix
\onecolumn

\section*{Appendix}

\section{Supplemental Material}
\label{sec:supplemental}

The complete source code of $\genrl$ tool along with our experimental setup has been made available at \href{https://anonymous.4open.science/r/GenRL_Zenith-7EEB/}{https://anonymous.4open.science/r/GenRL\_Zenith-7EEB/}. We provide comprehensive training and testing code for our experiments. In addition to the training and testing scripts, we include pre-trained models that allow users to generate rollouts and visualize the resulting trajectories. We have provided a \texttt{requirements.txt} within the artifact with detailed instructions. 

To get started, please install the required dependencies: \\
\texttt{pip install -r requirements.txt}. 

\paragraph{Training new models.}
To train the models, use the following command with the appropriate parameters: \\
\texttt{python -u -m spectrl.examples.car2d\_dirl -n 0 -d car2d\_k2/\{any\_name\_of\_your\_choice\} -s \{s\} -h \{h\} -b 'j2\_'}. 

\begin{itemize}
    \item For Towerstack with robotic arm experiments, set \texttt{s} from 0 to 5 and \texttt{h} to 8. 
    \item For choice experiments, set \texttt{s} to 6 or 7 and \texttt{h} to 10. 
    \item For classical control experiments, set \texttt{s} from 8 to 10 and \texttt{h} to 10. 
\end{itemize}

Running the training script also executes the testing script. The codebase is set to test for only \texttt{n} number of unseen tasks.  To test for a different number of task instances, change the \texttt{test\_rounds} variable in the \texttt{spectrl/hierarchy/reachability.py} file. Replace \texttt{\{any\_name\_of\_your\_choice\}}, \texttt{\{s\}}, and \texttt{\{h\}} with your specific values. For example, to run a tower-stack with a robotic arm experiment with \texttt{s} set to 2 and \texttt{h} set to 8, use: \\
\texttt{python -u -m spectrl.examples.car2d\_dirl -n 0 -d car2d\_k2/experiment1 -s 2 -h 8 -b 'j2\_'}. 

\paragraph{Only visualizing rollouts from pre-trained models.}
If you only want to visualize the rollouts from the pre-trained models and not train new ones, modify the parameters in the \texttt{spectrl/examples/car2d\_dirl.py} file: set \texttt{training} to \texttt{False} (line 521) and set \texttt{prepare\_rollouts} to \texttt{True} (line 522), and run the same training commands given above as per your requirements by choosing the right file name. For example, \texttt{python -u -m spectrl.examples.car2d\_dirl -n 0 -d car2d\_k2/choice\_test -s 6 -h 10 -b 'j2\_'}.
\section{\spectrl}
\label{ap:spectrlsemantics}

\paragraph{\spectrl semantics.}
Letting $\zeta$ be a finite trajectory of length $t$, this function is defined by
\begin{align*}
\zeta&\models\eventually{b} ~&&\text{if}~ \exists\ i \leq t,~s_i\models b \\
\zeta&\models \p \always{b} ~&&\text{if}~ \zeta\models\p ~\text{and}~ \forall\ i\leq t, ~ s_i\models b \\
\zeta&\models\p_1; \p_2 ~&&\text{if}~ \exists\ i < t, ~\zeta_{0:i}\models \p_1 ~\text{and}~ \zeta_{i+1:t}\models\p_2 \\
\zeta&\models\choice{\p_1}{\p_2} ~&&\text{if}~ \zeta\models\p_1 ~\text{or}~ \zeta\models\p_2.
\end{align*}

\section{Abstract Graph}

\begin{definition}[\cite{jothimurugan2021compositional}]
\rm
An {\em abstract graph} $\G = (U,E,u_0,F,\beta,\traj_{\safe})$ is a directed acyclic graph (DAG) with vertices $U$,
(directed) edges $E\subseteq U\times U$, initial vertex $u_0\in U$, final vertices $F\subseteq U$, subgoal region map $\beta:U\to2^S$ such that for each $u\in U$, $\beta(u)$ is a subgoal region, and \emph{safe trajectories}
$
\traj_\safe = \bigcup_{e \in E}\traj_\safe^e\cup\bigcup_{f \in F}\traj_\safe^f,
$
where $\traj_\safe^e\subseteq\traj$ denotes the safe trajectories for edge $e \in E$ and $\traj_\safe^f\subseteq\traj$ denotes the safe trajectories for final vertex $f\in F$.

Intuitively, $(U,E)$ is a DAG. Furthermore, $\beta$ and $\traj_{\safe}$ connect $(U,E)$ back to the original MDP $\M$; in particular, for an edge $e=u\to u'$, $\traj_{\safe}^e$ is the set of safe trajectories in $\M$ that can be used to transition from $\beta(u)$ to $\beta(u')$.

A trajectory $\zeta=s_0\xrightarrow{a_0}s_1\xrightarrow{a_1}\cdots\xrightarrow{a_{t-1}}s_t$ in $\M$ satisfies the abstract graph $\G$ (denoted $\zeta\models \G$) if there is a sequence of indices $0=k_0\leq k_1<\cdots<k_\ell\leq t$ and a path $\rho=u_0\to u_1\to\cdots\to u_\ell$ in $\G$ such that
(a). $u_\ell\in F$,
(b). for all $z\in\{0,\ldots,\ell\}$, we have $s_{k_z}\in \beta(u_z)$, 
(c). for all $z < \ell$, letting $e_z=u_z\to u_{z+1}$, we have $\zeta_{k_z:k_{z+1}}\in\traj_{\safe}^{e_z}$, and
(d). $\zeta_{k_\ell:t}\in\traj_\safe^{u_\ell}$.
The first two conditions state that the trajectory should visit a sequence of subgoal regions corresponding to a path from the initial vertex to some final vertex, and the last two conditions state that the trajectory is composed of subtrajectories that are safe according to $\traj_\safe$.

An {\em edge policy} $\pi_e$ for a single edge $e=u\to u'$
is one that safely transitions from a state in $\beta(u)$ to a state in $\beta(u')$. Then, 
given edge policies $\Pi$ along with a path
$
\rho=u_0\to u_1\to \cdots\to u_k = u
$
in $\G$, the \emph{path policy} ${\pi}_{\rho}$ navigates from $\beta(u_0)$ to $\beta(u)$. In particular, ${\pi}_{\rho}$ executes $\pi_{u_j\to u_{j+1}}$ (starting from $j=0$) until reaching $\beta(u_{j+1})$, after which it increments $j\gets j+1$ (unless $j=k$). That is, ${\pi}_{\rho}$ is designed to achieve the sequence of edges in $\rho$. 
Then, an optimal policy for \spectrl specification is the optimal path policy from the initial to the final state in the abstract graph of the specification.
\end{definition}

\section{Algorithm Details}
\label{sec:algoappen}

\subsection{Kappa Learning Algorithm (Algorithm~\ref{algo:kappatraining} \trainkappa )}
\label{ap-sec:algo:kappalearning}

Algorithm~\ref{algo:kappatraining} \trainkappa~  learns the kappa-coefficients using an ARS style algorithm with a softmin aggregator. 

In more detail, vector $\kappa = [\kappa_0, \kappa_1, \ldots, \kappa_{m-1}]$ is initialized as a normal distribution vector, where $m$ represents the number of elements in $\kappa$. $\sampledelta$ samples perturbation vectors $\delta$, conforming to the same dimensionality as $\kappa$. $\perturbkappa$ creates perturbed kappa vectors $\kplus = \kappa + (\delta_{scale} \cdot \delta)$ and $\kminus = \kappa - (\delta_{scale} \cdot \delta)$. For each perturbed kappa vector, the equation~\ref{eq:kappa-polynomail} ($\kpolicy$) generates policies by polynomially combining its elements with the base policy $\pi^0_e$. These generated policies are evaluated for each task $\task_i \in \traintasks$, with rewards accumulated in $\rplus$ and $\rminus$ for policies derived from $\kplus$ and $\kminus$ for each task, respectively. $\Score$ aggregates these rewards into collective performance scores $\Rplus$ and $\Rminus$ by computing $\Rplus \gets \mathit{softmin}(\rplus)$ and similarly for $\Rminus$. It then forms tuples $\delta_{samples}$, pairing each perturbation $\delta$ with its corresponding aggregate scores. $\ARSupdate$ computes a weighted average perturbation $\delta_{\kappa}$ from these samples, guiding the update of the kappa vector as $\kappa_{updated} \gets \kappa + \delta_{\kappa}$. This iterative process of sampling, evaluating, and updating is continued until convergence of the kappa vector is reached, optimizing the policies for the specified tasks in $\traintasks$. The reward function $\Reward$ based on the Euclidean distance between the agent's position and the goal position can be expressed as $\Reward = -\|\mathbf{p}_{\text{agent}} - \mathbf{p}_{\text{goal}}\|$ where $\mathbf{p}_{\text{agent}}$ represents the position vector of the agent, $\mathbf{p}_{\text{goal}}$ represents the position vector of the goal, $\|\cdot\|$ denotes the Euclidean norm (or Euclidean distance).

\begin{algorithm}[t]
\caption{{\trainkappa}($e, m, \pi_{e}^0, \Gamma_{u}, \traintasks$) \\ Kappa Training using a modified Augmented Random Search}
\begin{algorithmic}[1]
\STATE Initialize $\kappa(m) $\\ where $m \gets$ number of kappa in the polynomial template

\WHILE{$\kappa$ not converged}

    \STATE $\delta_{samples} \gets \emptyset$ 
    
    \FOR{$s =  0$ to $n\_samples$}
    
        \STATE $\rplus \gets \emptyset$, $\rminus \gets \emptyset$
        \STATE $\delta \gets \sampledelta(\kappa)$
        \STATE $\kplus \hp{2} \gets \perturbkappa(\kappa, \delta, \deltas)$
        \STATE $\kminus \gets \perturbkappa(\kappa, \delta, -\deltas)$  
        
        \FOR{$k = 0$ to $|\traintasks|$ and task $\task_i$ where $i \in \traintasks$}
        
            \STATE $\pplus \hp{4} \gets \kpolicy(\kplus,\pi^0_e, k, m)$
            \STATE $\rplus[k] \hp{2} \gets \Reward(\pplus, \task_i)$
            \STATE $\pminus \hp{2} \gets \kpolicy(\kminus, \pi^0_e,k, m)$
            \STATE $\rminus[k] \gets \Reward(\pminus, \task_i)$
            
        \ENDFOR
        
        \STATE $\Rplus \hp{2} \gets \Score(\rplus)$
        \STATE $\Rminus \gets \Score(\rminus)$
        \STATE $\delta_{samples}[s] \gets (\delta, \Rplus, \Rminus)$
        
    \ENDFOR
    
    \STATE $\delta_{\kappa} \gets  \ARSupdate(\delta_{samples})$
    \STATE Update $\kappa \gets \perturbkappa(\kappa, \delta_{\kappa}, 1)$
    
\ENDWHILE 
\STATE \textbf{return} $\kappa$
\end{algorithmic}
\label{algo:kappatraining}
\end{algorithm}

\subsection{Learning Guard Conditions}
\label{sec:algo:decisionboundary}

\paragraph{Overview}
Post-DAG traversal, Algorithm~\ref{algo:decisionboundary} conducts a reverse traversal to ascertain optimal tasks $\task_i$ for each edge $e$, with $i \in \traintasks$, guiding the task from the initial state $u_0$ to any one of the final states in $F$. The identified tasks for each task are stored in $\De$ where edge $e \in \G$. Subsequently, a dataset $D$ is generated, on which a decision tree classifier is trained to yield a $\boundary$. $\boundary$ directs the choice of edges for $\task$, such that the likelihood of reaching final state $f \in F$ with maximal success likelihood.

$\De(e):$ Given a set of task indices $i \in \traintasks$ and $i \in \De(e)$, the most optimal path from the initial state to the final state for the task $\task_i$ goes through the edge $e$. 

\paragraph{Details}

The Algorithm~\ref{algo:decisionboundary} commences by initializing the outgoing edges $O(v)$ for every vertex $v \in \G$ and populating a queue $Q$ with the final states $F$. In its reverse traversal phase, the algorithm, for each vertex $u$ dequeued from $Q$, examines the incoming vertices. For every incoming vertex $v$, it removes $u$ from $v$'s outgoing edges and for every task $i \in \traintasks$, it checks whether the vertex $v$ is in $\bestincomingedge(u, i)$  and if this condition is true, then the task index $i$ is appended to $De(e)$ where edge $e = (v \rightarrow u)$. Then, if \(O(v)\) becomes empty, meaning \(v\) has no more outgoing vertices to process, \(v\) is enqueued in \(Q\) for further processing. This iterative process continues until all vertices in $\G$ are traversed, ultimately determining the $\De)$ for each edge, which identifies the optimal tasks for which each edge forms a part of the best path to any of the final states $F$ from the initial state $u_0$.  

Now, using $\De$, we establish decision boundaries for vertices $u$ in $\G$ where $\outgoing(u) > 1$, identifying vertices necessitating decision-making (choose the optimal outgoing edge for a particular task). For each such vertex $u$, the algorithm examines every edge $e = (u \rightarrow v)$, with $v \in \outgoing(u)$. In this process, the algorithm iterates over each task $\task$ in $\De$ to create a dataset to train our decision tree classifier:

(Let dataset $D = \{(x, y) \mid x \in X, y \in Y\}$ denote the dataset, where $X$ is the set of input feature vectors and $Y$ is the set of output target labels. Each pair $(x, y)$ in $D$ corresponds to a feature vector $x$ from $X$ and its associated label $y$ from $Y$)

\begin{itemize}
    \item If all tasks $\task$ are common across all outgoing edges from $u$ to $v$, the feature set $X$ for the vertex $u$ is formed using the environmental input values $\EnvInputValues(\task)$ (here, Cartesian coordinates of the task's initial distribution points), and the target label $Y$ is set as the first outgoing edge for all tasks $\task$ for all $\De$.
    \item In cases where tasks $\task$ are not common to all edges, and a specific task appears in multiple but not all edges, the algorithm includes in the dataset for this task the environmental input values $X \gets \EnvInputValues(\task)$ and target label $Y \gets \text{ edge }e$ for the first edge it appears in while ignoring its appearances in subsequent edges.
    \item For tasks unique to an edge, the dataset is constructed such that $X \gets \EnvInputValues(\task)$ and the target label $Y \gets \text{ edge } e$, where the task $\task$ appears.
\end{itemize}

Upon completing data collection for each vertex $u$, the algorithm proceeds with training a decision tree classifier $\traindecisiontree(D(u))$ for each vertex’s dataset. The trained model, denoted as $\boundary(u)$, establishes a guard for that particular vertex $u$.

\begin{algorithm}[t]
\caption{\decisionboundary($\G, \bestincomingedge$) \\ Training a Decision Tree Classifier at every edge where decision making is involved}
\begin{algorithmic}[1]
\STATE \textcolor{blue}{\textbackslash\textbackslash\ Creating Decision Sets}
\STATE Initialize $O(v) \gets \outgoing(v)$ for all $v \in \G$ 
\STATE Initialize $Q \gets F$ \textcolor{blue}{\textbackslash\textbackslash\ $Q$ is a queue}
\WHILE{$Q$ is not empty}
    \STATE vertex $u \gets Q.\dequeue$
    
    \FOR{vertex $v \in \incoming(u)$}
    \STATE $O(v).\remove(u)$
    \FOR{$i \in \traintasks$}
    \IF{$v \in (\bestincomingedge(u,i))$}
    \STATE $\De(e).\append(i)$ where edge $e = (u \rightarrow v)$
    \ENDIF
    \ENDFOR
    \IF{$O(v)$ is empty}
 
    \STATE $Q.\enqueue(v)$
    \ENDIF
    \ENDFOR          
\ENDWHILE

\STATE \textcolor{blue}{\textbackslash\textbackslash\ Learning Guard}
\STATE  dataset $D = \{(x, y) \mid x \in X, y \in Y\} \text{ where } X \gets \EnvInputValues(\task), Y \gets \text{edge } e$ 
\STATE $\boundary \gets \traindecisiontree$($D$) where $\boundary = (f: X \rightarrow Y$)
\STATE \textbf{return} $\boundary$
\end{algorithmic}
\label{algo:decisionboundary}
\end{algorithm}

\section{Experimental Setup: Implementation Level Details}
\label{sec:setupappen}
\paragraph{Model Configuration with Augmented Random Search}

We use Augmented Random Search (ARS) algorithm to train both the kappa vector and the base policy with the following specific hyperparameters: 

\begin{itemize}[leftmargin = 10pt]
    \item \textbf{Learning Rate:} The learning rate is conditionally step decayed, starting from 1 and decreasing to 0.1.
    \item \textbf{Number of Directions Sampled:} The model samples 30 directions per iteration. This sampling is part of the exploration strategy of ARS, allowing the model to investigate various policy adjustments.
    \item \textbf{Number of Top Samples used for Policy Update:} We use the top 8 samples for updating the policy. This means that out of all the directions sampled, the 8 with the highest rewards are used to guide the policy update.
    \item \textbf{Number of Steps in Training:} During training, the agent is allowed 15 steps in each iteration. This setting defines the length of each episode or trial used to evaluate the policy during training.
    \item \textbf{Number of Steps in Testing:} In testing, the agent is allowed a longer leash with 60 steps per iteration. This extended step count enables a more comprehensive evaluation of the trained policy.
    \item \textbf{Network Architecture:} The model's neural network consists of one input layer, two hidden layer, and one output layer. 
    \item \textbf{Activation Function:} The ReLU (Rectified Linear Unit) activation function is used for the input and the hidden layer and the tanh activation function is used for the output layer. 
    \item \textbf{No. of rollouts tested on:} During testing, we conduct 1000 rollouts for each task.
\end{itemize}

\paragraph{Pre-Processing Tasks.}

While training $\kappa_e$ for an edge $e \in \G$, we start by filtering out all tasks $\task_i \in \traintasks$ for which no feasible policy exists. We do so by creating task-specific training sets, $\traintasks_e$ for edges $e \in \G$; if a policy cannot be learnt for a certain edge (using $\learnpolicy$) for a task $\task_i$, we remove the task from $\traintasks_e$. Doing this improves the learning for the other tasks for which a feasible policy exists in the edge $e$ and makes sure the infeasible task does not affect the training of the other tasks.

\paragraph{Testing Methodology for Successful Unseen Tasks.} To test for successful unseen tasks, we iterate over each inductive task instance $\task_i$ that is not included in $\traintasks$. For each task, we increment the index $i$ and check the success probability of the task. If the success probability is above or equal to the success threshold (in our success threshold is $0.9$), we consider the unseen task $\task_i$ successful. We continue this process until we encounter five consecutive indices where the success probability is below $0.9$. At this point, we terminate our testing for successful unseen tasks.

\section{Cartesian Plane (Car2D) Benchmarks}
\label{sec:ap:environmentcar}

\subsection{Environment Description}
We consider a continuous cartesian plane which consists of a car which is free to move in the plane. The base objective of the training agent is to teach the car to reach the final goal from its initial position which could be any point in the 2D plane. In this environment, both the state space and action space are continuous in nature.

We add complexity to this task by providing an intermediate point for the car to meet before reaching the end goal or providing rectangular obstacles in its path of traversal.

The coordinates of the points are given as $(x, y)$ where $x$ denotes its position along the x-axis and $y$ denotes its position along the y-axis. The dimensions and coordinates of the obstacle are given by $(x_1, y_1, x_2, y_2)$ where $x_1, y_1$ gives the coordinates of the bottom left point of the obstacle and $x_2, y_2$ gives the coordinates of the top right point of the obstacle.

We use certain pre-defined predicates to define our tasks in the environment. 
Predicate $\reach$ is interpreted as reaching the coordinates of the specified point.  Predicate $\avoid$ is interpreted as avoiding the cartesian space of the defined obstacle.

Assume a rectangular obstacle $obs$ with $a = (a_x, a_y)$ as the bottom-left corner and with $b = (b_x, b_y)$ as the top-right corner, then

\begin{itemize}

\item $\reach$ holds true when point $s$ is near the point $goal$ w.r.t euclidean norm  $\|.\|_2$
$$\reach(goal) =  (\|s - goal\|_2 < \epsilon_1)$$

\item $\avoid$ holds true when point  $s$  is outside the rectangular region defined by its bottom-left corner $a$ and its top-right corner $b$

$$\avoid(obs) = ( s \notin [a_x, b_x] \times [a_y, b_y])$$

\end{itemize}

\subsection{1-Reachability Tasks with Simple Specifications}

\begin{figure*}[t]
    \centering
    \begin{minipage}[t]{0.3\textwidth}
        \begin{subfigure}[t]{\textwidth}
            \includegraphics[width=\textwidth]{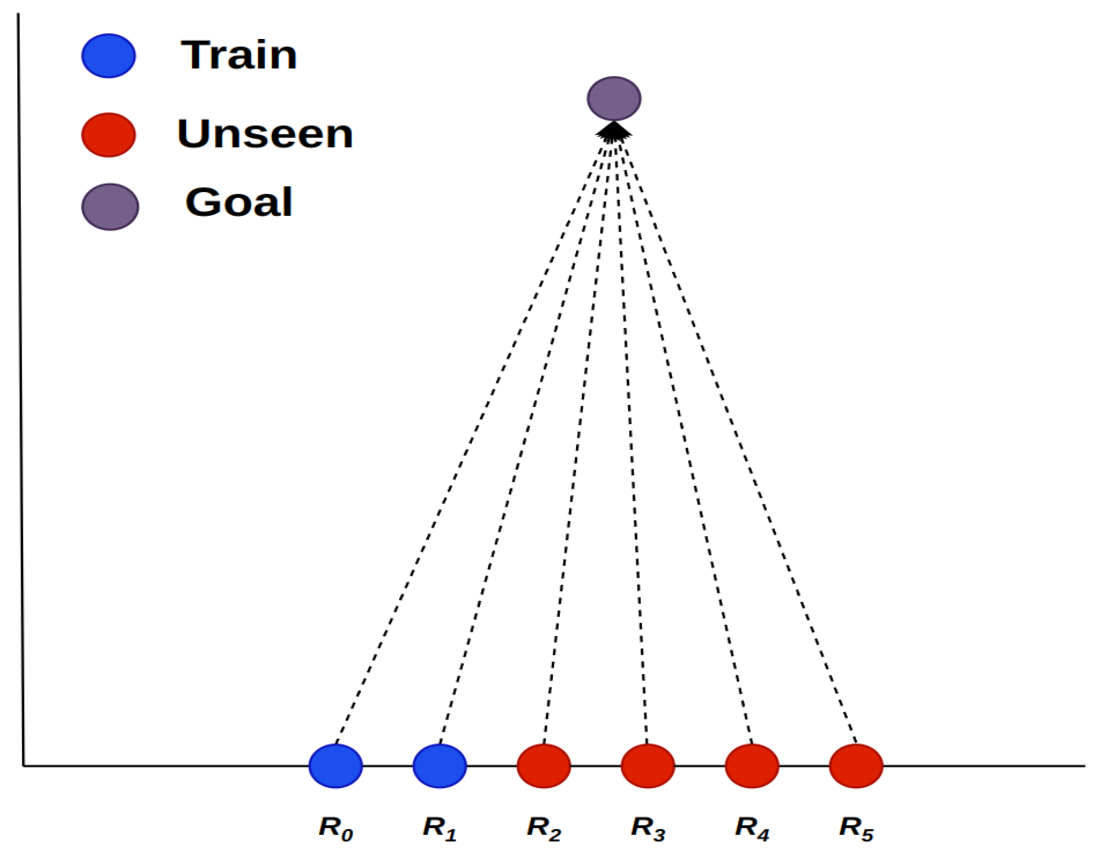} 
        \caption{}
        \label{fig:mov_init_illus}
        \end{subfigure}
        \vfill
        \begin{subfigure}[t]{\textwidth}
            \includegraphics[width=\textwidth]{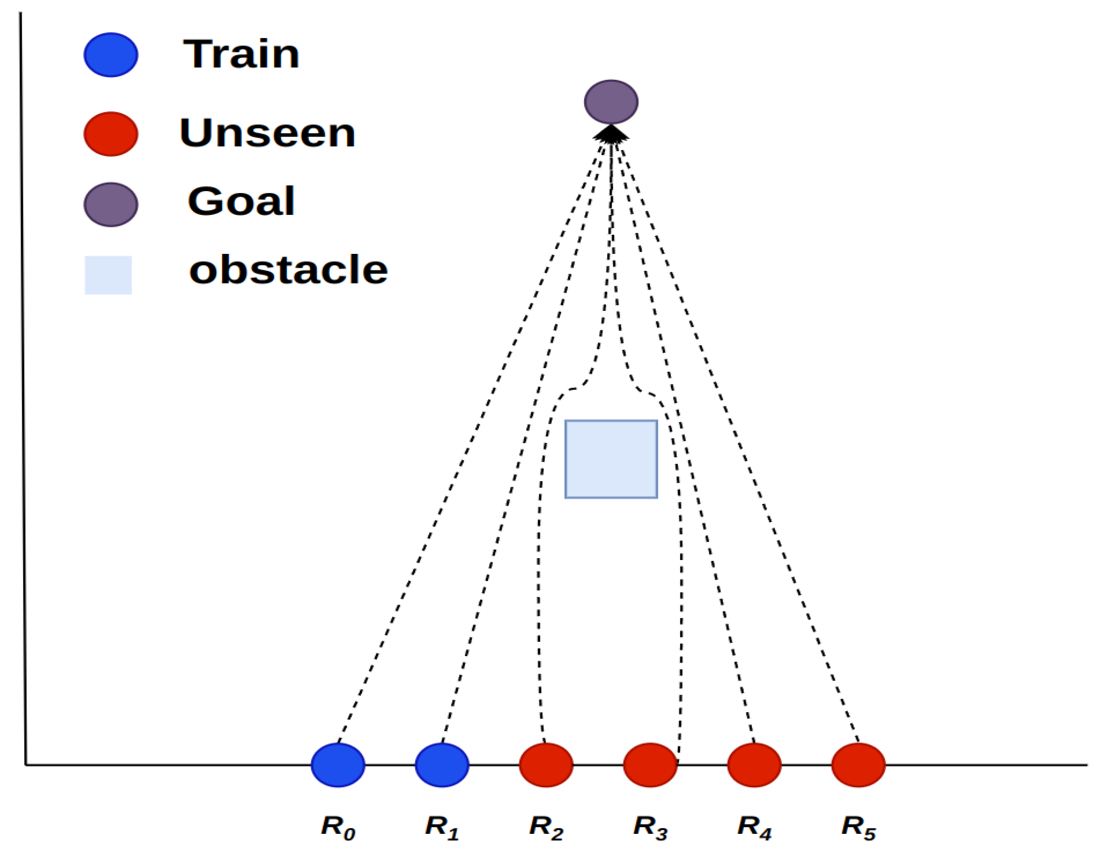}
        \caption{}
        \label{fig:mov_init_obs_illus}
        \end{subfigure}
        \caption{1-Reachability Inductive Task with moving initial distribution: a) without obstacle b) with obstacle}        
    \end{minipage}
    \hfill
    \begin{minipage}[t]{0.3\textwidth}
        \begin{subfigure}[t]{\textwidth}
            \includegraphics[width=\textwidth]{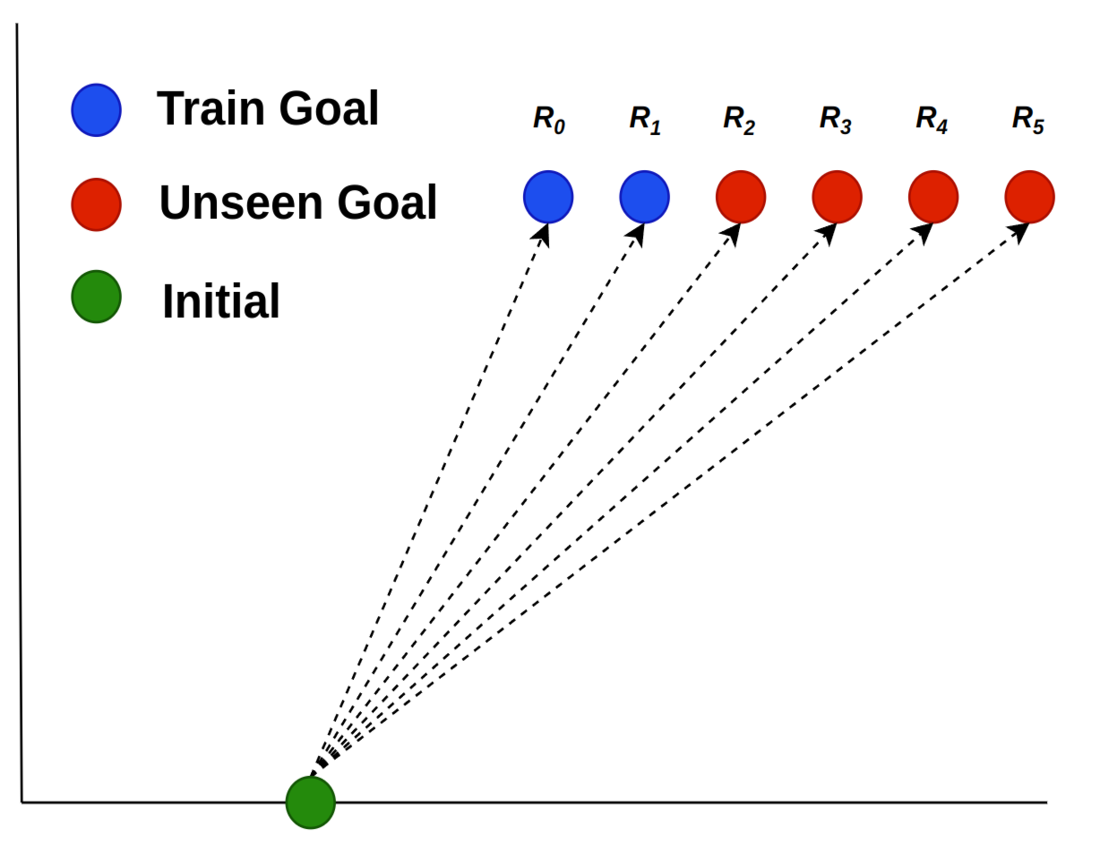} 
        \caption{}
        \label{fig:mov_goal_illus}
        \end{subfigure}
        \vfill
        \begin{subfigure}[t]{\textwidth}
            \includegraphics[width=\textwidth]{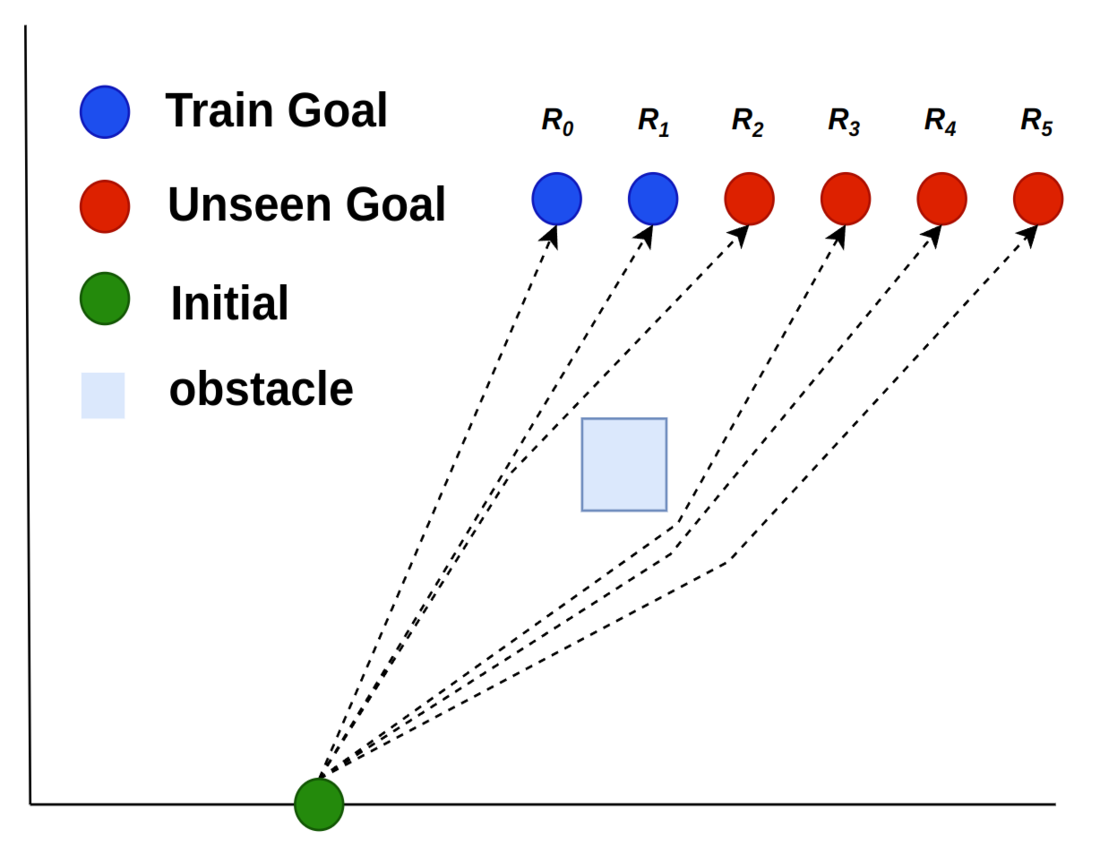}
        \caption{}
        \label{fig:mov_goal_obs_illus}
        \end{subfigure}
        \caption{1-Reachability Inductive Task with moving goal point: a) without obstacle b) with obstacle}
    
    \end{minipage}
    \hfill
    \begin{minipage}[t]{0.3\textwidth}
        \begin{subfigure}[t]{\textwidth}
            \includegraphics[width=\textwidth]{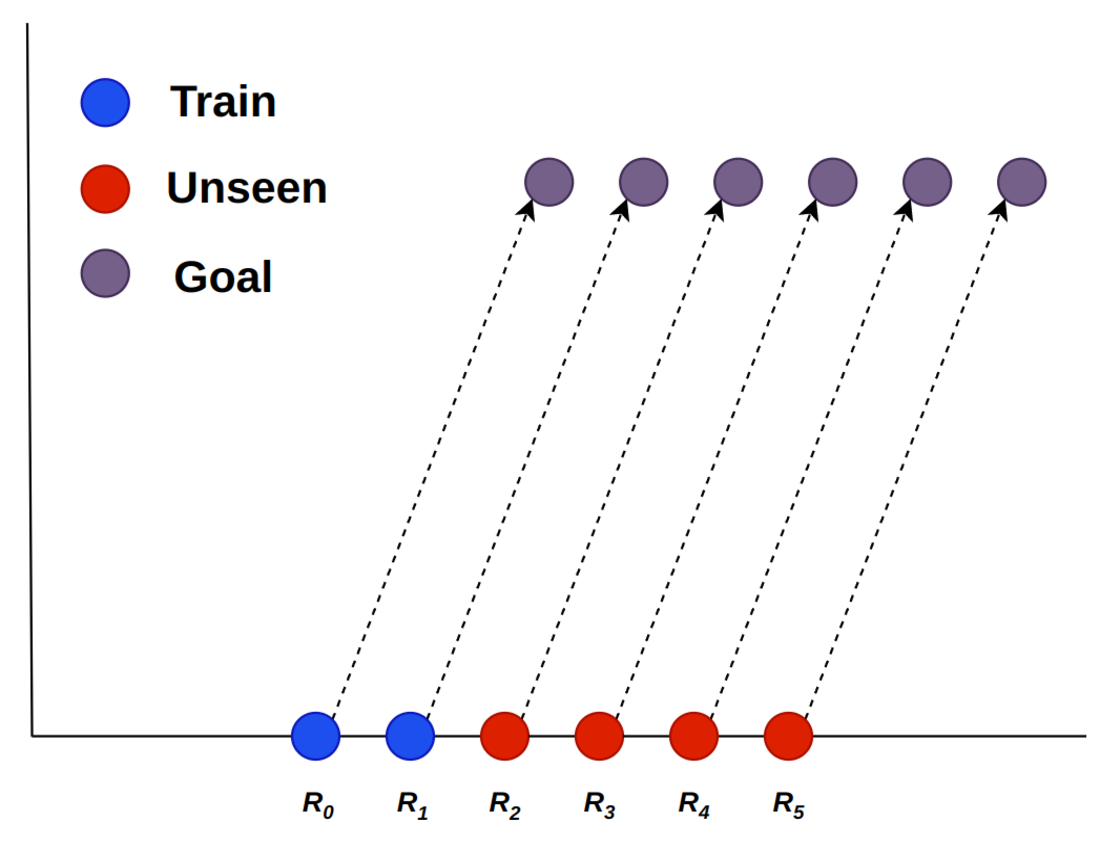} 
        \caption{}
        \label{fig:mov_illus}
        \end{subfigure}
        \vfill
        \begin{subfigure}[t]{\textwidth}
            \includegraphics[width=\textwidth]{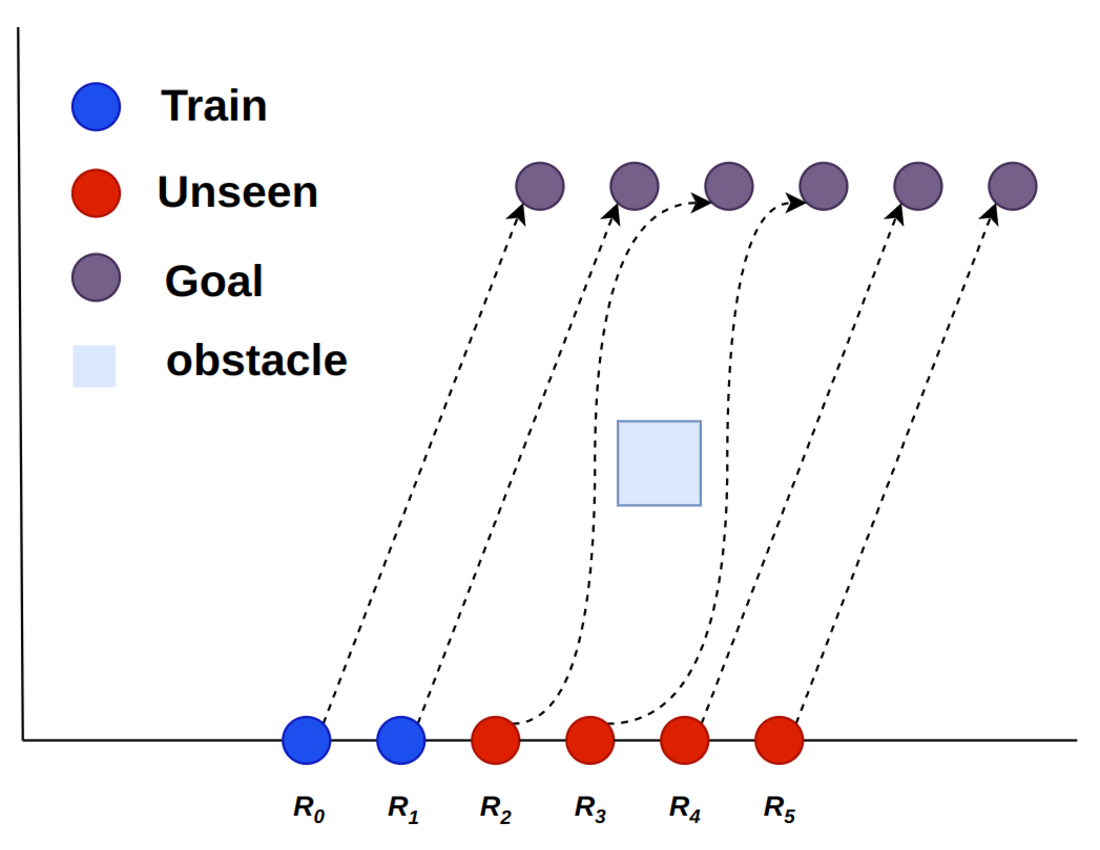}
        \caption{}
        \label{fig:mov_obs_illus}
        \end{subfigure}
        \caption{1-Reachability Inductive task with moving initial distribution and goal point: a) without obstacle b) with obstacle}
  
    \end{minipage}
\end{figure*}

In our 1-reachability experiments, we start from an initial distribution $\eta(s)$ and aim to reach a goal point $g_1$. We examine four different variations of 1-reachability tasks in our Car2D environment. These variations include inductive updates to the initial distribution and the goal state, both with and without obstacles which our car agent must avoid (Illustration in Figure~\ref{fig:mov_init_illus}, \ref{fig:mov_init_obs_illus}, \ref{fig:mov_goal_illus}, \ref{fig:mov_goal_obs_illus}, \ref{fig:mov_illus}, \ref{fig:mov_obs_illus}). These experiments help us assess the performance of our algorithms across a range of simple tasks.

\paragraph{RL Specifications.}

\begin{itemize}[leftmargin = 10pt]
    \item $1$-Reachability Task without obstacle: reach $g_1$ from $\eta(s)$ with updating initial distribution (Figure~\ref{fig:mov_init_illus}),
    \begin{align*}
        \eventually{(\reach(g_1))}; \ldots; \eventually{(\reach(g_n))}
    \end{align*}

    \item $1$-Reachability Task with obstacle: reach $g_1$ with updating initial distribution (Figure~\ref{fig:mov_init_obs_illus}),
    \begin{align*}
        \eventually{(\reach(g_1))}; \ldots; \eventually{(\reach(g_n))} \always (\avoid{(obs)})
    \end{align*}
\end{itemize}

The initial distribution is inductively updated by increasing the x-coordinate in successive instances of an inductive task, i.e. $\updateinit(\eta(s)) = \eta(s + (c_1, 0))$ where $c_1 = 0.5$ units.

\begin{itemize}[leftmargin = 10pt]
    \item $1$-Reachability Task without obstacle: reach $g_1$ from $\eta(s)$ with updating goal point (Figure~\ref{fig:mov_goal_illus}),
    \begin{align*}
        \eventually{(\reach(g_1))}; \ldots; \eventually{(\reach(g_n))}
    \end{align*}

    \item $1$-Reachability Task with obstacle: reach $g_1$ with updating goal point (Figure~\ref{fig:mov_goal_obs_illus}),
    \begin{align*}
        \eventually{(\reach(g_1))}; \ldots; \eventually{(\reach(g_n))} \always (\avoid{(obs)})
    \end{align*}
\end{itemize}

The goal is also inductively updated by increasing the x-coordinate in successive instances of an inductive task, i.e. $\updatepred(\reach(g_1)) = \reach(g_1 + (c_2, 0))$ where $c_2 = 0.5$ units. 

\begin{itemize}[leftmargin = 10pt]
    \item $1$-Reachability Task without obstacle: reach $g_1$ from $\eta(s)$ with updating initial distribution and goal point (Figure~\ref{fig:mov_illus}),
    \begin{align*}
        \eventually{(\reach(g_1))}; \ldots; \eventually{(\reach(g_n))}
    \end{align*}

    \item $1$-Reachability Task with obstacle: reach $g_1$ with updating initial distribution and goal point (Figure~\ref{fig:mov_obs_illus}),
    \begin{align*}
        \eventually{(\reach(g_1))}; \ldots; \eventually{(\reach(g_n))} \always (\avoid{(obs)})
    \end{align*}
\end{itemize}

The initial distribution and the goal are inductively updated by increasing their x-coordinates by \(c_1 = 0.5\) units and \(c_2 = 0.5\) units respectively in successive instances of an inductive task, i.e., \(\updateinit(\eta(s)) = \eta(s + (c_1, 0))\) and \(\updatepred(\reach(g_1)) = \reach(g_1 + (c_2, 0))\).

\begin{figure}[t]
    \centering
    \begin{minipage}[t]{0.50\textwidth}
        \centering
        \includegraphics[width=0.9\linewidth]{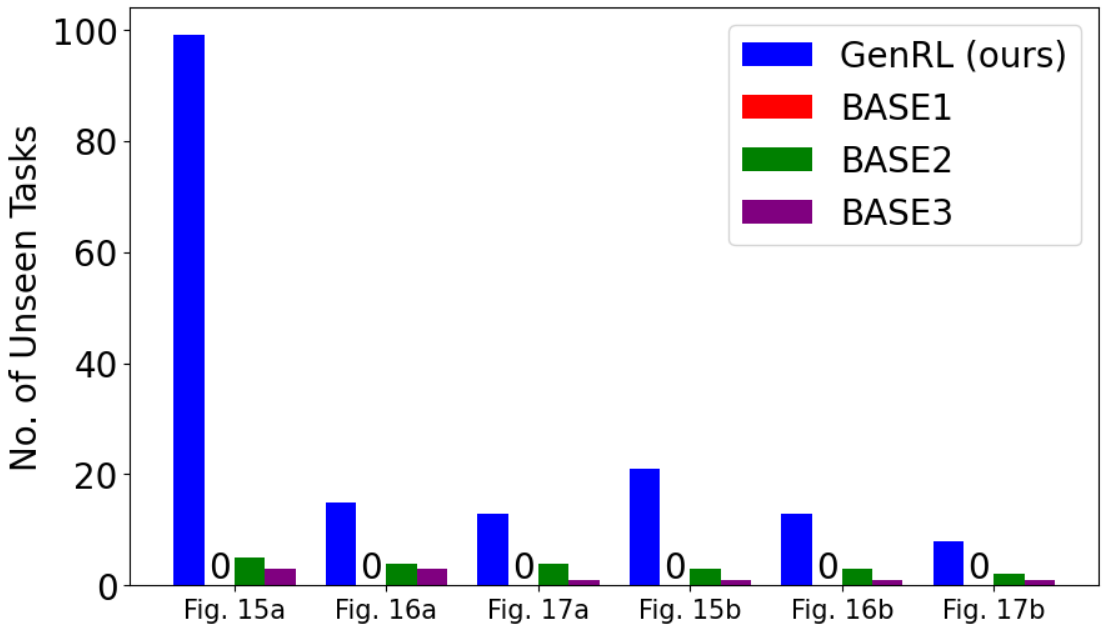}
        \caption{No. of successful Unseen Tasks for Simpler 1-Reachability experiments.}
        \label{fig:baselines_hist}
    \end{minipage}\hfill
    \begin{minipage}[t]{0.40\textwidth}
        \centering
        \includegraphics[width=\textwidth]{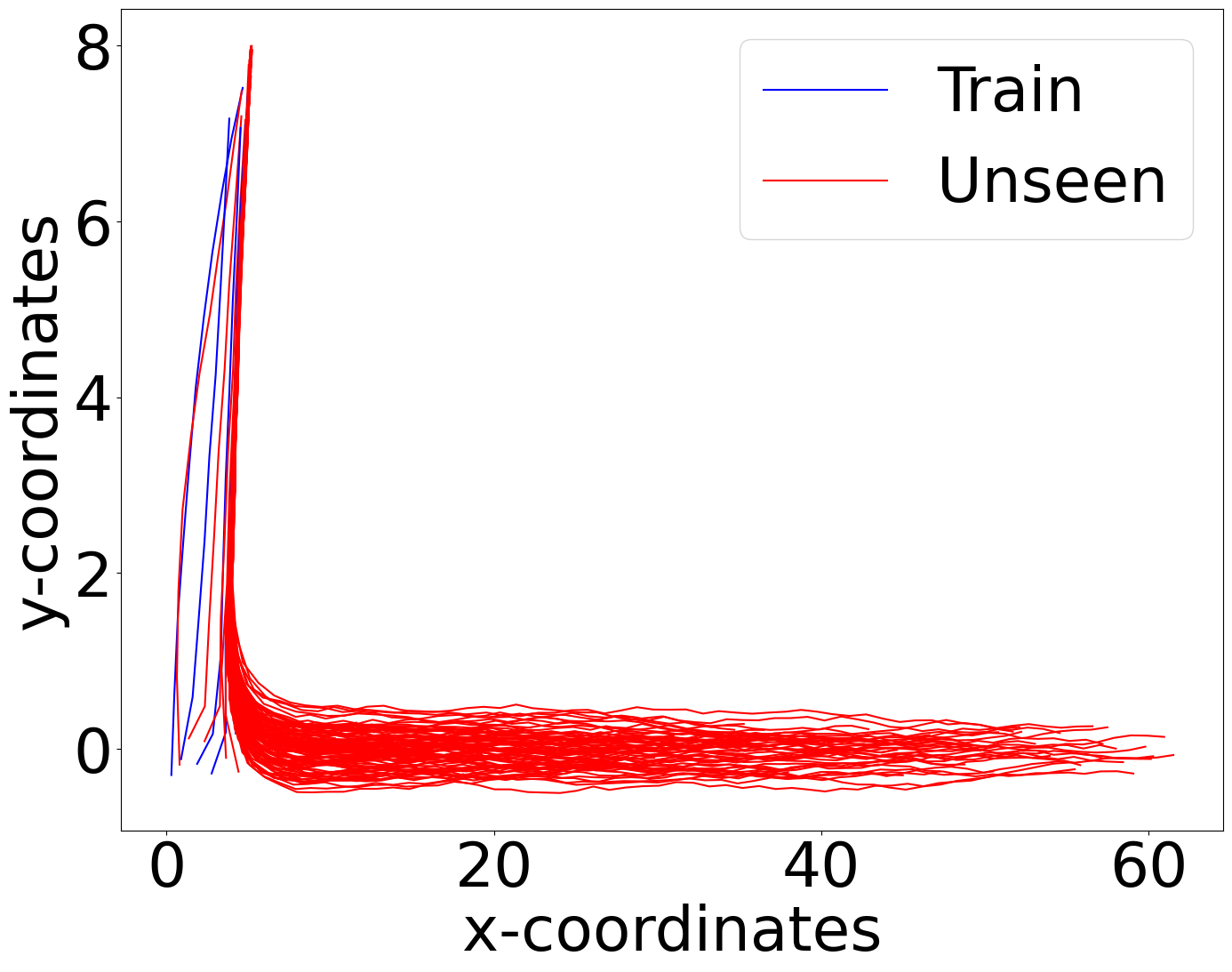}
        \caption{1-Reachability Task (updating initial distribution)}
        \label{fig:oneedgeTrajectory}
    \end{minipage}
\end{figure}

\begin{table}[t]
    \centering
    \setlength{\tabcolsep}{2pt}
    \begin{tabular}{|cc|ccccc|c|}
    \hline
    \multicolumn{2}{|c|}{\textbf{Benchmark}}  & \multicolumn{5}{c|}{\textbf{Iterations}}  & \multirow{2}{*}{\begin{tabular}[c]{@{}c@{}}\textbf{\% gen.}\\  \textbf{(Best Iter)}\end{tabular}} \\ \cline{3-7}
    \multicolumn{2}{|c|}{{$|\traintasks| = 6$}}                               & \multicolumn{1}{c|}{\textbf{200}} & \multicolumn{1}{c|}{\textbf{400}} & \multicolumn{1}{c|}{\textbf{600}} & \multicolumn{1}{c|}{\textbf{800}} & \textbf{1000} &                                                                                   \\ 
    \hline
    \multicolumn{8}{|c|}{\textbf{1-Reachability Task without Obstacle}} \\
    \hline
    \multicolumn{2}{|c|}{Figure~\ref{fig:mov_init_illus}}    & \multicolumn{1}{c|}{\textbf{116}} & \multicolumn{1}{c|}{99} & \multicolumn{1}{c|}{74} & \multicolumn{1}{c|}{15} & 109 & \multicolumn{1}{c|}{1933.33} \\
    \multicolumn{2}{|c|}{Figure~\ref{fig:mov_goal_illus}}          & \multicolumn{1}{c|}{\textbf{21}}  & \multicolumn{1}{c|}{15} & \multicolumn{1}{c|}{14} & \multicolumn{1}{c|}{10} & 10  & \multicolumn{1}{c|}{350} \\
    \multicolumn{2}{|c|}{Figure~\ref{fig:mov_illus}}          & \multicolumn{1}{c|}{\textbf{13}}  & \multicolumn{1}{c|}{\textbf{13}} & \multicolumn{1}{c|}{11} & \multicolumn{1}{c|}{11} & 11  & \multicolumn{1}{c|}{260} \\
    \hline
    \multicolumn{8}{|c|}{\textbf{1-Reachability Task with Obstacle}} \\
    \hline
    \multicolumn{2}{|c|}{Figure~\ref{fig:mov_init_obs_illus}}      & \multicolumn{1}{c|}{22}           & \multicolumn{1}{c|}{21} & \multicolumn{1}{c|}{16} & \multicolumn{1}{c|}{18} & \textbf{59} & \multicolumn{1}{c|}{983.33} \\
    \multicolumn{2}{|c|}{Figure~\ref{fig:mov_goal_obs_illus}}                 & \multicolumn{1}{c|}{\textbf{15}}  & \multicolumn{1}{c|}{13} & \multicolumn{1}{c|}{8}  & \multicolumn{1}{c|}{11} & 10  & \multicolumn{1}{c|}{250} \\
    \multicolumn{2}{|c|}{Figure~\ref{fig:mov_obs_illus}}          & \multicolumn{1}{c|}{\textbf{8}}  & \multicolumn{1}{c|}{\textbf{8}} & \multicolumn{1}{c|}{\textbf{8}} & \multicolumn{1}{c|}{\textbf{8}} & \textbf{8} & \multicolumn{1}{c|}{133.33} \\
    \hline
    \end{tabular}
    \caption{No. of successful Unseen task instances for Simpler 1-Reachability experiments across multiple iterations. The number in \textbf{boldface} under Iterations represents the best generalization for the specification.}
    \label{table:car2deasytasks:appendix}
\end{table}

\paragraph{Observations.}
From Table~\ref{table:car2deasytasks:appendix}, it is evident that our proposed $\genrl$ method demonstrates significantly superior generalizability across various iterations. We can also see that the $\genrl$ method consistently exhibits higher performance metrics than the baseline in every evaluated scenario from Figure~\ref{fig:baselines_hist}. Figure~\ref{fig:oneedgeTrajectory} shows the trajectory of the car agent on a 1-reachability task with updating initial distribution which shows our model's capability to generate policies that successfully satisfy unseen tasks for simpler environments and specifications.

\subsection{N-Reachability Tasks without obstacles}

\begin{figure}[t]
    \centering
    \begin{subfigure}{0.45\textwidth}
        \centering
        \includegraphics[width=\linewidth]{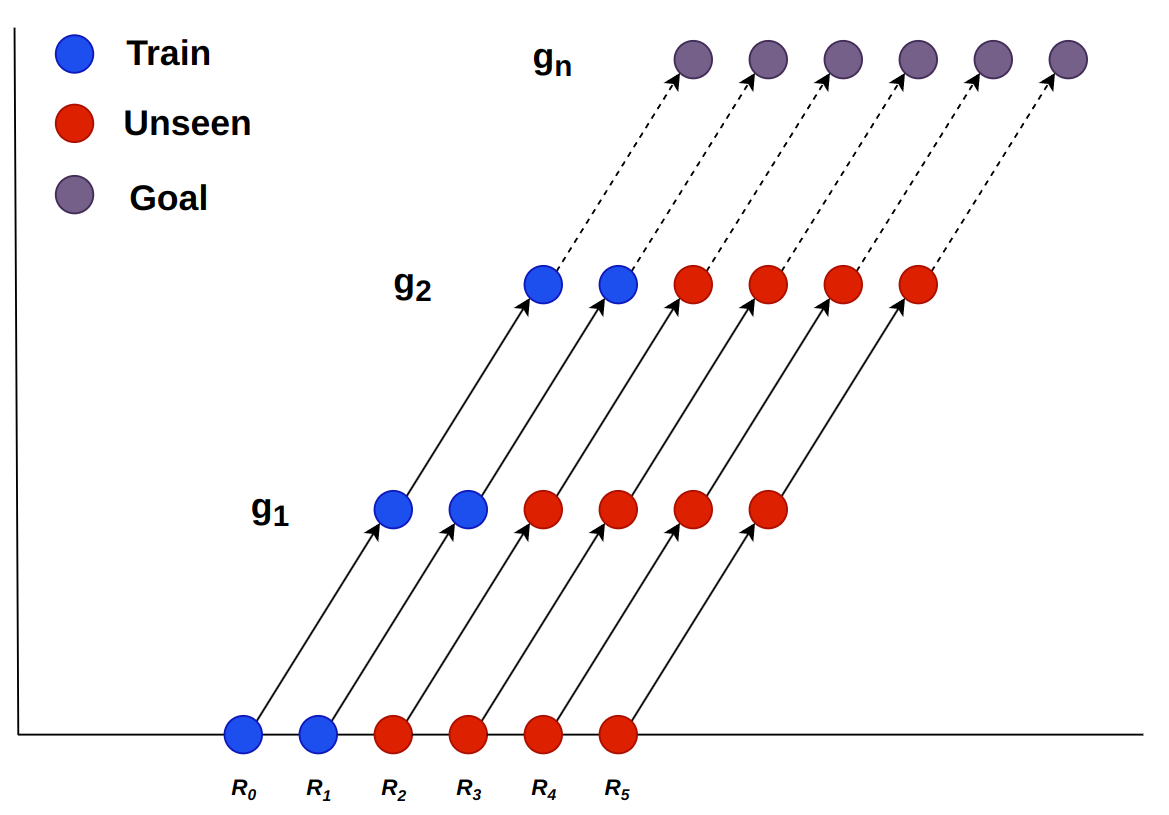}
        \caption{}
        \label{fig:nreach_illus}
    \end{subfigure}\hfill
    \begin{subfigure}{0.45\textwidth}
        \centering
        \includegraphics[width=\linewidth]{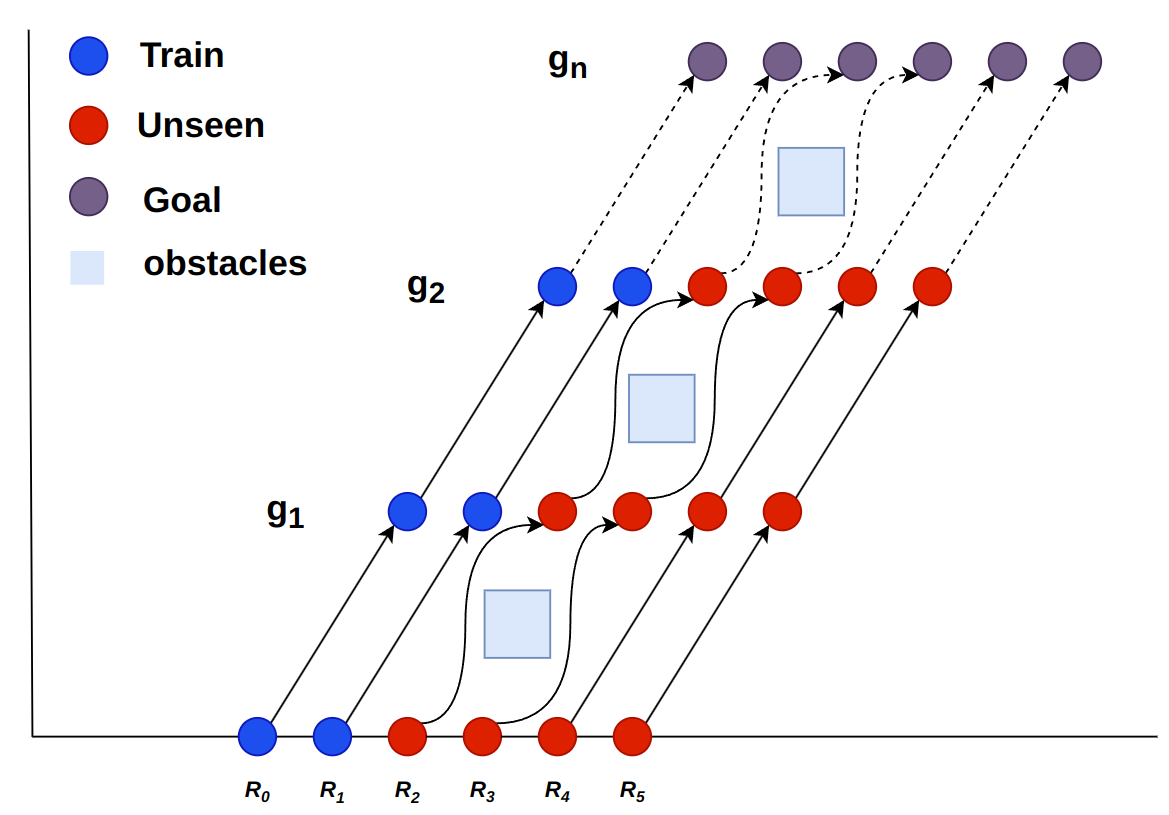}
        \caption{}
        \label{fig:nreach_obs_illus}
    \end{subfigure}
    \caption{N-Reachability Inductive Task: a) without obstacle b) with obstacle}
    \label{fig:combined_illustrations}
\end{figure}

In N-Reachability Tasks, we start from an initial distribution $\eta(s)$ and aim to reach the goal point $g_n$ while navigating via $g_1, \ldots, g_{n_1}$ intermediate goal points. This experiment is designed to test our model's generalization capabilities on long-horizon tasks. Here, the initial distribution $\eta(s)$ and all the goal points $g_1, \ldots, g_n$ inductively update.

\paragraph{RL Specifications.}
$N$-Reachability Task without obstacle (\textsf{NReach(n)}: reach a set of $n$ goals states $g_1, g_2, \dots, g_n$ in sequence (Figure~\ref{fig:nreach_illus}),

    \begin{align*}
        \eventually{(\reach(g_1))}; \ldots; \eventually{(\reach(g_n))}
    \end{align*}

The initial distribution is inductively updated by increasing the x-coordinate in successive instances of an inductive task, i.e. $\updateinit(\eta(s)) = \eta(s + (c_1, 0))$ where $c_1 = 0.5$ units. The goal is also inductively updated by increasing the x-coordinate in successive instances of an inductive task, i.e. $\updatepred(\reach(g_1, \ldots, g_n)) = \reach(g_1, \ldots, g_n + (c_2, 0))$ where $c_2 = 0.5$ units.

\begin{figure}[t]
    \centering
    \begin{minipage}[t]{0.50\textwidth}
        \centering
        \includegraphics[width=0.9\columnwidth]{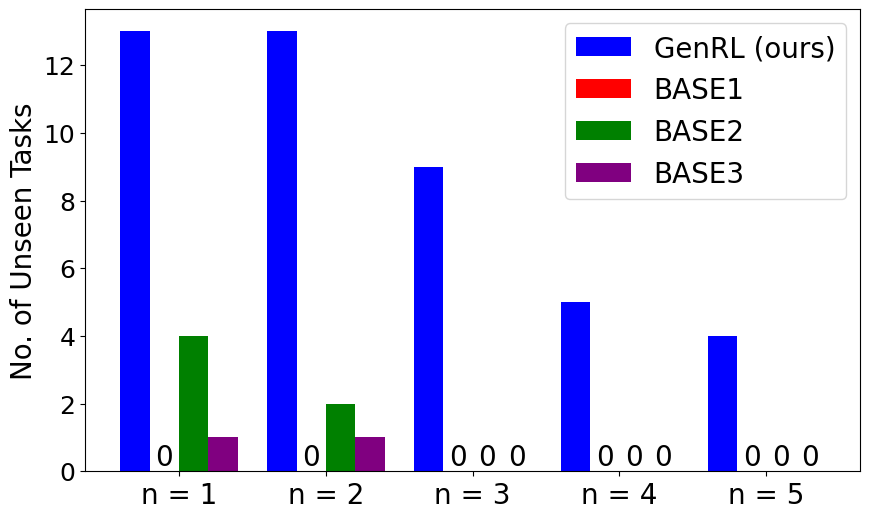}
        \caption{No. of successful Unseen Tasks for N-Reachability Experiments without obstacles.}
        \label{fig:nreachhist}
    \end{minipage}\hfill
    \begin{minipage}[t]{0.45\textwidth}
        \centering
        \includegraphics[width=0.9\columnwidth]{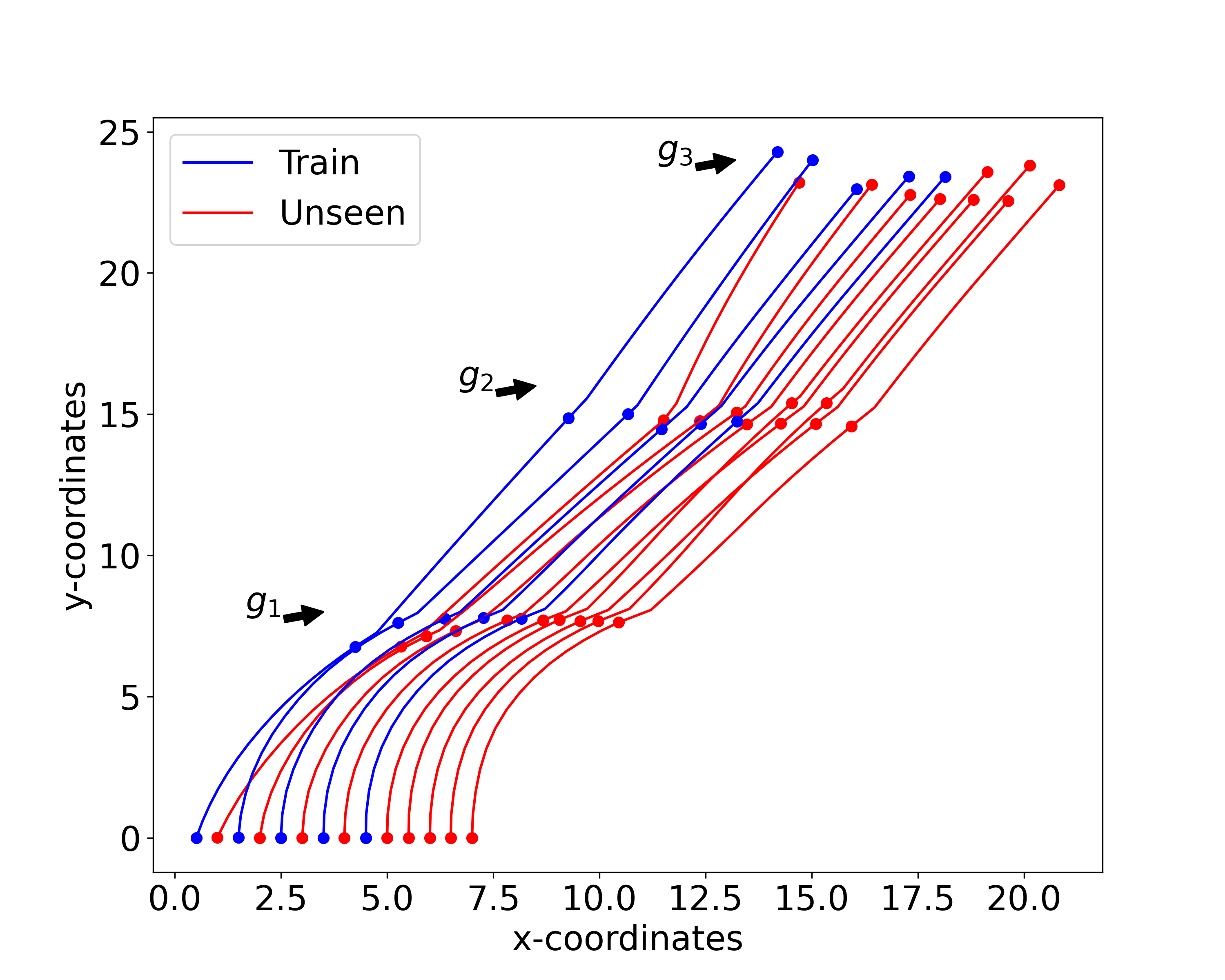}
        \caption{3-Reachability Task without obstacle (updating initial distribution and goal) - Trajectory}
        \label{fig:threereachtraj}
    \end{minipage}
\end{figure}

\begin{table}[t]
    \centering
    \setlength{\tabcolsep}{2pt}
    \begin{tabular}{|cc|ccccc|c|}
    \hline
    \multicolumn{2}{|c|}{\textbf{Benchmark}}                                                       & \multicolumn{5}{c|}{\textbf{Iterations}}                                                                   & \multirow{2}{*}{\begin{tabular}[c]{@{}c@{}}\textbf{\textbf{\% gen.}}\\  \textbf{(Best Iter)}\end{tabular}} \\ \cline{3-7}
    \multicolumn{2}{|c|}{$|\traintasks| = 6$} & \multicolumn{1}{c|}{\textbf{200}} & \multicolumn{1}{c|}{\textbf{400}} & \multicolumn{1}{c|}{\textbf{600}} & \multicolumn{1}{c|}{\textbf{800}} & \textbf{1000} & \\ 
    \hline
    \multicolumn{2}{|c|}{\textsf{NReach(1)} }         & \multicolumn{1}{c|}{\textbf{13}}  & \multicolumn{1}{c|}{\textbf{13}} & \multicolumn{1}{c|}{11} & \multicolumn{1}{c|}{11} & 11  & \multicolumn{1}{c|}{217} \\
    \multicolumn{2}{|c|}{\textsf{NReach(2)}}     & \multicolumn{1}{c|}{12}           & \multicolumn{1}{c|}{\textbf{13}} & \multicolumn{1}{c|}{11} & \multicolumn{1}{c|}{11} & 11  & \multicolumn{1}{c|}{217} \\
    \multicolumn{2}{|c|}{\textsf{NReach(3)} }                 & \multicolumn{1}{c|}{6}            & \multicolumn{1}{c|}{\textbf{9}}  & \multicolumn{1}{c|}{8}  & \multicolumn{1}{c|}{8}  & 8   & \multicolumn{1}{c|}{150} \\
    \multicolumn{2}{|c|}{\textsf{NReach(4)} }           & \multicolumn{1}{c|}{5}            & \multicolumn{1}{c|}{5}           & \multicolumn{1}{c|}{\textbf{6}} & \multicolumn{1}{c|}{\textbf{6}} & \textbf{6} & \multicolumn{1}{c|}{100} \\
    \multicolumn{2}{|c|}{\textsf{NReach(5)} }            & \multicolumn{1}{c|}{4}            & \multicolumn{1}{c|}{4}           & \multicolumn{1}{c|}{\textbf{5}} & \multicolumn{1}{c|}{\textbf{5}} & \textbf{5} & \multicolumn{1}{c|}{83} \\
    \hline
    \end{tabular}
    \caption{No. of successful Unseen task instances for N-Reachability experiments without obstacles across multiple iterations. The number in \textbf{boldface} under Iterations represents the best generalization for the benchmark.}
    \label{table:car2dreachability:merged}
\end{table}

\paragraph{Observations.}

Table~\ref{table:car2dreachability:merged} clearly demonstrates that $\genrl$ exhibits high generalizability across multiple iterations in N-reachability tasks. Figure~\ref{fig:nreachhist} highlights that $\genrl$ consistently outperforms the baselines in terms of satisfying unseen tasks across all N-reachability benchmarks. Figure~\ref{fig:threereachtraj} illustrates the trajectory of the car agent in a \textsf{NReach(3)} task where we can see how unseen tasks (marked in red) traverses through the intermediate goal points to reach $g_3$.

\subsection{N-Reachability Tasks with Obstacles}

In N-Reachability Tasks, we start from an initial distribution $\eta(s)$ and aim to reach the goal point $g_n$ while navigating via $g_1, \ldots, g_{n_1}$ intermediate goal points while avoiding the obstacles $obs$. This experiment is designed to test our model's generalization capabilities on long-horizon tasks. Here, the initial distribution $\eta(s)$ and all the goal points $g_1, \ldots, g_n$ inductively update.

\paragraph{RL Specifications.}
$N$-Reachability Task with obstacle (\textsf{NReachObs(n)}): reach a set of $n$ goals $g_1, g_2, \dots, g_n$ in sequence \textit{while avoiding the obstacles} in $obs$ (Figure~\ref{fig:nreach_obs_illus}),

    \begin{align*}
        &\eventually{(\reach(g_1))}; \ldots; \eventually{(\reach(g_n))} \\
        &\always(\avoid(obs))
    \end{align*}

The initial distribution is inductively updated by increasing the x-coordinate in successive instances of an inductive task, i.e. $\updateinit(\eta(s)) = \eta(s + (c_1, 0))$ where $c_1 = 0.5$ units. The goal is also inductively updated by increasing the x-coordinate in successive instances of an inductive task, i.e. $\updatepred(\reach(g_1, \ldots, g_n)) = \reach(g_1, \ldots, g_n + (c_2, 0))$ where $c_2 = 0.5$ units. 

\begin{figure}[t]
    \centering
    \begin{minipage}[t]{0.50\textwidth}
        \centering
        \includegraphics[width=0.9\textwidth]{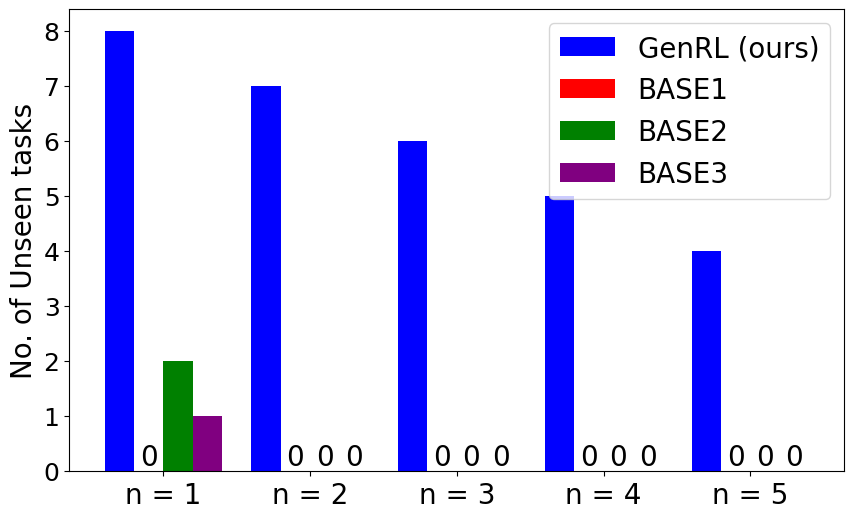}
        \caption{No. of successful Unseen Tasks for N-Reachability Experiments with obstacles.}
        \label{fig:baselines_hist_obs}
    \end{minipage}\hfill
    \begin{minipage}[t]{0.45\textwidth}
        \centering
        \includegraphics[width=0.9\textwidth]{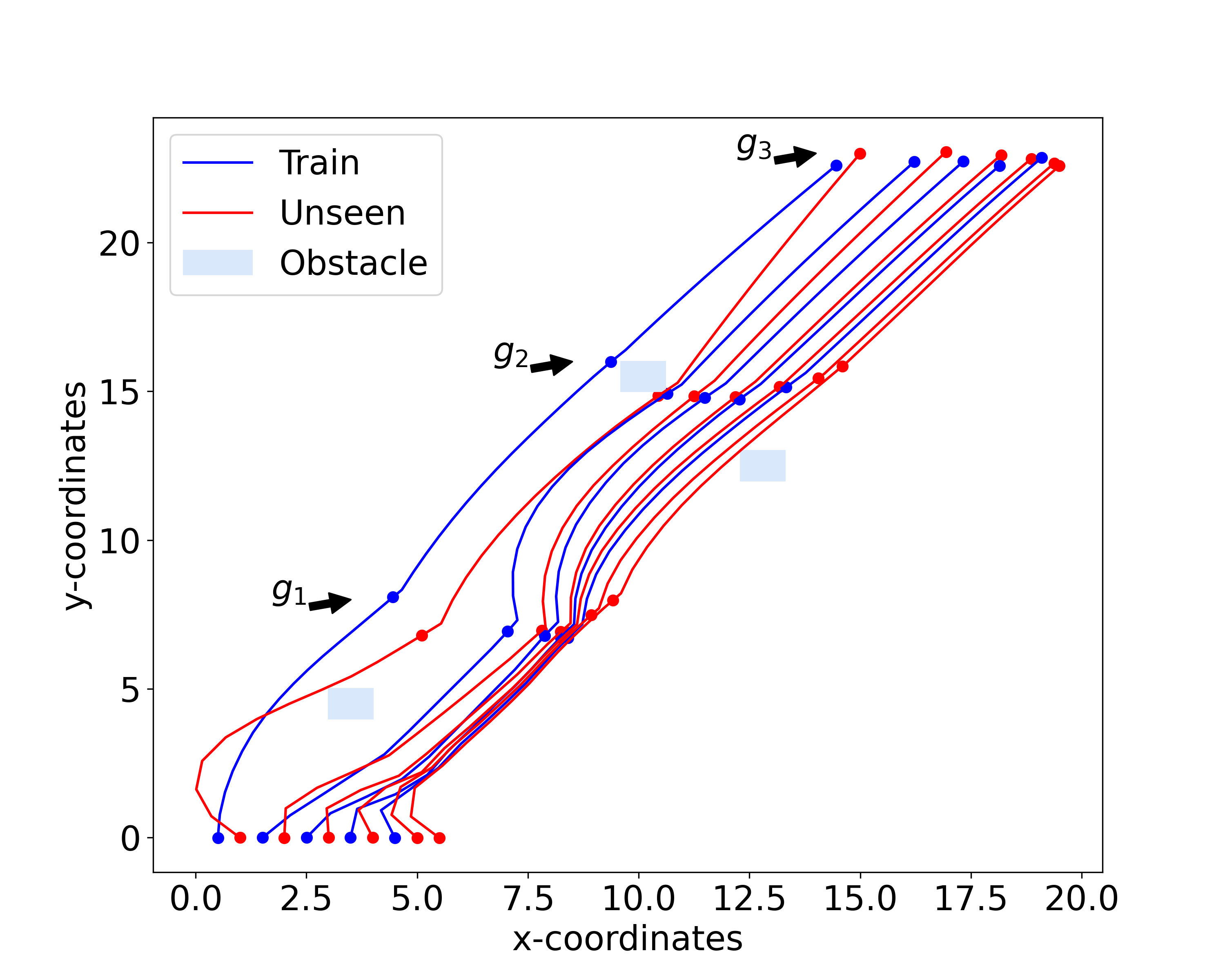}
        \caption{3-Reachability Task with obstacle (updating initial distribution and goal) - Trajectory}
        \label{fig:threereachobstraj}
    \end{minipage}
\end{figure}

\paragraph{Observations.}
Table~\ref{table:car2dreachability:mergedappendixobs} clearly shows that $\genrl$ has very high generalizability across multiple iterations in tasks with obstacles. Figure~\ref{fig:baselines_hist_obs} shows that $\genrl$ consistently achieves better generalizability than the baselines in all scenarios. However, it can be noted that the generalizability of $\genrl$ shows a marginal decline when compared to tasks without obstacles. This observation can be attributed to the increased complexity and difficulty presented by the obstacle specifications. Figure~\ref{fig:threereachobstraj} shows the trajectory of the car agent on a \textsf{NReachObs(3)} task.

\begin{table}[ht]
    \centering
    \setlength{\tabcolsep}{2pt}
    \begin{tabular}{|cc|ccccc|c|}
    \hline
    \multicolumn{2}{|c|}{\textbf{Benchmark}}  & \multicolumn{5}{c|}{\textbf{Iterations}}  & \multirow{2}{*}{\begin{tabular}[c]{@{}c@{}}\textbf{\textbf{\% gen.}}\\  \textbf{(Best Iter)}\end{tabular}} \\ \cline{3-7}
    \multicolumn{2}{|c|}{{$|\traintasks| = 6$}} & \multicolumn{1}{c|}{\textbf{200}} & \multicolumn{1}{c|}{\textbf{400}} & \multicolumn{1}{c|}{\textbf{600}} & \multicolumn{1}{c|}{\textbf{800}} & \textbf{1000} & \\ 
    \hline
    \multicolumn{2}{|c|}{\textsf{NReachObs(1)}}    &     \multicolumn{1}{c|}{\textbf{8}}   & \multicolumn{1}{c|}{\textbf{8}}  & \multicolumn{1}{c|}{\textbf{8}} & \multicolumn{1}{c|}{\textbf{8}} & \textbf{8} & \multicolumn{1}{c|}{133.33} \\
    \multicolumn{2}{|c|}{\textsf{NReachObs(2)}}          & \multicolumn{1}{c|}{\textbf{8}}   & \multicolumn{1}{c|}{7}           & \multicolumn{1}{c|}{7}          & \multicolumn{1}{c|}{7}          & 7   & \multicolumn{1}{c|}{133.33} \\
    \multicolumn{2}{|c|}{\textsf{NReachObs(3)}}           & \multicolumn{1}{c|}{\textbf{6}}   & \multicolumn{1}{c|}{\textbf{6}}  & \multicolumn{1}{c|}{\textbf{6}} & \multicolumn{1}{c|}{\textbf{6}} & \textbf{6} & \multicolumn{1}{c|}{100} \\
    \multicolumn{2}{|c|}{\textsf{NReachObs(4)}}          & \multicolumn{1}{c|}{4}            & \multicolumn{1}{c|}{\textbf{5}}  & \multicolumn{1}{c|}{\textbf{5}} & \multicolumn{1}{c|}{\textbf{5}} & \textbf{5} & \multicolumn{1}{c|}{83.33} \\
    \multicolumn{2}{|c|}{\textsf{NReachObs(5)}}   & \multicolumn{1}{c|}{4}            & \multicolumn{1}{c|}{4}           & \multicolumn{1}{c|}{\textbf{5}} & \multicolumn{1}{c|}{\textbf{5}} & \textbf{5} & \multicolumn{1}{c|}{83.33} \\
    \hline
    \end{tabular}
    \caption{No. of successful Unseen task instances for N-Reachability experiments with obstacles across multiple iterations. The number in \textbf{boldface} under Iterations represents the best generalization for the benchmark.}
    \label{table:car2dreachability:mergedappendixobs}
\end{table}

\subsection{Choice Tasks}

In choice tasks, we start from an initial distribution $\eta(s)$ and choose between navigating to intermediate goal $g_1$ or $g_2$ based on reachability and then reach $goal$ while avoiding obstacle $obs$. This experiment allows us to test the optimality of our guards (branching predicate to choose between the goals) and test long-horizon reachability on complex decision-involving specifications (Illustration in Figure~\ref{fig:choice1illus}, \ref{fig:choice2illus}, \ref{fig:choice3illus}).

\paragraph{RL Specifications.}
Choice Task - \textsf{Choice(l)} (Figure~\ref{fig:choice1illus}, \ref{fig:choice2illus}, \ref{fig:choice3illus}): A stack of $l$ sub-tasks (or \textit{levels}), where each sub-task $i$ requires reaching a goal $goal_i$ while avoiding the obstacle $obs_i$, either through the (sub)goal $g_{i1}$ or $g_{i2}$,
    \begin{align*}
        &(\eventually{(\choice{\reach(g_{i1})}{\reach(g_{i2})})}; \\
        &\hspace*{20pt}\eventually{(\reach(goal_i))})^l\\ 
        &\always(\avoid(obs))
    \end{align*}
    where, $1  \leq i \leq l$.

We use the superscript $l$ to indicate that the enclosed specification is repeated $l$ times. The inductive tasks on the above tasks inductively modify the initial distribution $\eta(s)$, and the goal positions $goal_i$. The initial distribution is inductively updated by increasing the x-coordinate in successive instances of an inductive task, i.e. $\updateinit(\eta(s)) = \eta(s + (c_1, 0))$ where $c_1 = 1$ units. The goal is also inductively updated by increasing the x-coordinate in successive instances of an inductive task, i.e. $\updatepred(\reach(goal_i)) = \reach(goal_i + (c_2, 0))$ where $c_2 = 1$ units. 

\paragraph{Observations.}

\begin{figure*}[t]
\centering
\begin{minipage}[t]{0.49\textwidth}
\centering
\includegraphics[width=\textwidth]{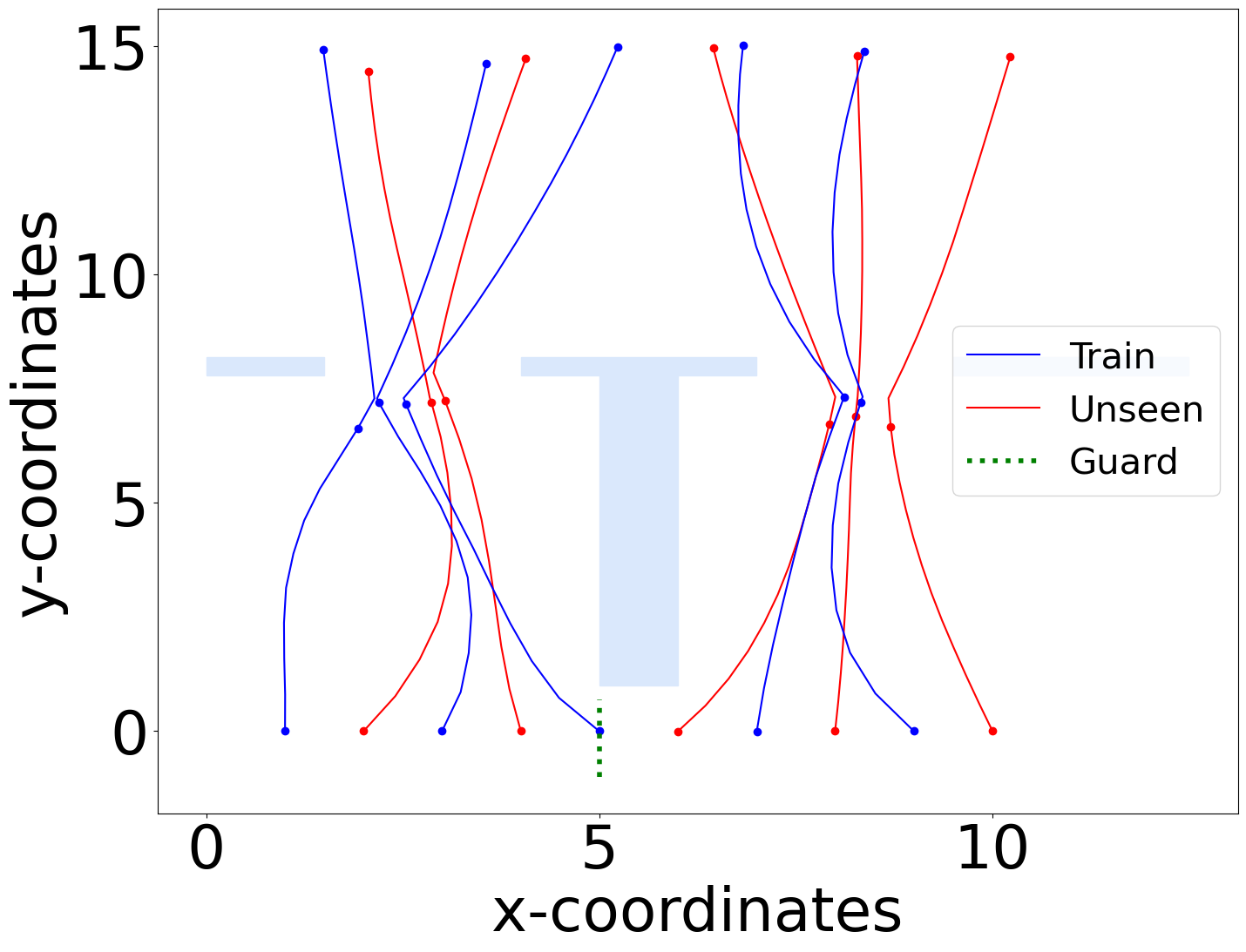}
\caption{Choice with moving goal - Trajectory}
\label{fig:choice2traj}
\end{minipage}
\hfill
\begin{minipage}[t]{0.49\textwidth}
\centering
\includegraphics[width=\textwidth]{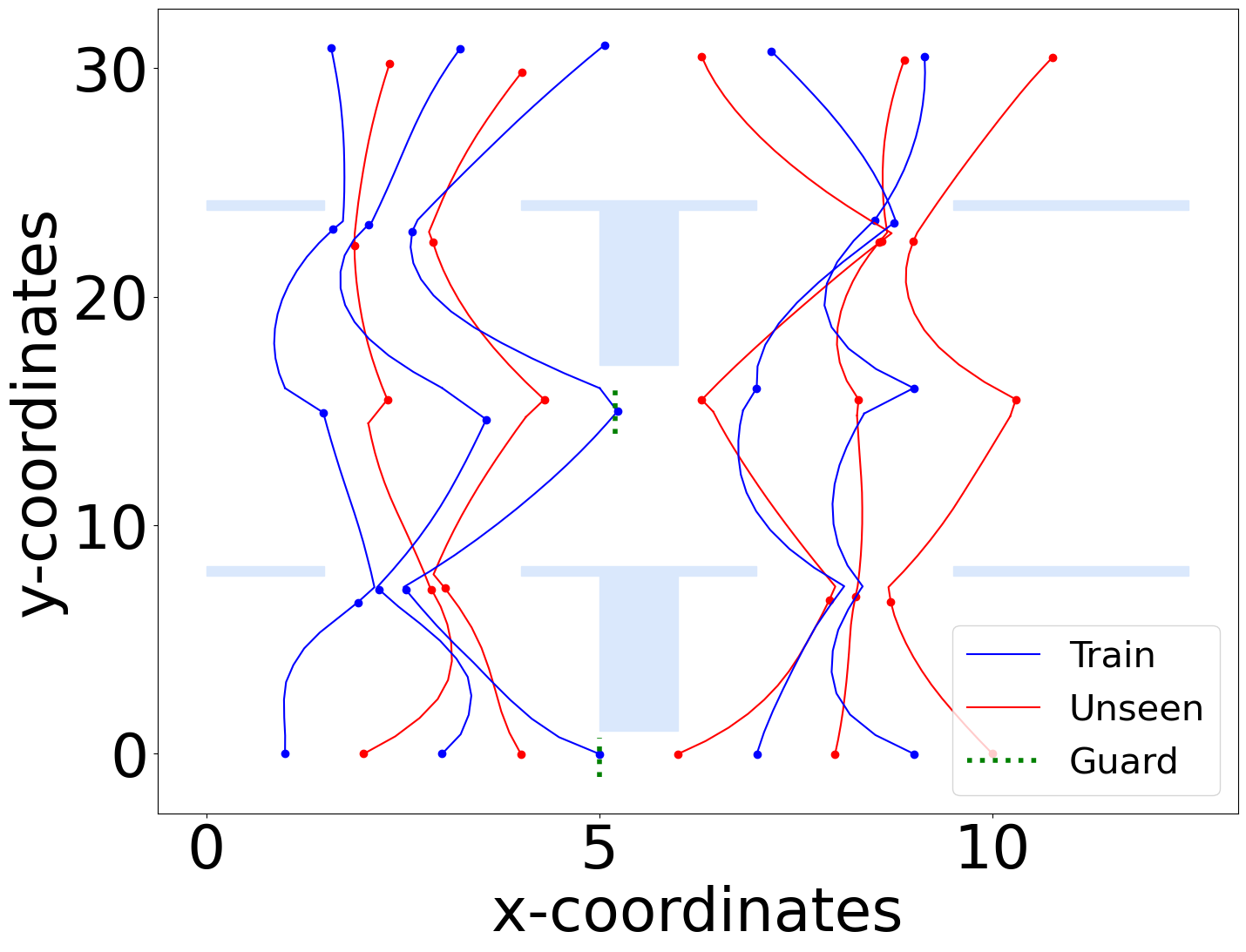}
\caption{Two levels of choice with moving goal - Trajectory}
\label{fig:choice3traj}
\end{minipage}
\end{figure*}

From Table~\ref{table:car2dchoice:mergedappendix}, it is evident that our proposed $\genrl$ method demonstrates significantly superior generalizability across various iterations. Figures~\ref{fig:choice1traj}, \ref{fig:choice2traj}, and \ref{fig:choice3traj} illustrate how our learned guard optimally indicates which edge to traverse based on the task index, enabling collision-free traversal. More results and observations on Section~\ref{sec:longhorsimdyn}.

\begin{table}[t]
\centering       
\setlength{\tabcolsep}{2pt}
\small
\begin{tabular}{|cc|ccccc|c|c|}
\hline
\multicolumn{2}{|c|}{\textbf{Benchmark}} & \multicolumn{5}{c|}{\textbf{Iterations}}                                                                 & \multirow{2}{*}{\begin{tabular}[c]{@{}c@{}}\textbf{\textbf{\% gen.}}\\  \textbf{(Best Iter)}\end{tabular}} & \multirow{2}{*}{\begin{tabular}[c]{@{}c@{}}\textbf{Guard}\\  \textbf{predicate}\end{tabular}}  \\   \cline{3-7}
\multicolumn{2}{|c|}{$|\traintasks|= 6$} & \multicolumn{1}{c|}{\textbf{200}} & \multicolumn{1}{c|}{\textbf{400}} & \multicolumn{1}{c|}{\textbf{600}} & \multicolumn{1}{c|}{\textbf{800}} & \textbf{1000} &   &  \\ 
\hline
\multicolumn{2}{|l|}{Figure~\ref{fig:motivating_example}}         & \multicolumn{1}{c|}{{6}} & \multicolumn{1}{c|}{6} & \multicolumn{1}{c|}{\textbf{7}} & \multicolumn{1}{c|}{\textbf{7}} & \textbf{7} & \multicolumn{1}{c|}{117} & $(i \leq 5)$\\
\multicolumn{2}{|l|}{Figure~\ref{fig:choice2illus}}                    & \multicolumn{1}{c|}{FAIL}  & \multicolumn{1}{c|}{\textbf{5}} & \multicolumn{1}{c|}{\textbf{5}} & \multicolumn{1}{c|}{\textbf{5}} & \textbf{5}  & \multicolumn{1}{c|}{84} & $(i\leq5)$\\
\multicolumn{2}{|l|}{Figure~\ref{fig:choice3illus}}                      & \multicolumn{1}{c|}{FAIL}  & \multicolumn{1}{c|}{{4}} & \multicolumn{1}{c|}{\textbf{5}} & \multicolumn{1}{c|}{\textbf{5}} & \textbf{5}  & \multicolumn{1}{c|}{84} & $(i\leq5), (i\leq5)$\\
\hline
\end{tabular}
\caption{No. of successful Unseen task instances for Choice experiments. The number in \textbf{boldface} under Iterations represents the best generalization for the benchmark.} 
\label{table:car2dchoice:mergedappendix}
\end{table}
\section{Tower-Destacking Benchmarks}
\label{sec:reacherenv}

\paragraph{Environment Description.} The 2-link planar arm environment is similar to the Reacher-v2 environment with a modified state and action space. The dynamics of the arm can be described through the joint angles $\theta_1$ and $\theta_2$, which dictate the Cartesian coordinates $x$ and $y$ of the end effector. The relationship between these variables is captured by the forward kinematics equations:

  \begin{equation*}
    x = l_1 \cos(\theta_1) + l_2 \cos(\theta_1 + \theta_2)
  \end{equation*}
  \begin{equation*}
    y = l_1 \sin(\theta_1) + l_2 \sin(\theta_1 + \theta_2)
  \end{equation*}

where $l_1$ and $l_2$ represent the lengths of the first and second links, respectively. The state space includes the x and y cartesian coordinates, while the action space consists of $\theta_1$ and $\theta_2$. 

The predicate $\reach$ holds true when point $s$ is inside the rectangular region (block) defined by its bottom-left corner $a$ and its top-right corner $b$

$$\reach(s) = ( s \in [a_x, b_x] \times [a_y, b_y])$$

\paragraph{RL Specifications.} We create five distinct benchmarks to evaluate our model's ability to generalize to long-horizon, complex tasks in this environment.

\begin{itemize}[leftmargin = 10pt]

 \item \textbf{Pick and Drop - Same Side (Figure~\ref{fig:r3})}

    The task involves performing a pick-and-drop operation. Starting from a designated target dropbox, the arm grabs a box from the source tower and moves it to the dropbox where both the target dropbox and the source tower are on the same side of the arm.
    
    Length of arm: $l_1 = 10$ units and $l_2 = 10$ units.

    \begin{align*}
    \eventually({\code{reach}(\mathsf{top\_block\_source\_tower}})); \eventually( \code{reach}(\mathsf{target\_box})
    \end{align*}

    \item \textbf{Pick and Vertical Stack - Same Side (Figure~\ref{fig:motivating_example_reacher})}

    The task involves performing a pick-and-drop operation. Starting from a designated source stack of boxes, the arm grabs a box from the source stack and stacks it vertically onto a target stack where both the source and the target stacks are on the same side of the arm.
    
    Length of arm: $l_1 = 10$ units and $l_2 = 10$ units.
    
    \begin{align*}
    \eventually({\code{reach}(\mathsf{top\_block\_target\_tower}})); \eventually( \code{reach}(\mathsf{top\_block\_source\_tower})
    \end{align*}

\item \textbf{Pick and Drop - Opposite Side (Figure~\ref{fig:r1})}

    The task involves performing a pick-and-drop operation. Starting from a designated target dropbox, the arm grabs a box from the source tower and moves it to the dropbox where both the target dropbox and the source tower are on the opposite side of the arm.
    
    Length of arm: $l_1 = 5$ units and $l_2 = 5$ units.
    
    \begin{align*}
    \eventually({\code{reach}(\mathsf{top\_block\_source\_tower}})); \eventually( \code{reach}(\mathsf{target\_box})
    \end{align*}

    \item \textbf{Pick and Vertical Stack - Opposite Side (Figure~\ref{fig:r2})}

    The task involves performing a pick-and-drop operation. Starting from a designated source stack of boxes, the arm grabs a box from the source stack and stacks it vertically onto a target stack where both the source and the target stacks are on the opposite side of the arm.
    
    Length of arm: $l_1 = 10$ units and $l_2 = 10$ units.
    
    \begin{align*}
    \eventually({\code{reach}(\mathsf{top\_block\_target\_tower}})); \eventually( \code{reach}(\mathsf{top\_block\_source\_tower})
    \end{align*}

    \item \textbf{Pick and Horizontal Stack - Same Side (Figure~\ref{fig:r5})}

    The task involves performing a pick-and-drop operation. Starting from a designated source stack of boxes, the arm grabs a box from the vertical stack and stacks it horizontally along the x-axis onto a target stack where both the source and the target stacks are on the same side of the arm.
    
    Length of arm: $l_1 = 10$ units and $l_2 = 10$ units.
    
    \begin{align*}
    \eventually({\code{reach}(\mathsf{leftmost\_block\_target\_tower}})); \eventually( \code{reach}(\mathsf{top\_block\_source\_tower})
    \end{align*}
\end{itemize}

\section{OpenAI Gym Classic Control Benchmarks}
\label{sec:gymexp}

\subsection{Cartpole}

\paragraph{Environment Description.} The objective in the Cartpole environment is to balance a pole on a moving cart by applying force to the cart's base. The action space is discrete, with two possible actions: moving the cart left or right. The observation space includes the cart position, cart velocity, pole angle, and pole angular velocity.

Predicate $\code{holdpole}$ can be defined as:

\begin{align*}
\code{holdpole}(goal) = (|\theta - \text{goal}| < \epsilon) \text{ for time } t 
\end{align*}

where $\theta$ is the angle of the pole.

\paragraph{RL Specifications.}  The goal is to balance a pole on a moving cart by applying force to the cart's base i.e. reach a certain theta $goal$ and hold it for time $t$.

$$\eventually{(\code{holdpole}(goal))}$$. 

The induction is on the length $l$ of the cartpole which involves increasing the length by $l + 0.4$ units. The training range for $l$ is from 0.4 to 2 units.

\subsection{Pendulum}

\paragraph{Environment Description.} The goal in the Pendulum environment is to keep a free pendulum standing up by applying torque at the pivot point. The action space consists of a single continuous value representing the torque applied to the pendulum's free end. The observation space provides the x-y coordinates of the pendulum's free end and its angular velocity, represented by cosine and sine of the angle and the angular velocity.

Predicate $\code{reachtheta}$ can be defined as:

\begin{align*}
\code{reachtheta}(goal) = (|\theta - \text{goal}| < \epsilon)
\end{align*}

where $\theta$ is the angle of the pendulum.

\paragraph{RL Specifications.} The goal is to keep a free pendulum standing up by applying torque at the pivot point i.e. reach a certain theta $goal$.

$$\eventually{(\code{reachtheta}(goal))}$$

The induction is on the mass $m_p$ of the pendulum which involves increasing the mass by $m_p + 0.1$ units. The training range for $m_p$ is from 1 to 1.4 units.

\subsection{Acrobot}

\paragraph{Environment Description.} The Acrobot's challenge is to swing up the end of a two-link robot arm above a certain threshold. The action space is discrete and deterministic, representing the torque applied at the joint between the two links. 

Predicate $\code{reachtip}$ can be defined as:

\begin{align*}
\code{reachtip}(goal) = (|-\cos(\theta_1) - \cos(\theta_2 + \theta_1)| > \epsilon)
\end{align*}

where $\theta_1$ is the angle of the first joint, and $\theta_2$ is the angle of the second joint w.r.t the first joint.

\paragraph{RL Specifications.} The goal is to swing up the end of a two-link robot arm above a certain threshold i.e the tip must reach a certain point $goal$

$$\eventually{(\code{reachtip}(goal))}$$ 

The induction is on the mass $m_a$ of the acrobot which involves increasing the mass by $m_a + 0.1$ units. The training range for $m_a$ is from 0.2 to 0.6 units.

\subsection{Observations}

Table~\ref{tab:combined} clearly shows that $\genrl$ satisfies a great number unseen tasks across multiple iterations in the control benchmarks. This shows $\genrl$s capability to even inductively generate policies where the update is on environment parameters rather than on specification parameters. More detailed results and observations in Section~\ref{sec:longhorizoncompdyn}.

\begin{table}[t]
\centering
\setlength\tabcolsep{2pt}
\begin{tabular}{|c|c|c|c|c|c|c|}
\hline
\textbf{Benchmark} & \multicolumn{5}{c|}{\textbf{Iterations}} & \textbf{\% Gen.}\\ \cline{2-6} 
   $|\traintasks| = 5$ & \textbf{100} &\textbf{200} &\textbf{300} & \textbf{400} & \textbf{500} & \textbf{(Best Iter)}\\ 
\hline
Cartpole &  9 & 11 &  11 & \textbf{14} &  12 & 280\\
Pendulum  & 9 & 8 & 10 & 9 & \textbf{12} & 240\\
Acrobot  & 0 & 6 & 7  & 9 & \textbf{10} & 200\\
\hline
\end{tabular}
\caption{No. of successful Unseen task instances for OpenAI Gym Classic Control experiments. The number in \textbf{boldface} under Iterations represents the best generalization for the benchmark.}
\label{tab:combined}
\end{table}

\section{Template Complexity Analysis}
\label{sec:template}

\begin{figure}[t]
    \centering
    \includegraphics[width=0.5\linewidth]{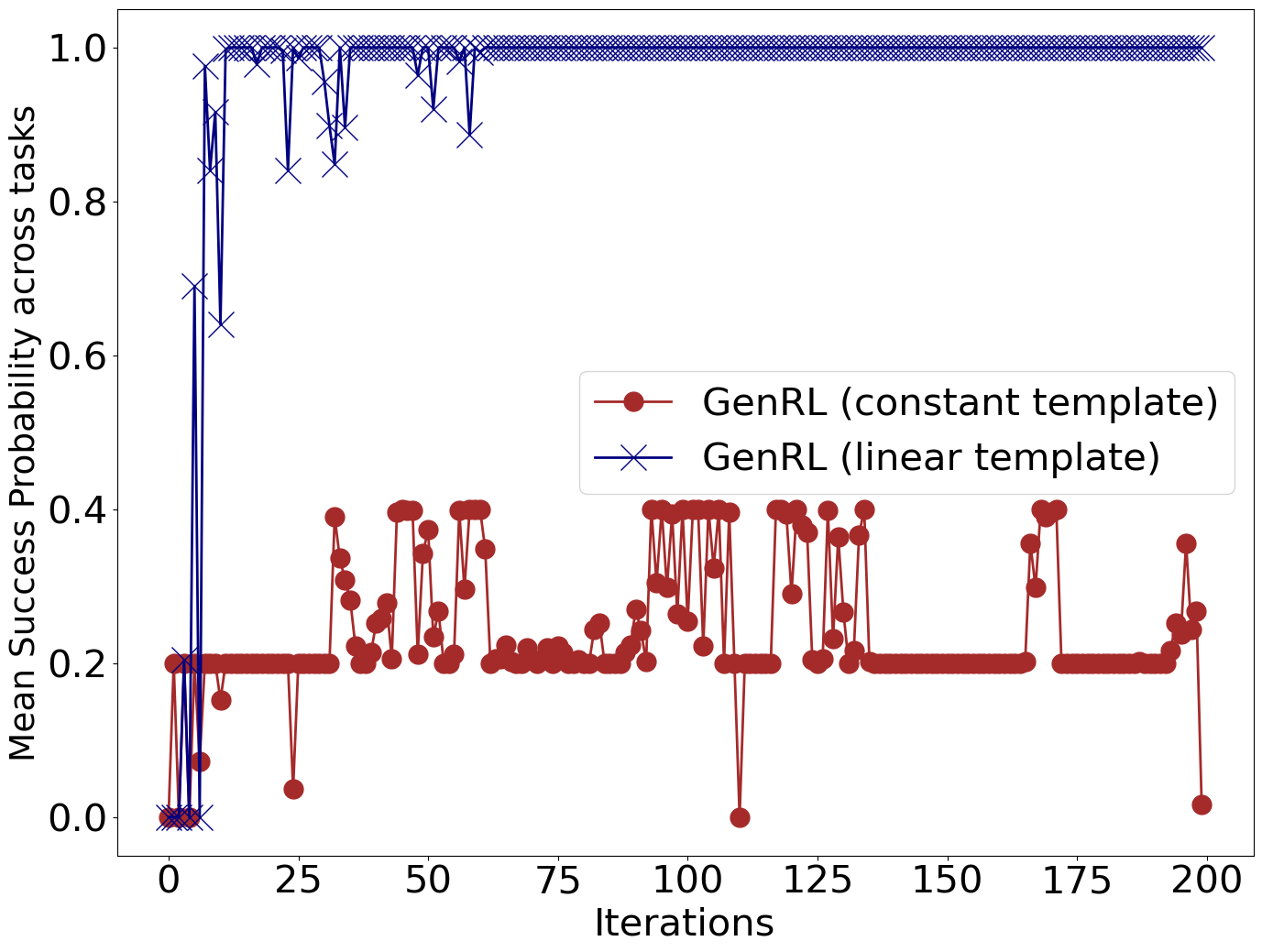}
    \caption{Template Comparison: Mean Succ. Prob. on $\traintasks$ in  1-Reachability Inductive Task with moving initial distribution}
   \label{fig:msp_both}
\end{figure}

The complexity of the policy generator template offers a tradeoff between generalizability and the difficulty of learning (Figure~\ref{fig:msp_both}): the x-axis represents the number of training iterations and the y-axis represents the success probabilities across all tasks, $\task_i \in \traintasks$ (each task shown in a different color). For successful learning, the success probability should increase with more training iterations and saturate at a high probability. We consider two templates:  (a). \textbf{constant update template}  $\pz{i+1} = \pz{i} + \kappa_0$ that only requires learning $\kappa_0$ and (b) \textbf{linear template} $\pz{i+1} = \kappa_{1}\cdot \pz{i} + \kappa_0$, with two coefficient vectors $\kappa_0$ and $\kappa_1$ to be learned. 

Figure~\ref{fig:msp_both} shows the mean success probability over the tasks $\task_i \in \traintasks$: \textit{constant update template} oscillates around a low success probability of around 0.2, demonstrating that this template is unable to learn a reasonable policy generator. On the other hand, \textit{linear template} seems to be effective, converging to a success high probability close to 1.0. At the same time, the linear templates has double the number of trainable parameters as compared to constant update template, making it a more involved training exercise. This study outlines the importance of selecting appropriate templates to learn the policy generator successfully.

\end{document}